\documentclass[conference]{IEEEtran}
\usepackage{cite}
\usepackage{amsmath,amssymb,amsfonts}
\usepackage{graphicx}
\usepackage{textcomp}
\usepackage{xcolor}
\def\BibTeX{{\rm B\kern-.05em{\sc i\kern-.025em b}\kern-.08em
    T\kern-.1667em\lower.7ex\hbox{E}\kern-.125emX}}
\usepackage{algorithm}
\usepackage{algorithmicx}
\usepackage{algpseudocode}
\usepackage{mathtools}

\usepackage{physics}

\usepackage{multirow}
\usepackage{pifont}

\usepackage{tabulary}
\usepackage{siunitx} 

\usepackage{subcaption}
\usepackage[font=small]{caption}
\usepackage[font=small]{subcaption}

\newcommand*{\QEDB}{\hfill\ensuremath{\blacksquare}}
\newcommand{\hide}[1]{}
\usepackage{enumitem}

\usepackage{makecell}

\usepackage{nicematrix,enumitem,booktabs} %for table footnote
\begin{document}

\title{\huge FedADMM: A Robust Federated Deep Learning Framework with Adaptivity to System Heterogeneity
\thanks{$\ast$ Equal contribution. $\dagger$ Corresponding author.}
}

\author{\IEEEauthorblockN{Yonghai Gong$^{\ast}$ $^1$, Yichuan Li$^{\ast}$ $^2$ and Nikolaos M. Freris$^{\dagger}$ $^1$}
\IEEEauthorblockA{
\textit{$^1$School of Computer Science and Technology, University of Science and Technology of China, Hefei, China}
 \\
\textit{$^2$Coordinated Science Laboratory, University of Illinois at Urbana-Champaign, IL 61820, USA}\\
gongyh@mail.ustc.edn.cn, yli129@illinois.edu, nfr@ustc.edu.cn}

%\and
%\and
%\IEEEauthorblockA{\textit{dept. name of %organization (of Aff.)} \\

}

% \author{\IEEEauthorblockN{1\textsuperscript{st} Yonghai Gong}
% \IEEEauthorblockA{\textit{School of Computer Science and Technology} \\
% \and
% \IEEEauthorblockN{1\textsuperscript{st} Yichuan Li}
% \and

\maketitle

\renewcommand{\thefootnote}{\fnsymbol{footnote}}
\footnotetext[1]{Equal contribution. $^\dagger$Corresponding author: Nikolaos M. Freris. This paper is accepted at the 38th IEEE International Conference on Data Engineering.} 

\renewcommand{\thefootnote}{\arabic{footnote}}

\begin{abstract}
% Federated Learning (FL) is an emerging framework for distributed processing of large data volumes by edge devices subject to limited communication bandwidths, heterogeneity in data distributions and computational resources, and privacy considerations. In this paper, we propose a new FL protocol termed \texttt{FedADMM} based on primal-dual optimization. The proposed method uses dual variables to tackle statistical heterogeneity and further accommodates system heterogeneity by tolerating variable amount of work performed by clients. \texttt{FedADMM} maintains identical communication costs per round as \texttt{FedAvg/Prox}, and generalizes them through the use of augmented Lagrangian. A convergence proof is established for general nonconvex objectives under less restrictive assumptions than the state-of-the-art, along with a general client participation scheme that imposes no restrictions on number of clients selected per round. We demonstrate the advantages of \texttt{FedADMM} through extensive experiments on real datasets, under both IID and non-IID data distributions across clients. Our experiments show that \texttt{FedADMM} consistently outperforms all baseline methods and achieves higher communication efficiency, with the number of rounds needed to reach a prescribed accuracy being reduced by up to $87\%$. Moreover, we show that \texttt{FedADMM} effectively adapts to heterogeneous data distributions through the use of dual variables without the need for hyperparameter tuning. 
Federated Learning (FL) is an emerging framework for distributed processing of large data volumes by edge devices subject to limited communication bandwidths, heterogeneity in data distributions and computational resources, as well as privacy considerations. In this paper, we introduce a new FL protocol termed \texttt{FedADMM} based on primal-dual optimization. The proposed method leverages dual variables to tackle statistical heterogeneity, and accommodates system heterogeneity by tolerating variable amount of work performed by clients. \texttt{FedADMM} maintains identical communication costs per round as \texttt{FedAvg/Prox}, and generalizes them via the augmented Lagrangian. A convergence proof is established for nonconvex objectives, under no restrictions in terms of data dissimilarity or number of participants per round of the algorithm.  We demonstrate the merits through extensive experiments on real datasets, under both IID and non-IID data distributions across clients. \texttt{FedADMM} consistently outperforms all baseline methods in terms of communication efficiency, with the number of rounds needed to reach a prescribed accuracy reduced by up to $87\%$. The algorithm effectively adapts to heterogeneous data distributions through the use of dual variables, without the need for hyperparameter tuning, and its advantages are more pronounced in large-scale systems.
\end{abstract}

% \begin{IEEEkeywords}
% federated learning, optimization, heterogeneous system
% \end{IEEEkeywords}

\section{Introduction} 
The abundance of data constitutes the fuel for the proliferation of Machine Learning (ML) in all aspects of life. In many applications, data are being generated in large quantities in a distributed fashion: for instance, crowdsourcing, mobile phones, autonomous vehicles, medical centers, distributed sensors in smart grids, to name but a few. The sheer size of the data alongside the urge to protect personal privacy prevent traditional distributed machine learning practices from being directly applied. The well-studied framework of data-center distributed learning \cite{datacenter2,datacenter1} often involves maneuvering raw data from machine to machine, which not only violates privacy restrictions but also becomes infeasible when limited networking resources fail to cope with large data sizes. Besides, the rapid increase of the computational power of personal devices such as smartphones surges pushing computation to the edge as opposed to the cloud. 

\begin{figure}[t]
\centering
\includegraphics[width=0.9\columnwidth]{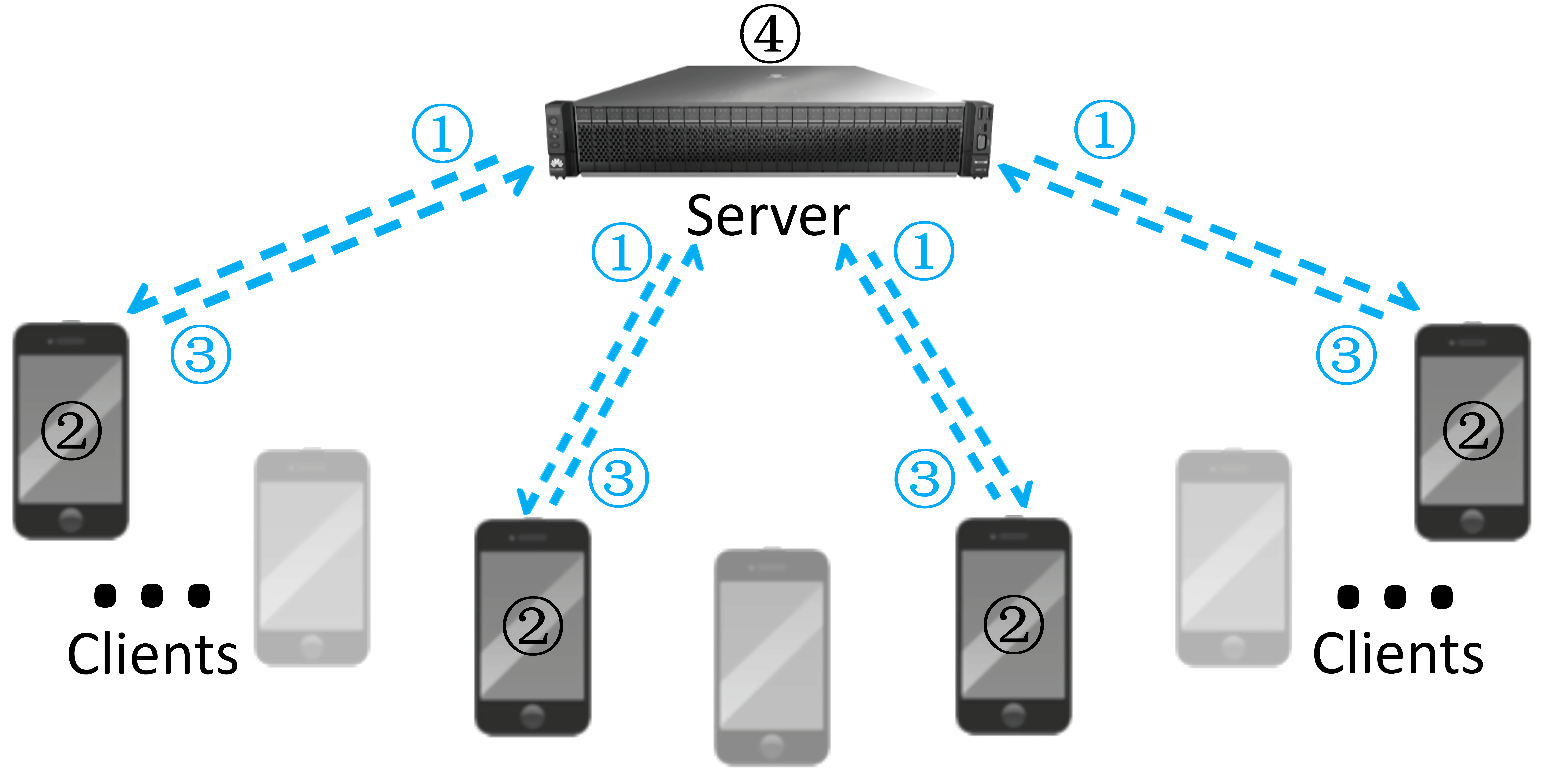} % Reduce the figure size so that it is slightly narrower than the column. Don't use precise values for figure width.This setup will avoid overfull boxes.
\caption{Architecture of FL. Active clients \ding{172} download the current global model from the server, \ding{173} perform local training based on their data, \ding{174} upload an update message to the central server. Server \ding{175} aggregates clients' updates to form a new global model. \hide{\ding{172} and \ding{174} are communication steps, \ding{173} and \ding{175} are computation steps.}}
\label{fig1}
\end{figure}

Federated Learning (FL) was proposed in \cite{Fed1,Fed2} as a new paradigm for tackling large-scale distributed machine learning problems with distinct features compared to the data-center setting: 
(i) massive client population; (ii) limited communication resources; (iii) low device participation rate and unreliable connections; (iv) stringent privacy considerations. A canonical setting of FL is shown in Fig. \ref{fig1}, where a server coordinates with clients to compute a global ML model. 
The introduction of this framework was carried in \cite{Fed2}, wherein the authors proposed \texttt{FedAvg} as the main algorithm. At each round of \texttt{FedAvg}, a small fraction of clients is selected and each selected client downloads the global model. The selected clients then proceed to update the model by running SGD using their local data, and upload the updated model to the server. Finally, a new global model is obtained by the server through averaging the updated models. Despite the fact that \texttt{FedAvg} is arguably the most widely adopted method of choice for FL, it does not address either \emph{system heterogeneity} or \emph{statistical heterogeneity}. Specifically, \texttt{FedAvg} does not explicitly handle the \textit{straggler} problem in a heterogeneous network: the process may stall in the face of unreliable network connections or client unavailability. Moreover, \texttt{FedAvg} may diverge when data distributions across clients violate the IID (independent and identically distributed) assumption \cite{Fed2,Fednoniid,FedAvg_conv2019,FedAvg_conv2020}. 

The authors in \cite{Fedprox} proposed \texttt{FedProx}, which augments a proximal term to the local training problem to counter possible divergence resulting from statistical heterogeneity. Additionally, this method accommodates system heterogeneity by allowing variable amount of work at each selected client. Nevertheless, the performance of \texttt{FedProx} is sensitive to the selection of the proximal coefficient, whose tuning depends on prior knowledge on statistical distributions and system sizes \cite{Fedprox}.
% Nevertheless, the convergence speed of the method is sensitive to proper tuning of the proximal term; this depends on the statistical distributions, as we have also observed in our experiments.
\texttt{SCAFFOLD} \cite{scaffold} introduces client and server \textit{control variates}, which effectively act as tracking variables for local and global information. It is reported to outperform \texttt{FedAvg/Prox} through utilizing the similarity of data across clients, but its communication cost per round is doubled due to the extra \textit{control variates}. 

Another branch of work has roots in primal-dual optimization methods, such as Generalized Method of Multipliers and Alternating Direction Method of Multipliers (ADMM) \cite{pd1,admm1,admm2} that have enjoyed much success in distributed optimization. Such methods provide a general framework which not only encompasses a wide range of objectives---such as low-rank matrix completion, nonconvex neural network training, and reinforcement learning \cite{LR-ADMM,NN-ADMM,RL-ADMM}---but it can also be tailored for specific application demands, such as event-triggered communication and inexact computing \cite{event_admm,inexact_admm,cocoa,dnadmm,bfgsadmm}.

In this paper, we propose \texttt{FedADMM}, a primal-dual algorithm that automatically adapts to data heterogeneity while achieving significant convergence speedup compared to the state-of-the-art. Each round of \texttt{FedADMM} involves a small fraction of clients and allows a plethora of computing choices for per user. A distinctive attribute of \texttt{FedADMM} is the use of dual variables to guide the local training process (see Section \ref{3.1} for detailed discussion), which effectively account for automatic adaptation to data distributions without tuning. This specifically addresses a key challenge encountered in the FL setting known as avoiding \emph{client drift}: local training performed at clients has to be carefully designed according to statistical variations so as to prevent the model from overfitting to a specific selected client's data. We note that compared to \texttt{SCAFFOLD}, where the acceleration is achieved through storing and communicating the \textit{control variates} that capture local gradient information, \texttt{FedADMM} maintains the exact same communication cost per round as \texttt{FedAvg/Prox}. Using a limited amount of extra storage for the dual variables (which is readily available and inexpensive in practical FL scenarios), \texttt{FedADMM} effectively harnesses the capabilities of edge devices and achieves significant communication and computation savings due to the automated adaptation to statistical variations. 

\noindent \textbf{Contributions}:

\indent We present a new framework for Federated Learning, termed \texttt{FedADMM}.\begin{itemize}[noitemsep,topsep=0pt]
    \item Through the use of local augmented Lagrangian and dual variables, \texttt{FedADMM} addresses both system and statistical heterogeneity. 
    \item By introducing a generalized server update step size, the proposed method achieves a fine balance between convergence speed and robustness against oscillations.
    \item We establish a convergence rate of $\mathcal{O}(\tfrac{1}{T})+\mathcal{O}(1)$ for the proposed algorithm under less restrictive assumptions than the state-of-the-art. Moreover, this rate matches the complexity lower bound for reaching an $\varepsilon$-stationary solution.
    \item A crucial feature of \texttt{FedADMM} is that no restrictions are imposed on the client activation scheme other than the necessary infinitely often participation. This is achieved by clients solving local augmented Lagrangian subproblems that are strongly convex, which ensures progress is made without diverging under a most general participation scheme.\hide{A crucial feature of \texttt{FedADMM} is that a constant hyperparameter can be employed without prior knowledge on system sizes, number of participating clients, and data distributions. Extensive experiments across various settings corroborate this observation.}
    \hide{The proposed solution concurrently addresses both system and statistical heterogeneity, and outperforms the baselines in terms of convergence speed, communication efficiency, as well as robustness against statistical variations.}
    \hide{\item We provide a convergence proof of the algorithm without imposing assumptions on data distributions. The rate of convergence $\mathcal{O}(\tfrac{1}{T})+\mathcal{O}(1)$ is established by selecting a constant hyperparameter without prior knowledge on system sizes, number of participating clients, and data distributions. Our result is also confirmed by extensive experiments showing that constant hyperparameters work well under various settings.}
    \item We have conducted numerous experiments on distributed neural network training from real datasets in both IID and non-IID settings, which attest that \texttt{FedADMM} achieves high accuracy in fewer rounds compared to the state-of-the-art (an average $72\%$ and a maximum $87\%$ of communication savings over the best performing baseline are reported over several datasets, population sizes, and both statistical and system heterogeneity). Notably, the advantages of \texttt{FedADMM} are most pronounced in large-scale networks with heterogeneous data distributions. 
\end{itemize}
\section{Related work}\label{section2}
Existing primal-dual algorithms for FL include \texttt{FedHybrid} \cite{fedhybrid} and \texttt{FedPD} \cite{fedpd}. \texttt{FedHybrid} is a synchronous method (requiring all clients to update at each round) that tackles system heterogeneity by separating clients into two groups, one performing gradient updates and the other Newton updates. However, this setting does not encapsulate federated deep learning (Newton's method is not applicable for large neural networks, while also convergence is established under strong convexity assumptions). \texttt{FedPD} employs gradient-based (GD or SGD) local training with variable amount of work at clients. Similar to our proposed method, it invokes dual variables to capture the discrepancy between local models and the global model. Nevertheless, \texttt{FedPD} adopts full client participation, in that (i) \emph{all clients} update their local models and dual variables at each round and (ii) at each round, with fixed probability, either \emph{all clients} communicate with the server to update the global model, or there is no communication. Therefore, the frequency of updating the global model is limited by the probability of communicating. We note that all clients in \texttt{FedHybrid/PD} are continuously engaged in computing, which incurs large computational overhead to devices. Moreover, requiring all clients to communicate at the same time not only imposes strict requirements on clients’ availability and network bandwidth, but also exacerbates the straggler problem (where the server has to wait for the slowest client before proceeding to the next round). \hide{The same holds true for computation, since full participation incurs large computational overhead to the devices.}

 In contrast, our proposed solution (i) applies partial client participation for both computation and communication, (ii) updates the global model at each round, and (iii) leverages tracking in the global model update rule (which accounts for accelerated convergence with smoother paths). The server in \texttt{FedADMM} communicates with a fraction of clients during each round and clients only perform local training when selected. We establish convergence without restrictions on statistical heterogeneity among the networks, or the fraction of participating users. This is in contradistinction to the analysis of other baseline methods \cite{Fedprox, Fednoniid,localsgd} which requires assumptions on data dissimilarity that may be unrealistic in practices when data are individually generated by each client \cite{edge_AI,FLapp2019,FLnetwork}. Our proposed method has close connections to randomized ADMM \cite{hongadmm}, which in the FL setting amounts to randomly activating clients and the server. However, it requires that subproblems to be solved exactly, which may be impractical in consideration of heterogeneous hardware conditions and data volumes. On the other hand, \texttt{FedADMM} accounts for inexact solutions from local problems and allows variable amount of training performed at clients (quantified by a tunable parameter). Another related line of work is termed asynchronous ADMM \cite{asynadmm14,asynadmm16,asynadmm2018}, whose focus is on alleviating the straggler problem mentioned earlier. However, assumptions imposed therein such as the \textit{bounded delay} assumption (i.e., each user needs to be active at least once every some number of rounds) may never be satisfied in FL settings. 

\begin{figure}[t]
\centering
\includegraphics[width=1\columnwidth]{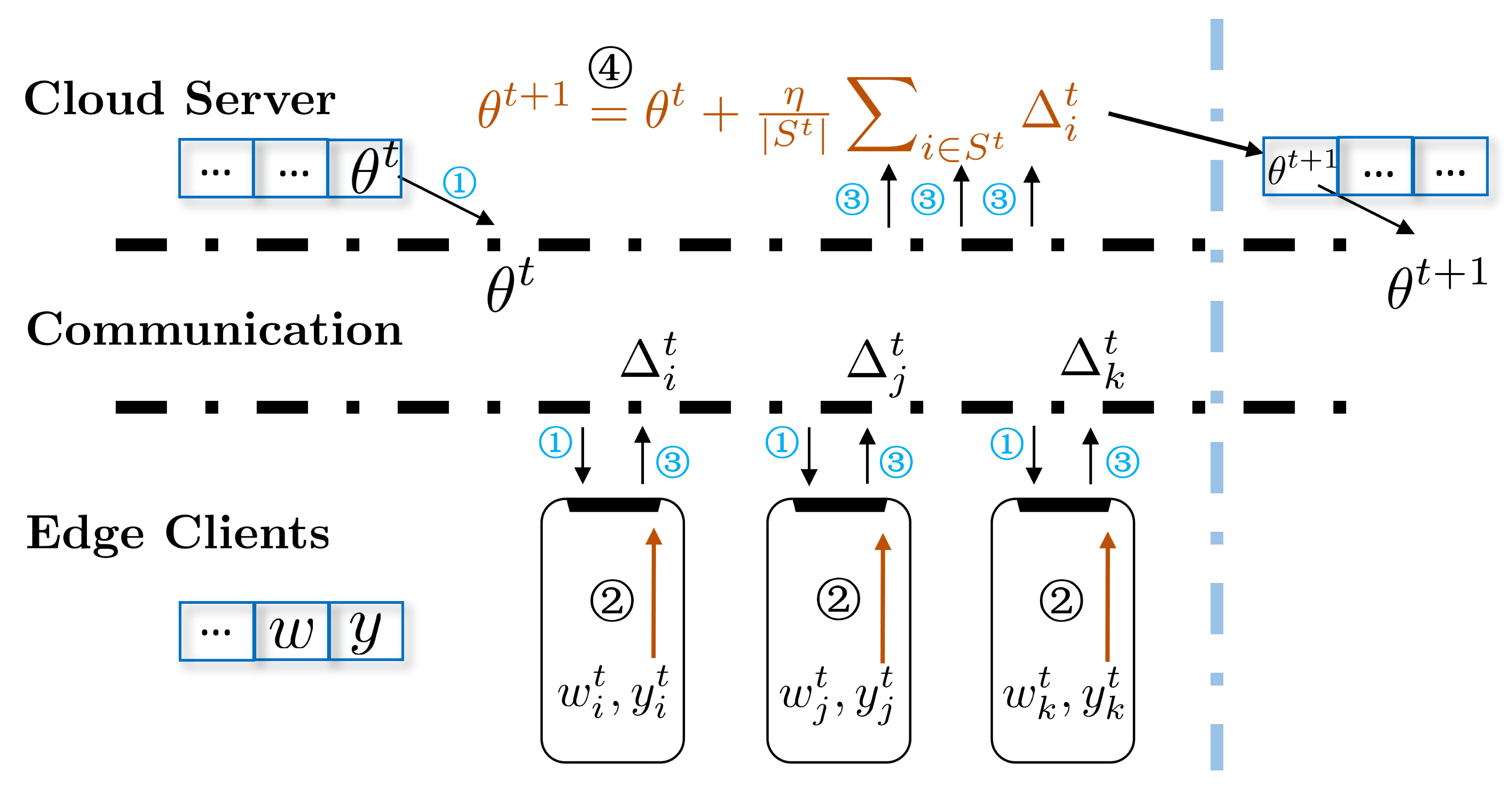}
\caption{\texttt{FedADMM} on selected clients during the $t$-th round. Selected clients \ding{172} download the current server model, \ding{173} carry local training and update their dual variables (Alg. \ref{algorithm 1}, line 6), \ding{174} upload \hide{\textit{augmented model}} update messages (Alg. \ref{algorithm 1}, line 8). Server \ding{175} aggregates messages from selected edge clients to update the global model.}
\label{fig2}
\end{figure}

\section{Algorithm} \label{section3}
The objective of FL is cast as a loss minimization problem: 
\begin{gather}
        \underset{\theta \in \mathbb{R}^d}{\mathrm{minimize}}\ \ \sum_{i=1}^{m}f_i(\theta),\label{prob1}
\end{gather}
where $m$ is the number of clients and $f_i(\theta)=\tfrac{\alpha_i}{n_i} \sum_{k=1}^{n_i}l_{ik}(\theta)$ captures the weighted local training loss. Common choices of weights include $\alpha_i=\frac{n_i}{n}$, where $n=\sum_{i=1}^m n_i$ (i.e., weighting clients proportionally to their data volumes) or $\alpha_i = 1$ (i.e., equal weights, which serves to avoid overfitting to clients with more data, and is also the choice in our experiments). The goal of FL is to solve (\ref{prob1}) over model parameter $\theta$ using data distributed across clients. In contrast to data-center distributed machine learning, 
% where computation resources are under strict budget, 
communication is the bottleneck for FL applications. The large number of clients in federated systems renders infeasible synchronous communication (i.e., all clients communicating with the server at a given round) with large message sizes; this is due to bandwidth limitations and the existence of stragglers. Moreover, the large data volumes involved along with the leakage of personal information from raw data prevents the server from collecting local data for centralized processing. Thus, a candidate algorithm for FL should meet the following requirements: (i) communication efficiency; (ii) low client participation rate per round; (iii) handling both system and statistical heterogeneity; (iv) no transmission of client raw data. We proceed to present \texttt{FedADMM} whose design meets all these guidelines\footnote{Although not the focus of this paper, we note that standard privacy-preserving methods, such as differential privacy and secure multi-party computation \cite{privacy_2008,privacy_2020,privacy2019} can be combined with \texttt{FedADMM}.}.

\subsection{Proposed Algorithm: \texttt{FedADMM}}\label{3.1}
We first reformulate (\ref{prob1}) into the following consensus setting that is more suitable for our development: 
\begin{gather}
            \underset{w_i,\theta \in \mathbb{R}^d}{\mathrm{minimize}}\ \ \sum_{i=1}^{m}f_i(w_i),\ \ \text{s.t.}\ \ w_i=\theta,\ \ \forall i \in [m].\label{prob2} 
\end{gather}
Problem (\ref{prob2}) is equivalent to (\ref{prob1}) in the sense that the optimal solutions coincide. This formulation naturally fits into the FL setting, where $w_i$ can be interpreted as the local model held by client $i$ and $\theta$ as the global model held by the server.

\hide{\begin{algorithm}[t]
\caption{\texttt{FedADMM}}\label{algorithm 1}
\begin{algorithmic}[1] %[1] enables line numbers
\State \textbf{Input}: Total number of rounds $T$, server step size $\eta$, proximal coefficient $\rho$.
\For{$t=0,1,\dots,T-1$}
    \State Server selects $S^t\subset [m]$
    \Statex \texttt{Clients}:
    \If{$i \in  S^t$}
    \State {download $\theta^{t}$ from the server.}
    \State {$w_i^{t+1}\leftarrow$\texttt{ClientUpdate}($w_i^t,y_i^t,\theta^t$)}
    \State {$y_i^{t+1} \leftarrow y_i^t + \rho (w_i^{t+1} - \theta^t)$}
    \State {compute update message $\Delta_i^t$ as in (\ref{updatemessage})}
    \State {upload $\Delta_i^{t}$ to the server}
    \EndIf 
    \Statex \texttt{Server:}
    \State {$\theta^{t+1} \leftarrow \theta^t + \frac{\eta}{\left | S^t \right |} \sum_{i \in S^t} \Delta_i^{t+1} $}
\EndFor
\Statex
\Statex
    \texttt{ClientUpdate$(\theta^t)$}:
    \State \textbf{Input}: Local epoch number $E_i$, client learning rate $\eta_i$.
    %\State {Initialize $w_i\leftarrow w_i^t$}
    \For{$k=0,1,\dots, E_i-1$}
    \State Draw local batch $b$ and compute batch gradient $\nabla f_i(w_i,b)$
     \State{$w_i\leftarrow w_i-\eta_i \left(\nabla f_i(w_i,b)+y_i^t+\rho(w_i-\theta^t)\right)$}
    \EndFor
    \State return $w_i$
\end{algorithmic}
\end{algorithm}}

\begin{algorithm}[t]
\caption{\texttt{FedADMM}}\label{algorithm 1}
\begin{algorithmic}[1] %[1] enables line numbers
\State \textbf{Input}: Total number of rounds $T$, server step size $\eta$, proximal coefficient $\rho$.
\For{$t=0,1,\dots,T-1$}
    \State \label{select}Server selects $S^t\subset [m]$
    \Statex \texttt{Clients}: // \textit{In parallel}
    \For{$i \in  S^t$} 
    \State {download $\theta^{t}$ from the server}
    \State {$(w_i^{t+1},y_i^{t+1})\leftarrow$\texttt{ClientUpdate}($i,\theta^t$)}
    \State {compute update message $\Delta_i^t$ as in (\ref{updatemessage})}
    \State {upload $\Delta_i^{t}$ to the server}
    \EndFor
    \Statex \texttt{Server:} 
    \State {$\theta^{t+1} \leftarrow \theta^t + \frac{\eta}{\left | S^t \right |} \sum_{i \in S^t} \Delta_i^{t} $}
\EndFor
\Statex
\Statex
    \texttt{ClientUpdate$(i,\theta)$}: // \textit{Store $w_i$ and $y_i$}
    \State \textbf{Input}: Local epoch number $E_i$, client learning rate $\eta_i$.
    \State create batches $\mathcal{B}$
    \For{$k=0,1,\dots, E_i-1$}
    \For {batch $b\in \mathcal{B}$}
    \State compute batch gradient $\nabla f_i(w_i,b)$
     \State{$w_i\leftarrow w_i-\eta_i \big(\nabla f_i(w_i,b)+y_i+\rho(w_i-\theta)\big)$}
    \EndFor
    \EndFor
    \State {$y_i \leftarrow y_i + \rho (w_i - \theta)$}
    \State return $w_i,y_i$
    \label{algorithm}
\end{algorithmic}
\end{algorithm}
At the beginning of the $t$-th round, a subset of clients is selected (denoted as $S^t$) and each selected client downloads the server model $\theta^t$. We note that a favorable attribute of \texttt{FedADMM} lies in that client selection (step \ref{select} of Alg. \ref{algorithm 1}) can be carried by any mechanism (e.g., an adaptive rule based on device operating conditions such as energy and bandwidth) as long as it guarantees non-zero probability of participation. Local training is performed to update the client model as $w_i^{t+1}\approx \underset{w_i}{\text{argmin}}\,\,\mathcal{L}_i(w_i,y_i^t,\theta^t)$, where 
\begin{gather}
    \mathcal{L}_i(w_i,y_i^t,\theta^t)=  f_i(w_i)+(y_i^t)^\top (w_i-\theta^t)
    +\frac{\rho}{2}\norm{w_i-\theta^t}^2. \label{AL}
\end{gather}
We note that $y_i^t\in\mathbb{R}^d$ is the local dual variable held by client $i$ and $\rho>0$ is the coefficient of the quadratic term. Both the dual variables and the quadratic term serve to strike a balance between updating model parameters using local data while staying consistent with the server model (which serves to incorporate information from all participants). We note that \texttt{FedProx} \cite{Fedprox} similarly solves (\ref{AL}) with $y_i^t\equiv 0$. While this provides some level of safeguard against \textit{client drift}, competitive performance of \texttt{FedProx} relies on careful tuning of $\rho$ \cite{Fedprox,scaffold}. The addition of the dual variables helps \texttt{FedADMM} to achieve automatic adaptation to heterogeneous data distributions and significantly alleviates the problem of hyperparameter tuning; we postpone detailed discussions on dual variables to the end of this section. In addition to the local model update (Alg. \ref{algorithm 1}, lines 14-19), the dual variable for each selected client is updated in line 20. We combine the primal and dual variables to what we call \textit{augmented model} $\big(w_i + \frac{1}{\rho} y_i\big)$, and use $\Delta_i^t\in \mathbb{R}^d$ to denote the update message of client $i$ to the server, that is the difference between successive \textit{augmented models}:
\begin{gather}
    \Delta_i^t = \left(w_i^{t+1}+\tfrac{1}{\rho}y_i^{t+1}\right)-\left(w_i^t+\tfrac{1}{\rho}y_i^t\right). \label{updatemessage}
\end{gather}
The server updates the global model after gathering update messages from selected clients as:
\begin{gather}
    \theta^{t+1}=\theta^t+\tfrac{\eta}{\abs{S^t}}\sum_{i\in S^t} \Delta_i^t, \label{server step size}
\end{gather}
where $\eta>0$ is the server gathering step size. Different choices of $\eta$ are suitable for different scenarios in terms of scale of the system and statistical variations. We empirically observe that setting $\eta=1$ gives rise to fast training speed, while setting $\eta=\abs{S^t}/m$ helps to eliminate oscillatory behaviors when significant heterogeneity is detected. In contrast to \texttt{FedAvg/Prox}, where the global model is updated using information only from the current models of clients $i\in S^t$, the server in \texttt{FedADMM} effectively incorporates past information. This is accomplished through the tracking update rule in (\ref{server step size}) and serves to provide additional safeguard against oscillations emanating from heterogeneous data and the stochastic nature of the algorithm. 

We emphasize that in a practical scenario (\ref{AL}) is not required to be solved exactly, but instead the updated local model $w_i^{t+1}$ can be computed inexactly in the following sense:
\begin{gather}
    \norm{\nabla_{w_i}\mathcal{L}_i(w_i^{t+1},y_i^t,\theta^t)}^2\leq \varepsilon_i ,\label{varepsilon_i}
\end{gather}
where $\varepsilon_i\geq 0$ prescribes the local accuracy. Note that a variable accuracy level is allowed for different clients (a smaller $\varepsilon_i$ corresponds to a more accurate solution). This is useful since user devices in FL systems feature varying hardware conditions such as computational power, network connectivity, battery level, etc. For the sake of simplicity and comparison with baseline methods, this is done in Algorithm \ref{algorithm 1} by running $E_i$ epochs of SGD to compute $w_i^{t+1}$ (line 14); the equivalence with (\ref{varepsilon_i}) emanates from the fact that local subproblems (\ref{AL}) are strongly convex. Nevertheless, other updating schemes are also feasible such as gradient descent and quasi-Newton updates like L-BFGS. Therefore, we accommodate system heterogeneity by letting clients decide to perform different amount of work according to their local environments. Finally, we note that each client in \texttt{FedADMM} holds the primal-dual pair $(w_i,y_i)$, while it only communicates the augmented model difference $\Delta_i^t$, i.e., the communication cost per round is identical with \texttt{FedAvg} and \texttt{FedProx}.

\noindent\textbf{Dual variables}: Contrary to primal only methods (\texttt{FedAvg/Prox} and \texttt{SCAFFOLD}), the augmented Lagrangian $\mathcal{L}_i$ is used for the training of local model parameters $w_i$. Invoking the augmented Lagrangian is effective for two reasons: (i) by a proper selection of the quadratic proximal coefficient $\rho$, it is possible to make $\mathcal{L}_i$ in (\ref{AL}) strongly convex with respect to $w_i$ (this is true even for non-convex $f_i$); therefore, theoretical guarantees for reaching a prescribed local accuracy level $\varepsilon_i$ in (\ref{varepsilon_i}) are readily obtainable. (ii) the dual variable $y_i$ in (\ref{AL}), in addition to penalizing the discrepancy between the local model and the global model, further provides a quantitative measure of the benefit brought by letting local models be different from the server model during the initial stage of training. To illustrate, note that the stationary points for each subproblem, i.e., the solutions to $\nabla f_i(\theta) = 0$, tend to be different from one another, especially in FL settings with heterogeneous data distributions across clients. The stationary condition of (\ref{prob1}), $\sum_{i=1}^m \nabla f_i(\theta^\star)=0$, indicates that the cost reduction by following any given $-\nabla f_i(\theta^\star)$ may be partly or entirely cancelled by the increase of other $f_j(\theta^\star), j\neq i$. This is encoded within the KKT conditions for problem (\ref{prob2}): $\nabla f_i(w_i^\star)+y_i^\star = 0$ for all $i\in [m]$ and $\sum_{i=1}^m y_i^\star=0$. Therefore, $y_i\in\mathbb{R}^d$ can be interpreted as a signed ``price vector'' (with entries being positive or negative) which not only quantifies the cost of $w_i^{t+1}$ being different from $\theta^t$, but also provides a \emph{direction} of the adjustments needed for agreement. \hide{This is achieved as $y_i$ accumulates the discrepancy between the client model and the server model $w_i^{t+1}-\theta^t$ as in line 17 of the algorithm. When a price entry $[y_i]_k$ is positive (which roughly indicates $[w_i-\theta]_k>0$), the corresponding model entry $[w_i]_k$ tends to decrease so as to minimize the cost in (\ref{AL}), and vice versa.} This effectively achieves an automatic adaptation to statistical variations in that it safeguards against \textit{client drift}, i.e., $w_i$ converging to client optima.  

\hide{\begin{table}[t]
\centering
\resizebox{.95\columnwidth}{!}{
\begin{tabular}{c|c|c}
 \hline
 \textbf{Algorithm} & \textbf{Message size}  &  \textbf{Storage} (server, client)  \\  
 \hline
 \texttt{FedAvg/Prox} & $d$ & $(d,0)$\\ 
 \hline
  \texttt{SCAFFOLD} & $2d$ & $(2d,d)$ \\ 
  \hline 
  \texttt{FedADMM}& $d$ & $(d,2d)$ \\
  \hline
\end{tabular}}
\caption{Message size and storage costs comparison in terms of model dimension $d$.}
\label{table1}
\end{table}}

\subsection{Connections with Existing Work}\label{3.2}
We show how \texttt{FedADMM} generalizes \texttt{FedAvg} and \texttt{FedProx}. Recall the augmented Lagrangian defined in (\ref{AL}). By setting $y_i \equiv 0$ (i.e., also omitting line 20 in Alg. \ref{algorithm 1}), we recover the local training problem of \texttt{FedProx}. If additionally $\rho$ is set to $0$, one recovers the local training problem of \texttt{FedAvg}. The main motivation of these two terms is to tackle a fundamental challenge reminiscent of FL, i.e., the balance between (i) extensive local training (aiming for faster convergence speed of the entire FL process, thus a reduction of the total number of communication rounds), and (ii) deterring client models from overfitting to their local data. In addition to the quadratic proximal term introduced in \texttt{FedProx}, \texttt{FedADMM} further employs dual variables, whose merits in guiding the update of local models $w_i$ were discussed in the prequel. We note that \texttt{SCAFFOLD} similarly stores additional variables at clients and the server to counter statistical variations. However, the variable introduced therein can not be combined into a single message (as is accomplished with our \textit{augmented model} and update messages in (\ref{updatemessage})), whence \texttt{SCAFFOLD} \emph{doubles} the size of the uploading message to the server. On the other hand, \texttt{FedADMM} only stores an additional vector at clients but maintains the exact same upload size. To this end, we believe \texttt{FedADMM} is an effective scheme for harnessing local device capabilities to alleviate
communication costs, which constitutes a primordial motivation for FL. Extensive experiments illustrate a substantial reduction of communication in all test cases ($72\%$ on average) over the best performing baseline.

\section{Analysis} 
\label{section 4}
In this section, we provide the convergence analysis for \texttt{FedADMM}. We adopt the following standard assumptions.\\
\noindent \textbf{Assumption 1}. Each local loss function $f_i(\cdot)$ is $L$-Lipschitz smooth, i.e., $\forall\,w,w^{\prime}\in\mathbb{R}^d$, the following inequality holds:
\begin{gather*}
    \norm{\nabla f_i(w)-\nabla f_i(w^\prime)}\leq L\norm{w-w^\prime},\,\, i\in[m].
\end{gather*}
\noindent\textbf{Assumption 2}. The objective of problem (\ref{prob1}) is lower bounded, i.e., there exists $f^\star\in \mathbb{R}$ such that 
$
    \sum_{i=1}^m  f_i(w)\geq f^\star ,\forall\,w\in\mathbb{R}^d
$.

The above assumptions are minimally restrictive, since loss functions are typically non-negative and the structure of the deep neural networks considered in the experiments ascertains Lipschitz smoothness. We further denote the aggregated Lagrangian as $\mathcal{L}=\sum_{i=1}^m \mathcal{L}_i$ and we define a non-negative function $V\left (\{w^t_i\},\{y^t_i\},\theta^t \right )$ to measure the optimality gap:
\begin{gather}
    V^t := \norm{\nabla_\theta \mathcal{L}^t}^2+\sum_{i=1}^m \left(\norm{\nabla_{w_i}\mathcal{L}^t}^2+\norm{w_i^t-\theta^t}^2\right).\label{V^t}
\end{gather}
Since each term of $V^t$ is non-negative, $V^t=0$ if and only if $\norm{\nabla_\theta\mathcal{L}^t}=\norm{\nabla_{w_i}\mathcal{L}_i^t}=\norm{w_i^t-\theta}=0$, $\forall\,i$. It can be further verified that these terms are $0$ if and only if a stationary solution of (\ref{prob2}) is reached. \hide{Using $V^t$ defined in (\ref{V^t}), we show that \texttt{FedADMM} converges sublinearly for general nonconvex problems without \emph{imposing any assumptions on data distributions}}

\noindent \textbf{Theorem 1}. Let assumptions 1 and 2 hold. Assume each client has a probability of being selected at each round that is lower bounded by a positive constant $p_{\mathrm{min}}>0$. Let $\eta=\abs{S^t}/m$, and select $\rho>(1+\sqrt{5})L$, then the following holds:
\begin{gather}
   \tfrac{1}{mT}\sum_{t=0}^{T-1} \mathbb{E}[V^t] \leq \tfrac{1}{mT}\tfrac{c_2}{c_1}\left(\mathcal{L}^0-f^\star+\tfrac{m}{2L}\varepsilon_{\mathrm{max}}\right)+c_3\varepsilon_{\mathrm{max}} ,\label{theorem1}
\end{gather}
where $\varepsilon_\mathrm{max}= \max_i\,\varepsilon_i$, $c_1 = p_{\mathrm{min}}\left(\tfrac{\rho-2L}{2}-\tfrac{2L^2}{\rho}\right)$, $c_2 =3(L^2+\rho^2)+2(1+\tfrac{2L^2}{\rho^2})$, and $c_3 = \left(3+\tfrac{16}{\rho^2}+\tfrac{c_2}{c_1}\cdot \tfrac{\rho+16L}{2L\rho}\right)$.

\noindent \textbf{Remark 1}: Note that the loss functions considered in our experiments admit $L=\mathcal{O}(1)$, i.e., the Lipschitz constant does not scale with the client population
(this is because the input data are uniformly bounded, e.g., images). To reach an $\varepsilon$-exact solution in such cases, \textbf{Theorem 1} suggests a complexity of $\mathcal{O}(1/\varepsilon p_{\mathrm{min}})$ by setting $\rho\in \mathcal{O}(L)=\mathcal{O}(1)$. We compare the convergence property of \texttt{FedADMM} with other baselines in Table \ref{conv_comp}. Note that only \texttt{FedProx} and \texttt{FedPD} among baseline methods enjoy the same favorable dependency on solution accuracy as \texttt{FedADMM}, i.e., $\mathcal{O}(1/\varepsilon)$. However, they both impose additional restrictive conditions to achieve this. In particular, \texttt{FedProx} requires selecting $S>B^2$, where $B$ is a measure of data dissimilarity across users as in (\ref{fedprox assumption}). This imposes constraints on minimum number of active clients. On the contrary, our analysis for \texttt{FedADMM} establishes convergence under arbitrary client participation and does not require (\ref{fedprox assumption}) (i.e., $B=\infty$ is also acceptable). Besides, \texttt{FedPD} requires all clients to communicate simultaneously (i.e., it does not allow partial client participation), which we believe is unrealistic for large networks where stragglers and unavailability of machines is ubiquitous. On the other hand, \texttt{SCAFFOLD} and \texttt{FedADMM} both allow arbitrarily large values of $G$ (gradient norm bound as in (\ref{grad_assump})) and $B$, but \texttt{SCAFFOLD} doubles the communication cost per client compared to \texttt{FedADMM}.

\begin{table}[t]
    \caption{Number of communication rounds required to reach an $\varepsilon$-stationary solution. Constants $B,G$ measure data dissimilarity and boundedness of gradients in (\ref{fedprox assumption}) and (\ref{grad_assump}) respectively, $m$ is the total number of clients, and $S$ denotes the number of active clients. }
    \centering
    \begin{tabular}{lcccc}
    \hline\hline
    Method & Number of communication rounds  \\
    \hline
    FedAvg\cite{Fed2,scaffold}& $\mathcal{O}(1/\varepsilon^2\cdot(m-S)/mS+G/\varepsilon^{3/2}+B^2/\varepsilon)$ \\
    FedProx$^1$\cite{Fedprox}      & $\mathcal{O}(B^2/\varepsilon)$   \\
    SCAFFOLD\cite{scaffold}    & $\mathcal{O}(1/\varepsilon^2+1/\varepsilon\cdot(m/S)^{\tfrac{2}{3}})$ \\
    FedPD$^2$ \cite{fedpd}        & 
    $\mathcal{O}(1/\varepsilon)$    \\
    FedADMM    & 
    $\mathcal{O}(1/\varepsilon\cdot (m/S))$\\
    \hline
    \multicolumn{2}{l}{$^1$\footnotesize \texttt{FedProx} requires $S> B^2$ to ensure convergence. }\\
    \multicolumn{2}{l}{$^2$\footnotesize \texttt{FedPD} requires all clients to communicate at the same time.} 
    \end{tabular}
    \label{conv_comp}
\end{table}

\noindent \textbf{Remark 2}: For simplicity, we stated \textbf{Theorem 1} assuming users are active with a non-zero probability at each round. In fact, the weaker condition is needed that each client is active infinitely often, which is necessary for the correctness of any iterative distributed algorithm (otherwise some clients are ignored). This in turn allows for a general activation scheme and thus incurs no lower bound on minimum number of active clients per round.
% This allows arbitrary activation scheme for the clients without imposing lower bounds on minimum number of active clients. 
The reason that this is possible is the introduction of dual variables and the quadratic penalty term which alleviate the effect of possible unbalanced client activation. Further insight can be gained by inspecting the inequality (\ref{condi}), where we derive the expected decrement of the augmented Lagrangian. This suggests that any activation scheme can be supported by our analysis: using time dependent participating probability $p_i^t$ for each client, our convergence proof carries as long as $\sum_{t=1}^\infty p_i^t = \infty$ (infinitely often participation) by invoking the  Borel-Cantelli Lemma II. 

We emphasize that our choice of $\rho$ can be made without any a priori knowledge on system sizes, number of participating clients, and statistical variation across clients. The dynamic nature of FL makes such selection rule of the hyperparameter useful and practical. This is corroborated by our experiments in Section \ref{section5} which shows that \texttt{FedADMM} performs well with constant $\rho$ in settings with different system sizes and data distributions.

In addition to the assumptions 1-2, the following conditions have been imposed when analyzing existing state-of-the-art methods. Our analysis \emph{does not require any of these conditions}.
 \\
(\textit{Data dissimilarity}) The local client loss functions satisfy: 
\begin{gather}
    \sum_{i=1}^m \norm{\nabla f_i(w)}^2\leq \norm{ \sum_{i=1}^m \nabla f_i(w)}^2 B ^2 \label{fedprox assumption}
,\end{gather}
for some constant $B$ and all $w\in \{w\,|\,\norm{\sum_{i=1}^m \nabla f_i(w)}^2>\varepsilon\}$. Note that since the accuracy $\varepsilon$ can be chosen arbitrarily small, this condition is restrictive as it would effectively force $B\to \infty$ as $\varepsilon\to 0$, unless data distributions are similar across clients. \hide{Moreover, the number of selected clients in \texttt{FedProx} has to increase as $B$ increases so as to ensure $\tfrac{B}{\sqrt{\abs{S^t}}}<1$, see also Remark 5 in \cite{Fedprox}.} In other words, in scenarios where accurate solutions are pursued and when data distributions are heterogeneous, the analysis in \cite{Fedprox} dictates that all clients need to participate in each round (which is highly undesirable in FL) when such an assumption is imposed.\\
\noindent (\textit{Bounded Gradient}) There exists a constant $G$ that uniformly bounds the client loss function gradient, i.e., 
\begin{gather}
    \norm{\nabla f_i(w_i^t)}\leq G,\,\, \forall w_i^t \in\mathbb{R}^d.\label{grad_assump}
\end{gather}
The above assumption was imposed in \cite{localsgd,Fednoniid} to analyze  \texttt{FedAvg} for bounding iterate paths during local training. However, it may be unrealistic to impose a uniform bound in the face of large populations and data heterogeneity.

\begin{table}[b]
\caption{Experimental Setup.}
\centering
\begin{tabular}{c|c|c|c}
 \hline\hline
 \textbf{Model} & \textbf{\# of para}  &  \textbf{Dataset} & \textbf{Target Accuracy} \\  
 \hline
 \multirow{2}{*}{CNN 1} & \multirow{2}{*}{1,663,370} & MNIST  & 97\%  \\ 
 \cline{3-4}
  &  & FMNIST  & 80\% \\ 
 \hline
 CNN 2  & 1,105,098 & CIFAR-10  & 45\% \\
 \hline
\end{tabular}
\label{table2}
\end{table}

\begin{table*}[t]
    \caption{Comparison of number of communication rounds along with the speedup relative to \texttt{FedSGD} to reach a target accuracy (100+ indicates the target accuracy was not reached in 100 rounds). For MNIST dataset with 100 clients, $E = 5, B = 200$ was used for both IID and non-IID data distributions. In all 1,000 client settings, we set $E=20, B=10$ for the non-IID case, and $E=20, B=\infty$ (full-batch) for the IID case. Reduction of number of communication rounds using \texttt{FedADMM} is computed over the \emph{best baseline method} in each experiment.}
	\centering
	\begin{tabular}{lllllllll}  
	\hline\hline 
     & \multicolumn{2}{l}{MNIST (100 clients)} & \multicolumn{2}{l}{MNIST (1,000 clients)} &  \multicolumn{2}{l}{FMNIST (1,000 clients)} &  \multicolumn{2}{l}{CIFAR-10 (1,000 clients)}  \\
	       & IID   & Non-IID  & IID  & Non-IID  & IID  & Non-IID  & IID  & Non-IID \\

% 	Accuracy &  $98\%$   &  $98\%$    &  $97\%$    &  $97\%$    &  $80\%$    &  $80\%$    &  $45\%$    &  $45\%$    \\
	\hline
	\texttt{FedSGD} &  297   &  250    &  201    &  269    &  390    &  530    &  186    &  202    \\
	
	\texttt{FedADMM}& 
	$\mathbf{10(29.7\times)}$   &  
	$\mathbf{33(7.6 \times)}$    &  
	$\mathbf{8(25.1\times)}$    &  
	$\mathbf{13(20.7\times)}$    &   
	$\mathbf{3(130.0\times)}$   &   
	$\mathbf{7(75.7\times)}$   &   
	$\mathbf{7(26.6\times)}$    &   
	$\mathbf{9(22.4\times)}$   \\
	
	\texttt{FedAvg} &   19($15.6\times$)   &   77($3.2\times$)    &   61($3.3\times$)    &   73($3.7\times$)    &   10($39.0\times$)    &   33($16.1\times$)    &  24($7.8\times$)   &   50($4.0\times$)   \\
	
	\texttt{FedProx} &   29($10.2\times$)   &  100+    &  78($2.6\times$) &  100+    &   14($27.9\times$)    &   61($8.7\times$)    &   32($5.8\times$)    &   68($3.0\times$)    \\
	
	\texttt{SCAFFOLD} &  27($11.0\times$)   &  76($3.3\times$)    &   61($3.3\times$)    &   84($3.2\times$)    &   12($32.5\times$)    &   40($13.3\times$)    &   37($5.0\times$)    &   100+    \\
	
	\textbf{Reduction} &  47.4 \% &  56.6\%& 86.9\% & 82.2\% &  70.0\% & 78.8\% & 70.8\% & 82.0\% \\
	
    \hline
	\end{tabular}
	\label{big_table}
\end{table*}

\section{Experiments}\label{section5}
We conducted experiments on three datasets, namely MNIST \cite{cnn1}, Fashion MNIST (FMNIST) \cite{fmnist}, and CIFAR-10 \cite{cifar}.\hide{, that constitute popular benchmarks for testing deep learning algorithms.} We compare our proposed method against \texttt{FedSGD}, \texttt{FedAvg} \cite{Fed2}, \texttt{FedProx} \cite{Fedprox}, and \texttt{SCAFFOLD} \cite{scaffold} using two popular CNN models \cite{cnn1, Fed2} with two convolutional layers each and number of parameters as detailed in Table \ref{table2}. We do not compare against  \texttt{FedHybrid} \cite{fedhybrid} and \texttt{FedPD} \cite{fedpd}, since they are both involve communication rounds that require participation of all users, which we deem unrealistic for large-scale FL scenarios. In brief, our experiments unravel three main findings:
(i) \texttt{FedADMM} achieves high accuracy in both IID and non-IID settings within \emph{substantially fewer rounds}, which translates to large communication\hide{and computation} savings; (ii) \texttt{FedADMM} features \emph{robust convergence} against both statistical and system heterogeneity, and requires no hyperparameter tuning across the IID and non-IID settings. This is in contrast to \texttt{FedProx} whose proximal coefficient needs to be carefully chosen according to data dissimilarity, which is difficult to assess in a practical scenario; (iii) \texttt{FedADMM} outperforms all baseline methods in \emph{all test scenarios}, with the advantages being most pronounced in large-scale systems with heterogeneous data distributions. This is as opposed to other methods, where an improved performance is restricted to either the IID or the non-IID setting.

\subsection{Experimental Setup}
We study ten-class image classification on three real datasets with details specified below. Two CNN models are used for our experiments, both with a convolutional module (two $5\times5$ convolutional layers, each followed by $2\times2$ max pooling layers), and a fully connected layer module. The input of the two models is a flattened image with dimension $784$ and $3,072$, respectively, while the output for both is a class label from $0$ to $9$.
Table \ref{table2} summarizes the datasets,  model sizes, and target accuracies used in our experiments. We denote $E$ as the local epoch number, $B$ as the local batch size, and $C$ as the fraction of clients that participate in the training process during each round. All experiments are implemented in PyTorch on a system with 2 Intel$^\circledR$ Xeon$^\circledR$ Silver 4210 CPUs and 2 NVIDIA$^\circledR$ Tesla$^\circledR$ V100S GPUs. Clients are selected uniformly at random during each global round, and the number of participating clients is set to $10\%$ ($C=0.1$) of total clients in all cases, while SGD was chosen as the local solver in all cases. This was done for the sake of comparison with the baselines, but our framework allows for arbitrary client participation and local solver. The system heterogeneity (i.e., variable computational capabilities across clients) is captured by letting each client select the local epoch number uniformly between $1$ and $E$ in \texttt{FedADMM} as well as in \texttt{FedProx}. The number of local epochs for \texttt{FedAvg} and \texttt{SCAFFOLD} are fixed to be $E$; this was done in order to compare against baselines in their principal description. We adopt random initialization for the global model in all algorithms, zero initialization for dual variables in \texttt{FedADMM}, and zero initialization for \textit{control variates} in \texttt{SCAFFOLD} (as recommended). Recall that the use of \textit{control variates} doubles the upload cost of \texttt{SCAFFOLD} compared to other baselines. We set the server gathering step size $\eta$ to be 1 unless otherwise specified and present all results averaged over five runs. The code is available at: github.com/YonghaiGong/FedADMM.

\begin{figure*}[t]
  \centering
    \begin{subfigure}[b]{0.245\linewidth}
    \includegraphics[width=\textwidth]{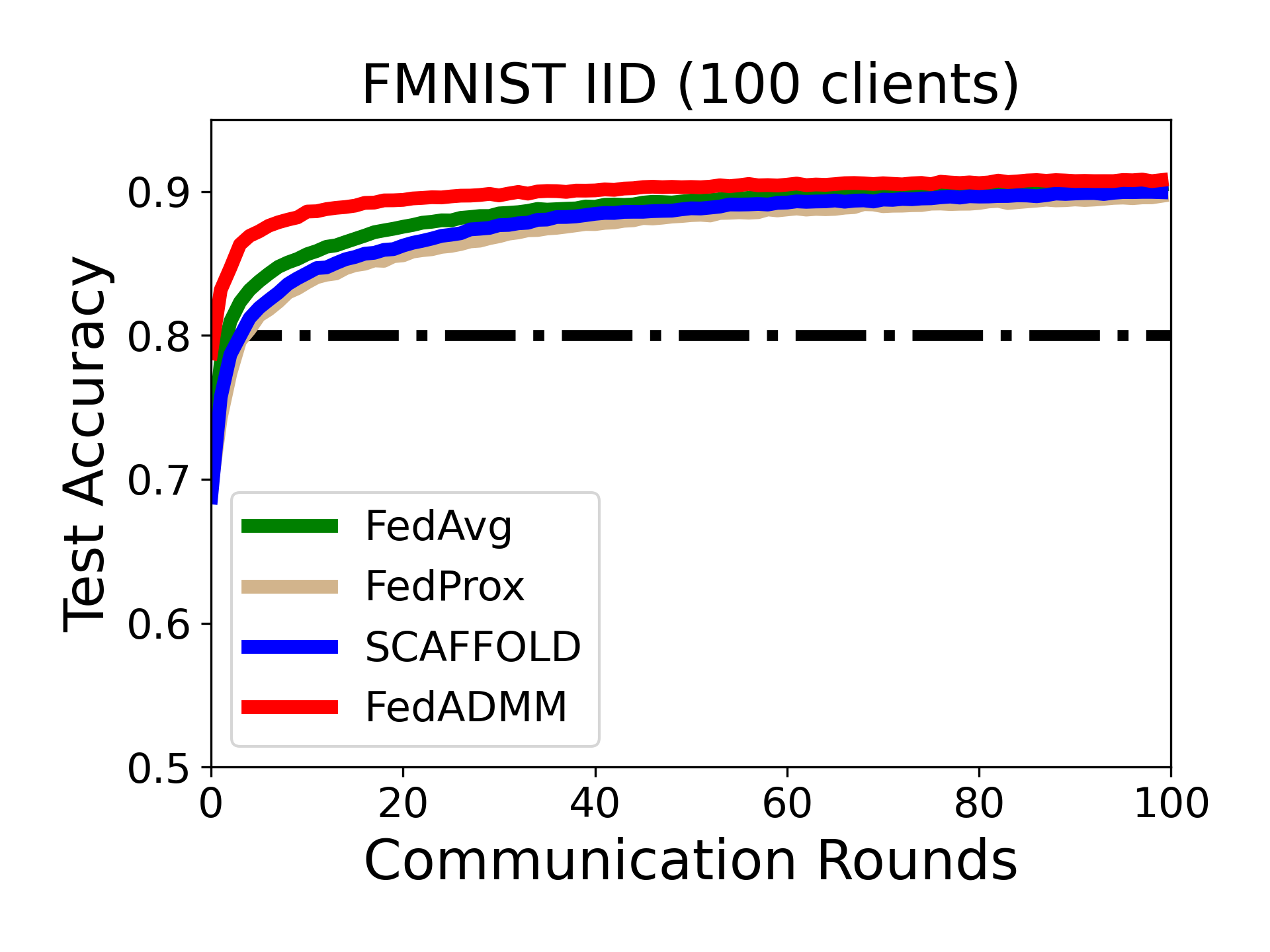}
  \end{subfigure}
  \begin{subfigure}[b]{0.245\linewidth}
    \includegraphics[width=\textwidth]{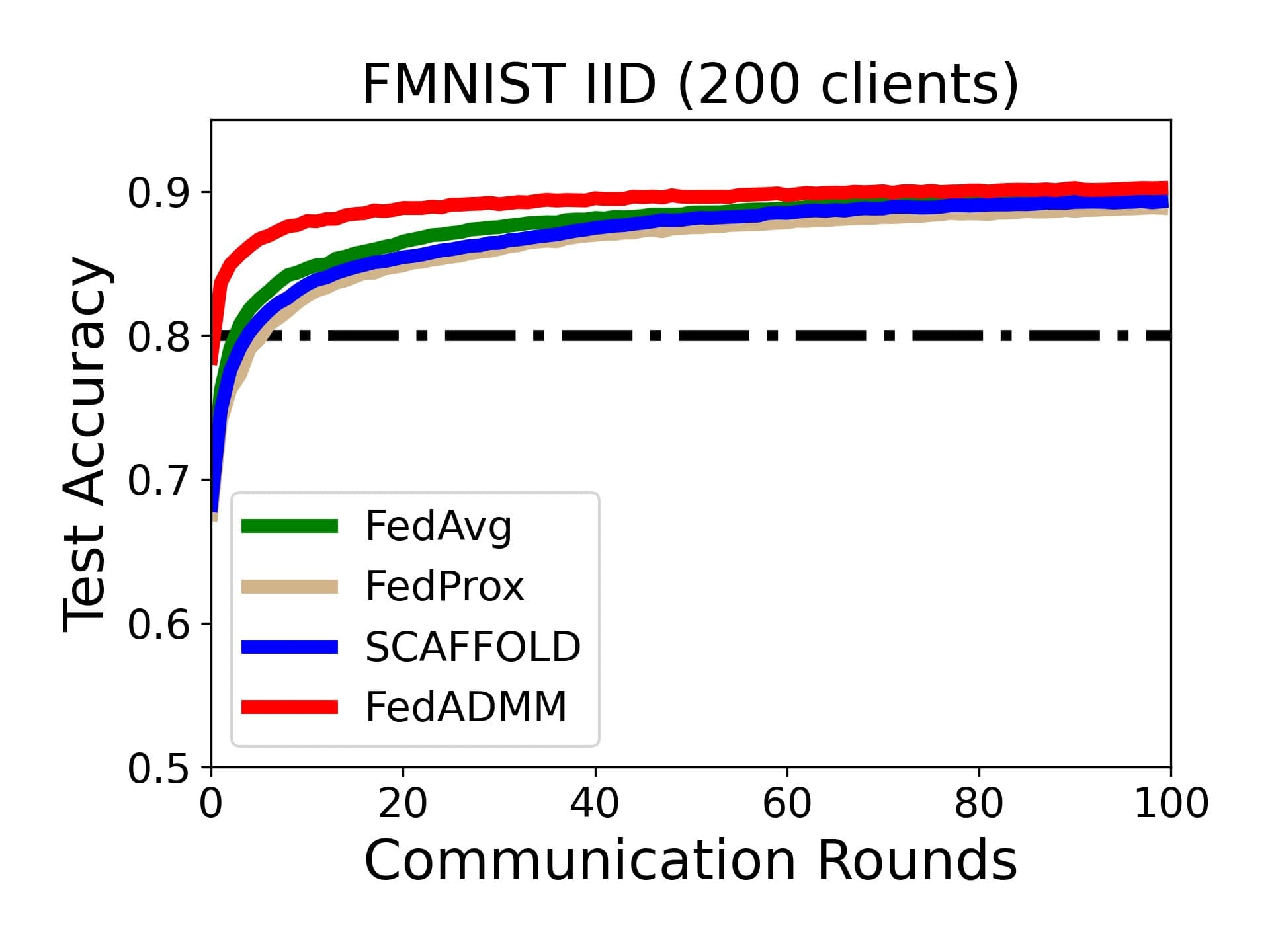}
  \end{subfigure}
    \begin{subfigure}[b]{0.245\linewidth}
    \includegraphics[width=\textwidth]{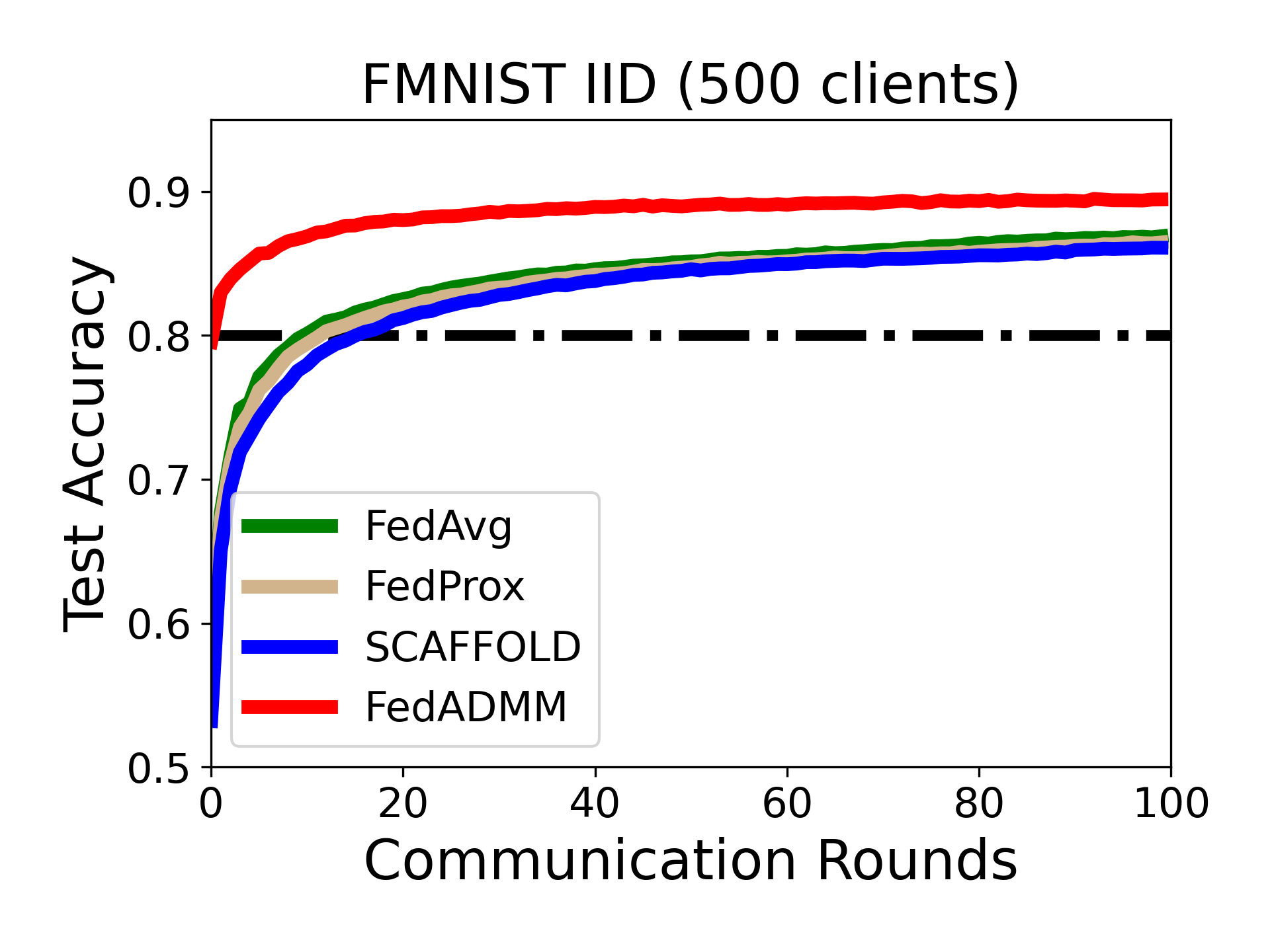}
  \end{subfigure}
      \begin{subfigure}[b]{0.245\linewidth}
    \includegraphics[width=\textwidth]{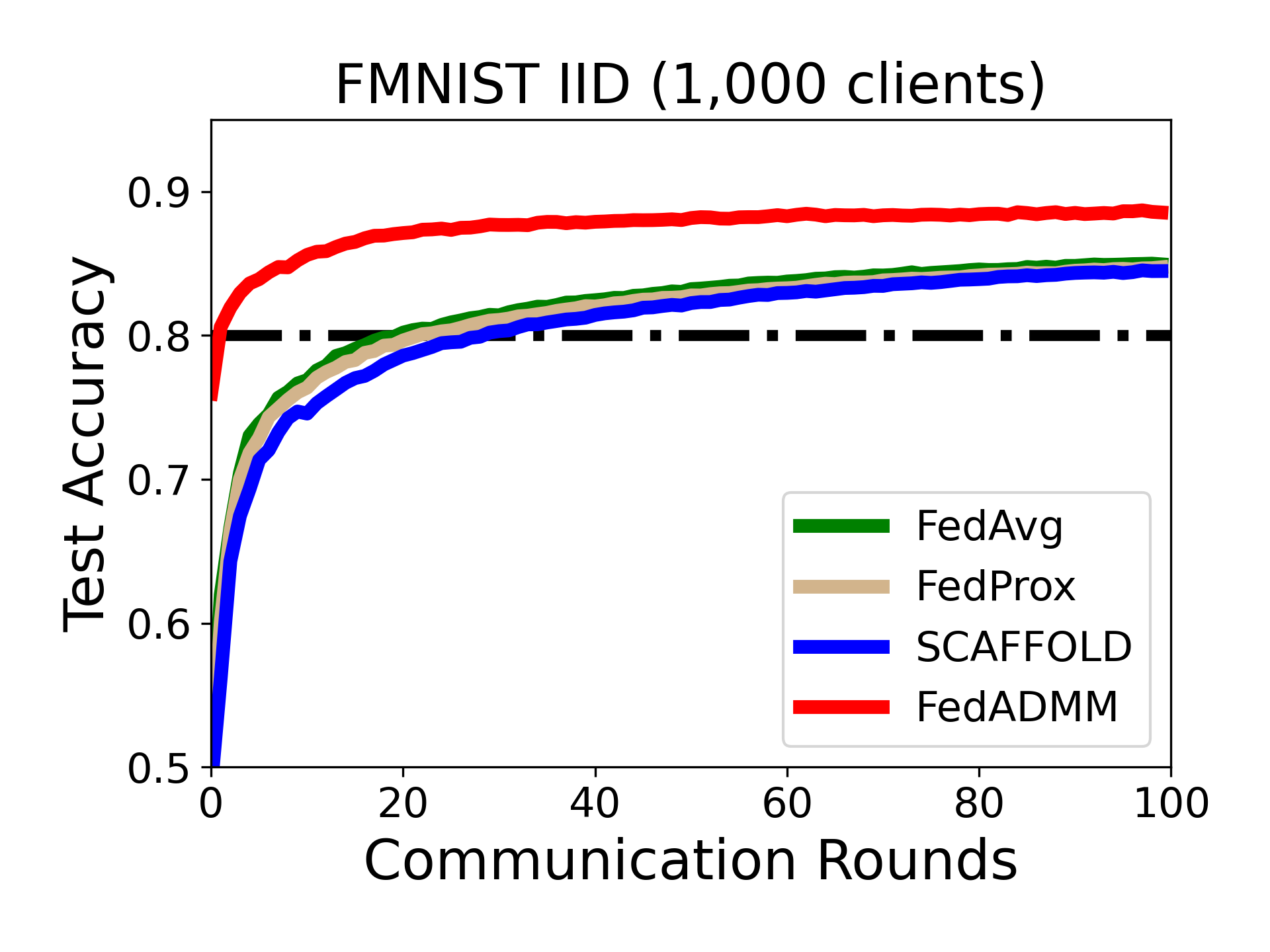}
  \end{subfigure}
      \begin{subfigure}[b]{0.245\linewidth}
    \includegraphics[width=\textwidth]{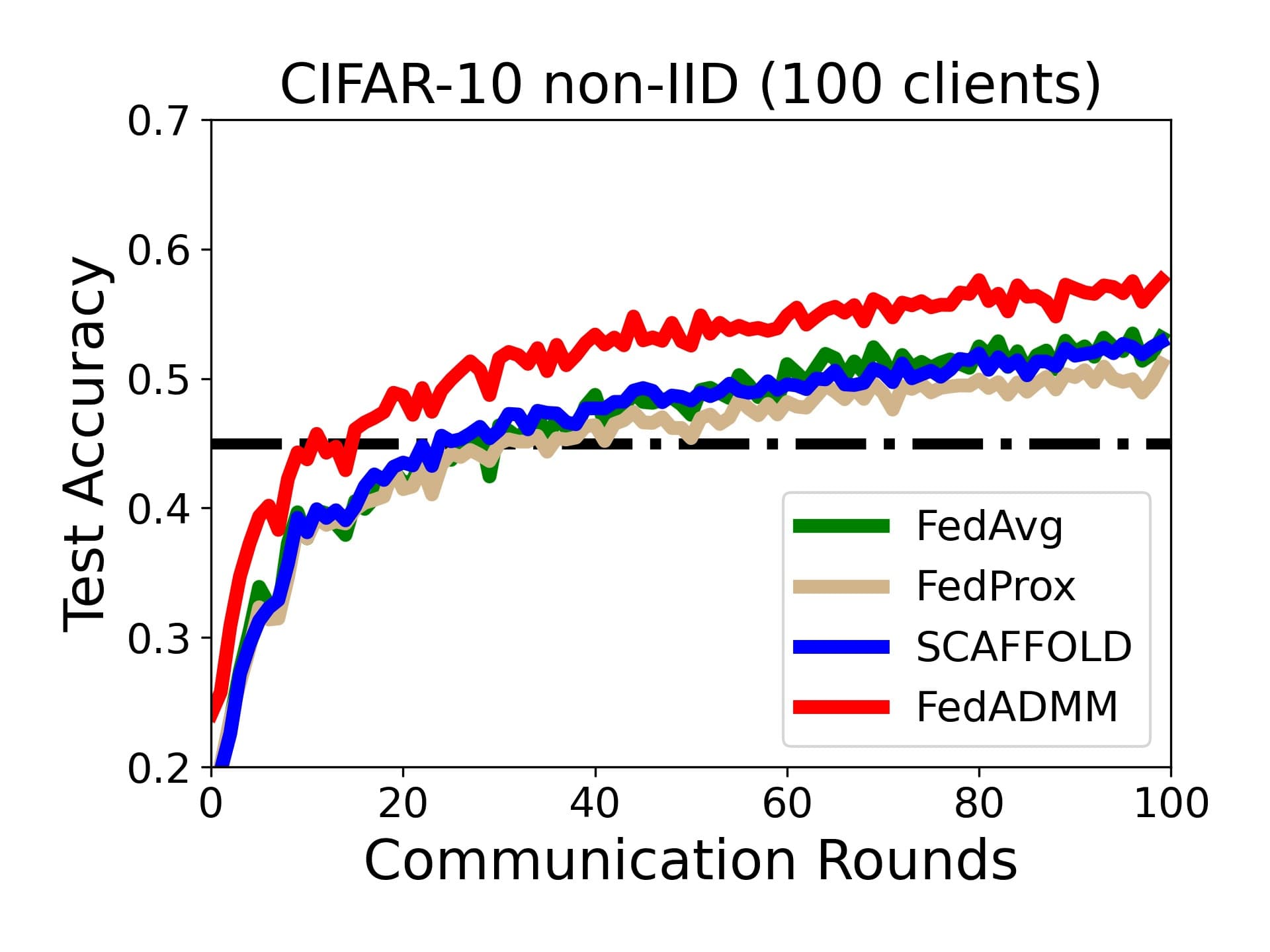}
  \end{subfigure}
  \begin{subfigure}[b]{0.245\linewidth}
    \includegraphics[width=\textwidth]{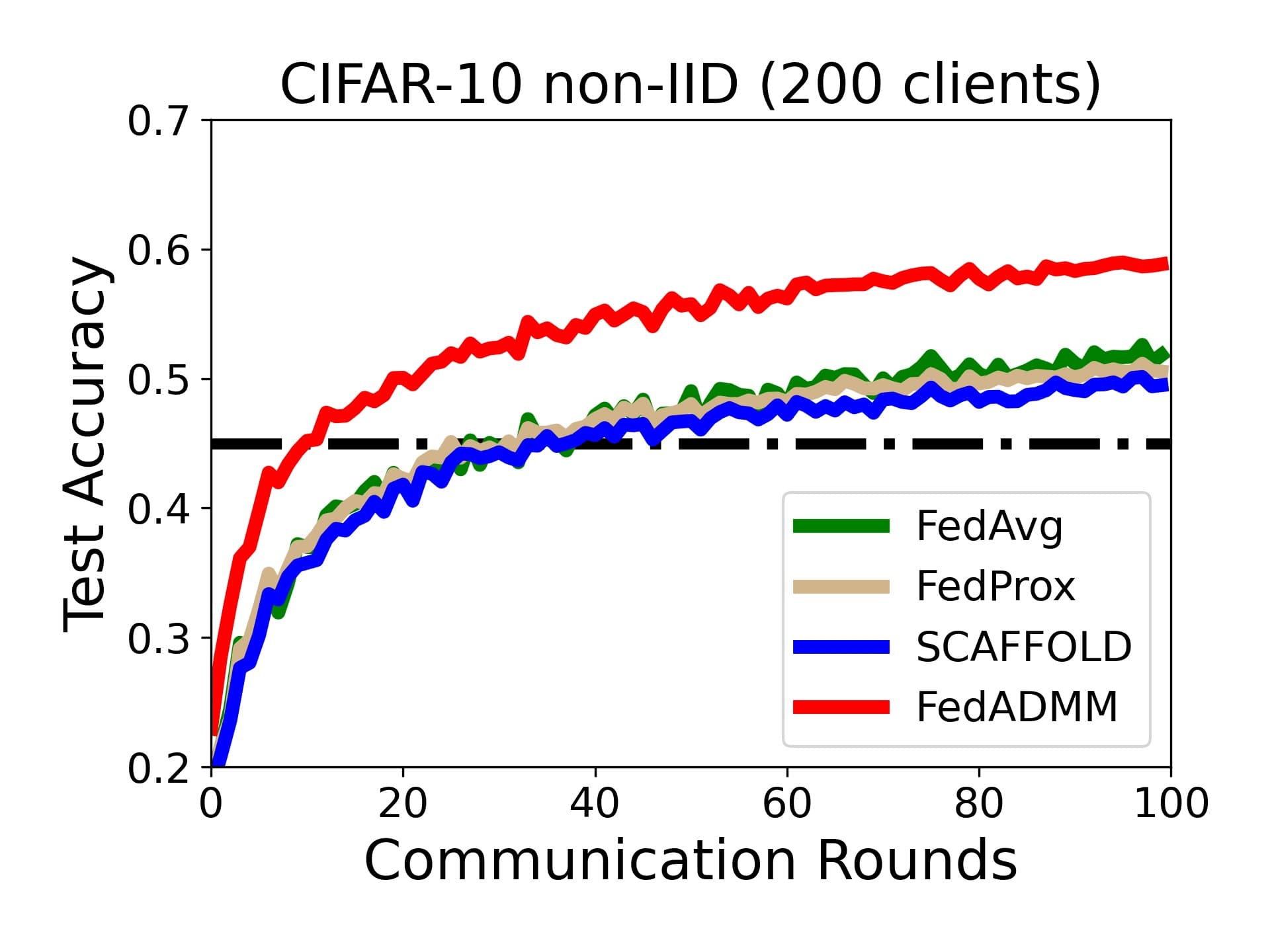}
  \end{subfigure}
    \begin{subfigure}[b]{0.245\linewidth}
    \includegraphics[width=\textwidth]{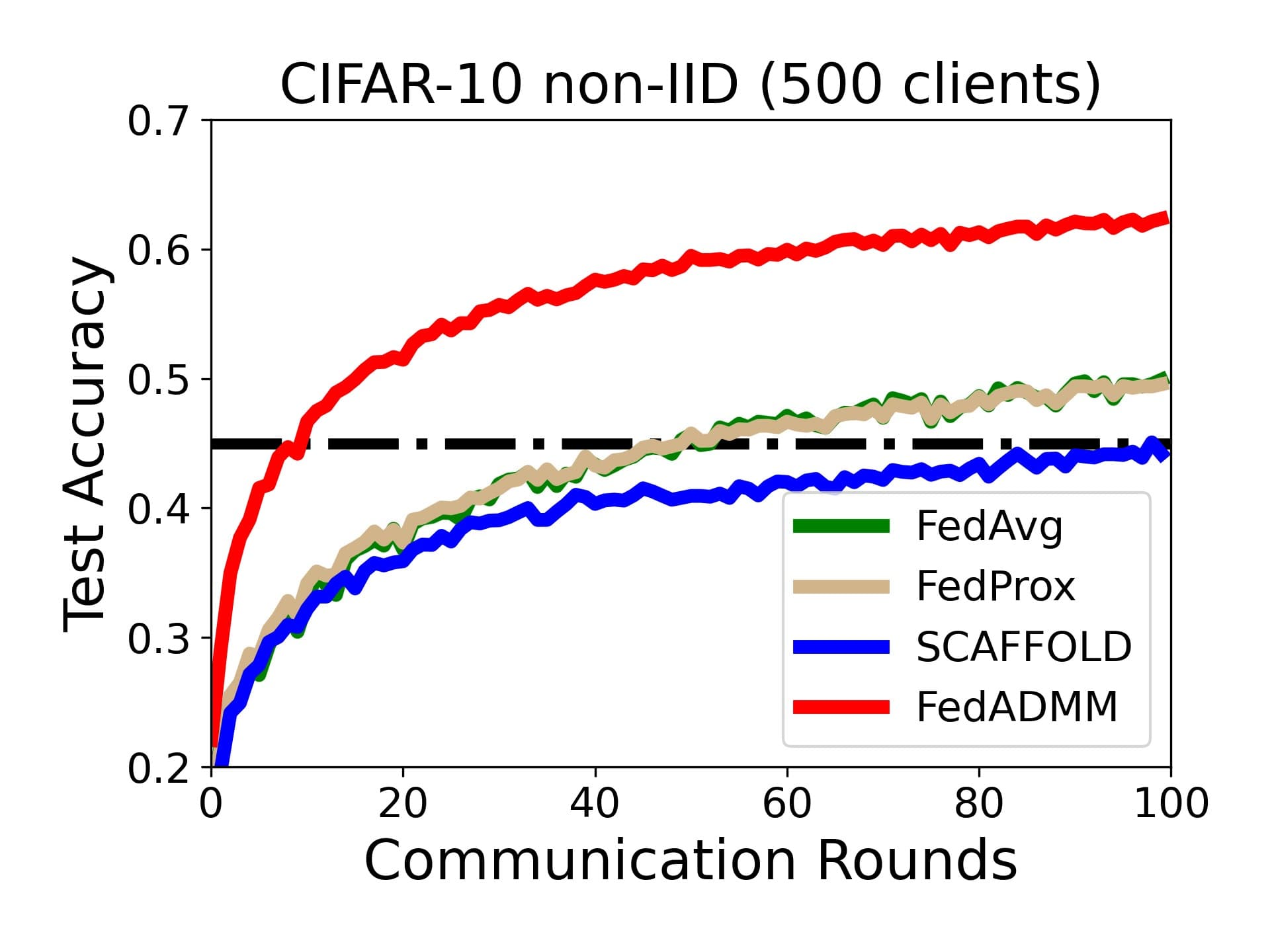}
  \end{subfigure}
      \begin{subfigure}[b]{0.245\linewidth}
    \includegraphics[width=\textwidth]{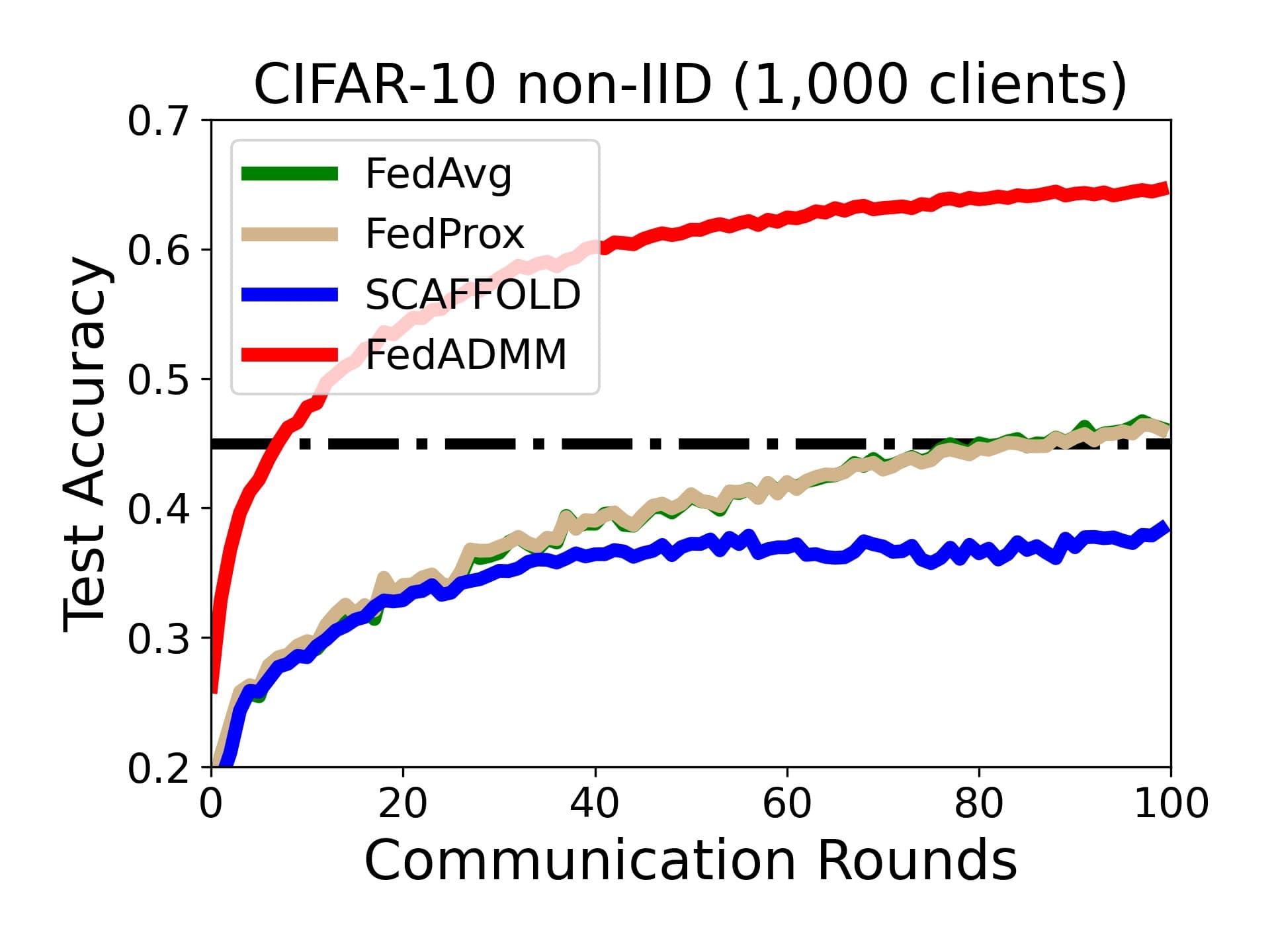}
  \end{subfigure}
  \caption{Convergence paths in systems with variable number of users (prescribed accuracies of $80\%$ for FMNIST and $45\%$ for CIFAR-10 are plotted as black dashed lines). We test how performance varies with fixed hyperparameters (tuned for best performance in the 100-client setting) for different system scales. We conclude that the performance gap of \texttt{FedADMM} increases with the client population.}
  \label{scaling_figure}
\end{figure*}

\begin{figure}[t]
  \centering
    \begin{subfigure}[b]{0.49\linewidth}
    \includegraphics[width=\textwidth]{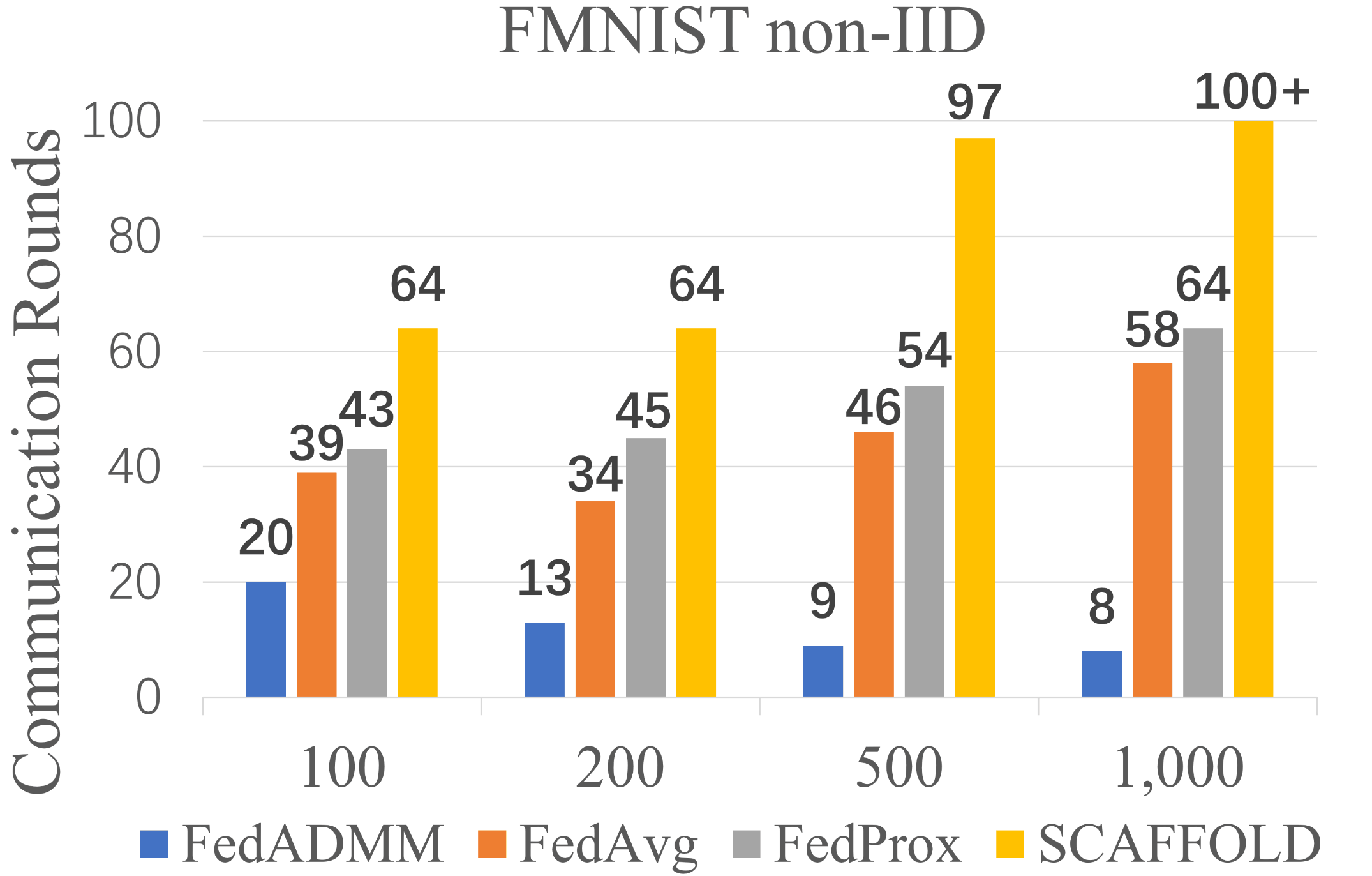}
  \end{subfigure}
  \begin{subfigure}[b]{0.49\linewidth}
    \includegraphics[width=\textwidth]{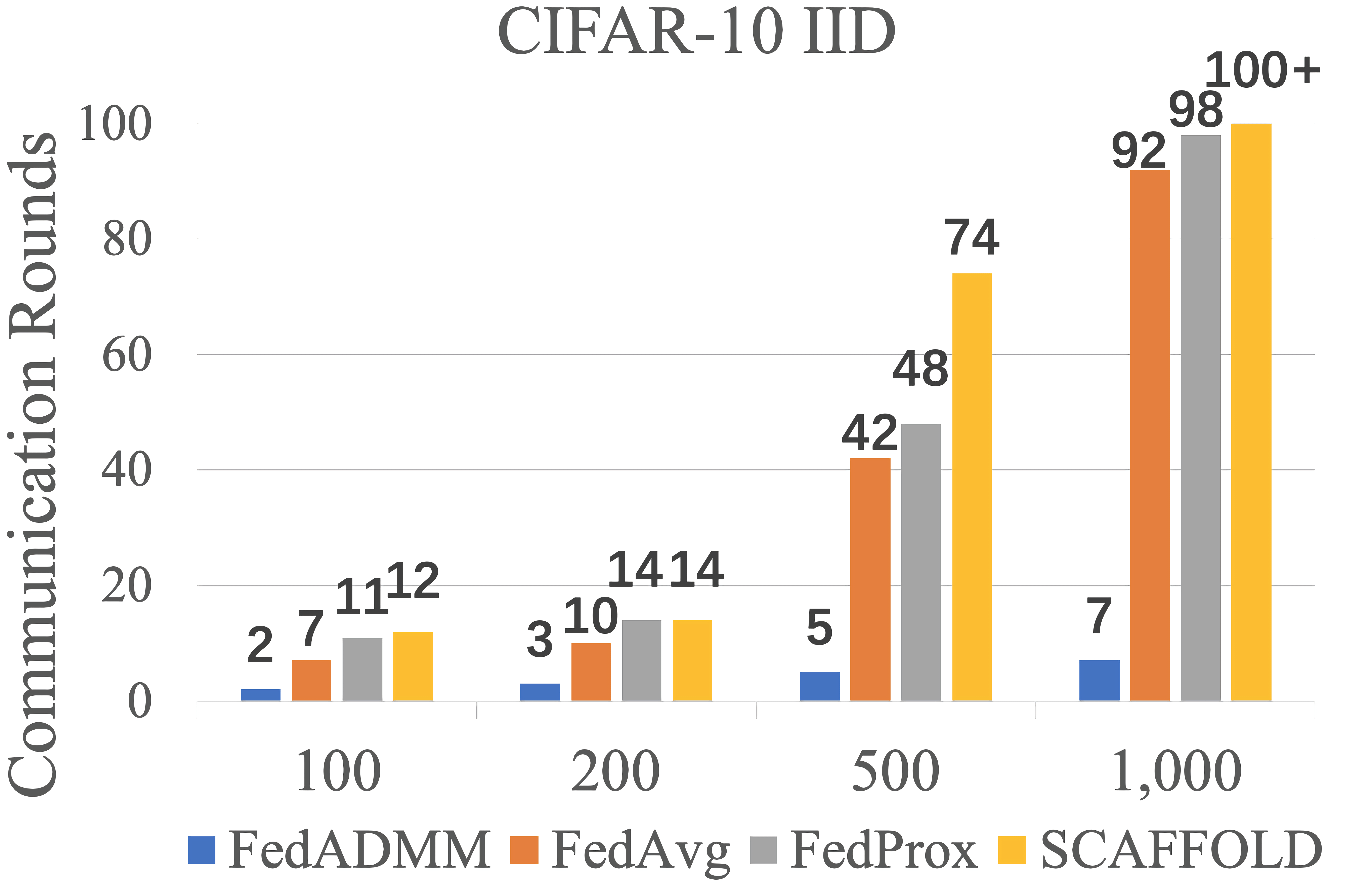}
  \end{subfigure}
    \begin{subfigure}[b]{0.49\linewidth}
    \includegraphics[width=\textwidth]{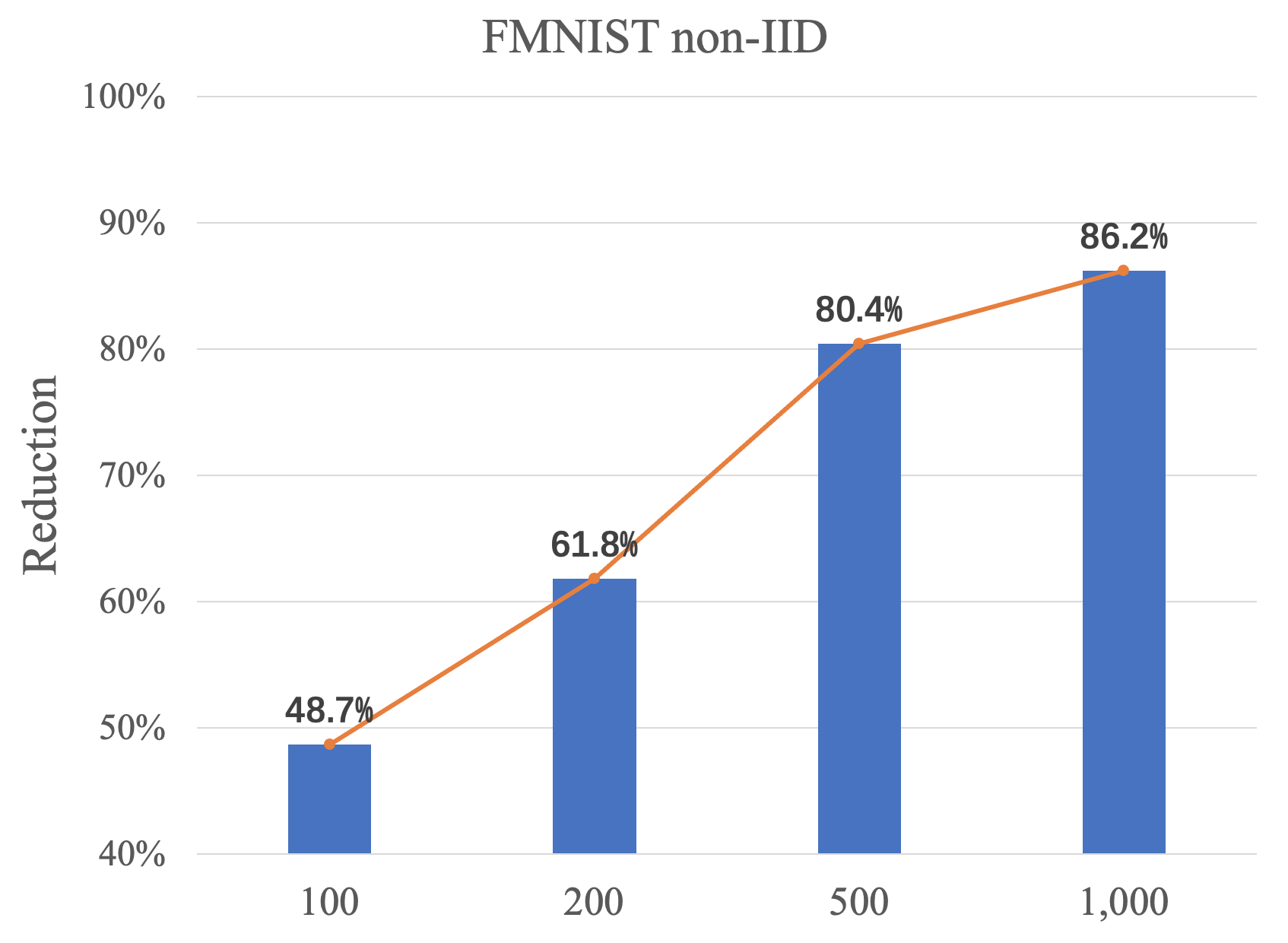}
  \end{subfigure}
      \begin{subfigure}[b]{0.49\linewidth}
    \includegraphics[width=\textwidth]{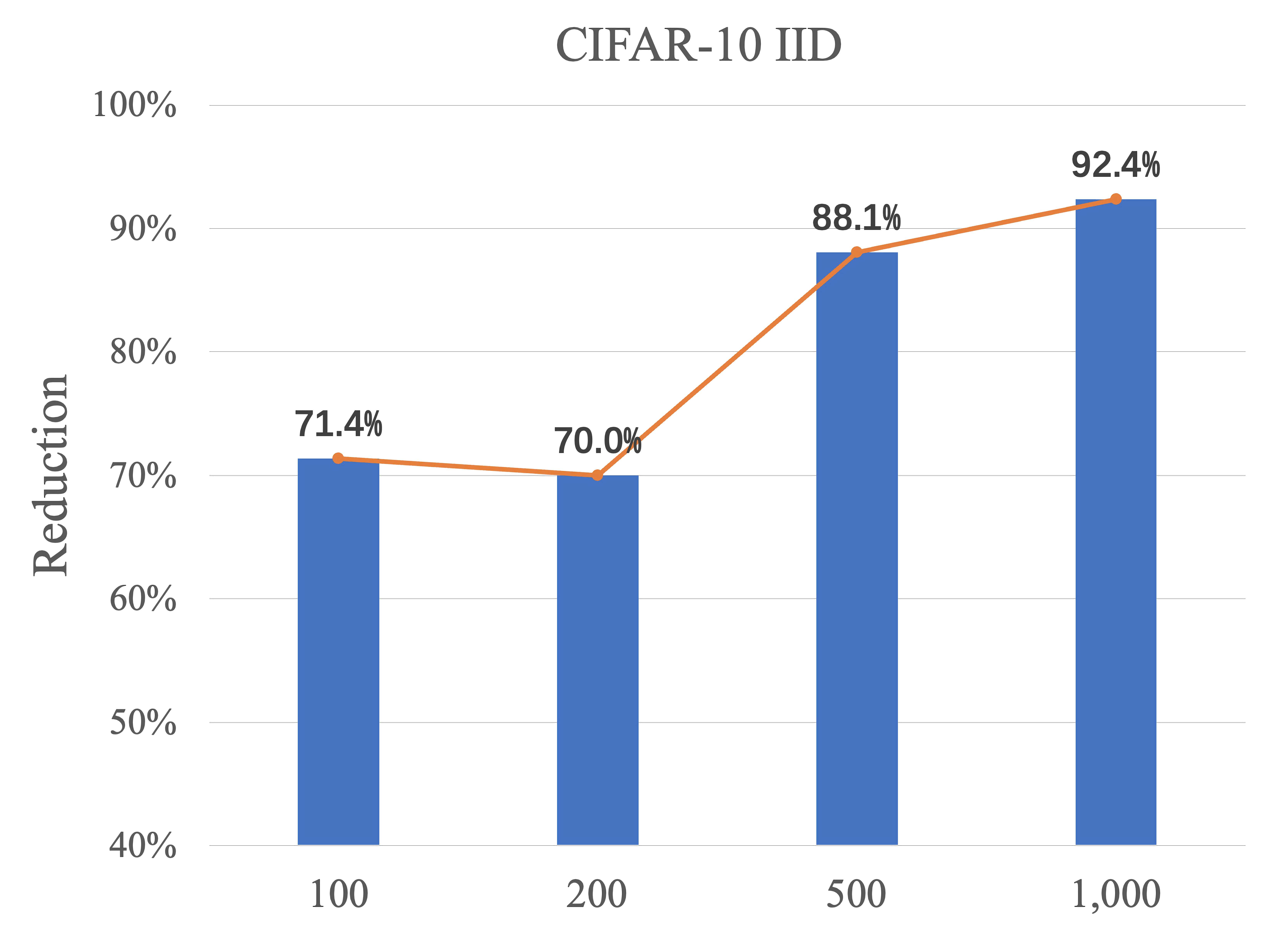}
  \end{subfigure}
  \caption{Number of rounds to reach a prescribed accuracy for variable client population along with the reduction of \texttt{FedADMM} over the best baseline method.}
  \label{figure8}
\end{figure}

\textbf{Data Distribution.} For the IID setting, data are evenly distributed to clients. In contrast, to account for data heterogeneity (non-IID setting), we first arrange the training data by label and then distribute them evenly into shards: each client is assigned two shards uniformly at random. We note that this is
a rather extreme representative of data heterogeneity, since clients tend to overfit to the specific classes of assigned data. \\
%   To account for data heterogeneity (non-IID setting), we first arranged the training data by label and then divided them into shards of equal size. Each client is assigned two shards, i.e., it typically contains no more than two classes of training data. In contrast, for the IID setting we uniformly assigned data to each client from the dataset. \\
\indent \textbf{Hyperparameters.} The learning rates of the local SGD solver for all algorithms are selected from the candidate set $\{0.01, 0.1, 0.2, 0.5 \}$ for best performance. Recall that both \texttt{FedADMM} and \texttt{FedProx} make use of a quadratic penalty term and require the setting of hyperparameter $\rho$. It has been observed in \cite{Fedprox}, \cite{scaffold}, and is further supported in our experiments, that careful tuning of $\rho$ is required to achieve the best performance for \texttt{FedProx} (see discussion on \textbf{Proximal Parameter} $\rho$ in Section V-B). For fairness, we fixed $\rho=0.1$ for \texttt{FedProx} except for Table \ref{big_table}, where $\rho$ is selected from the candidate set $\{0.001, 0.01, 0.1, 1 \}$ to test for its best performance. In contrast, \texttt{FedADMM} outperforms all baselines with fixed $\rho=0.01$ (empirically fixed) in all experiments. This is in full alignment with our theoretical analysis (see \textbf{Theorem 1} and \textbf{Remark 1}) that suggests that $\rho$ can be set as constant, independent of system size and data distributions.

\begin{figure}[t]
%   \centering
    \begin{subfigure}[b]{0.49\linewidth}
    \includegraphics[width=\textwidth]{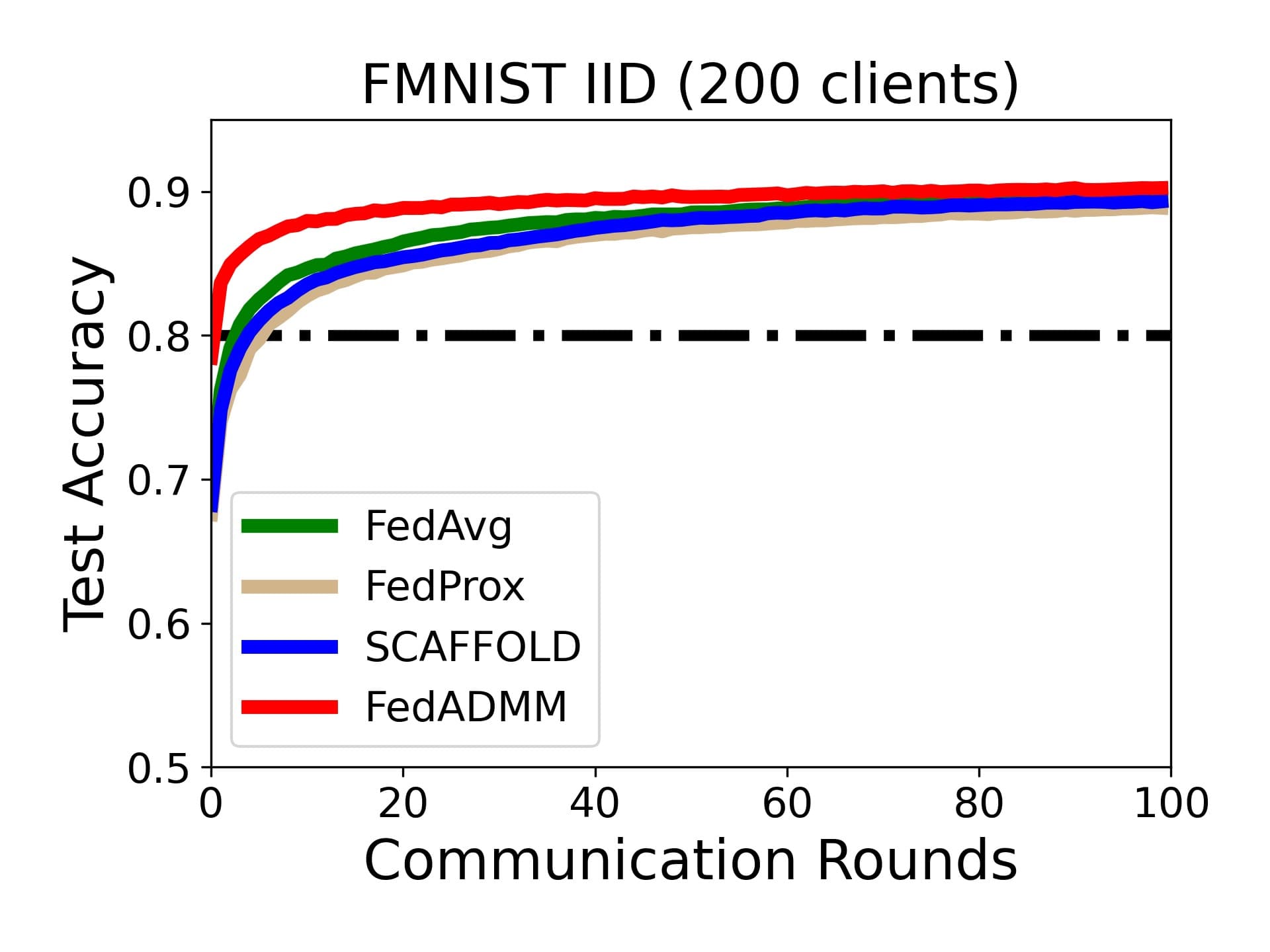}
  \end{subfigure}
  \begin{subfigure}[b]{0.49\linewidth}
    \includegraphics[width=\textwidth]{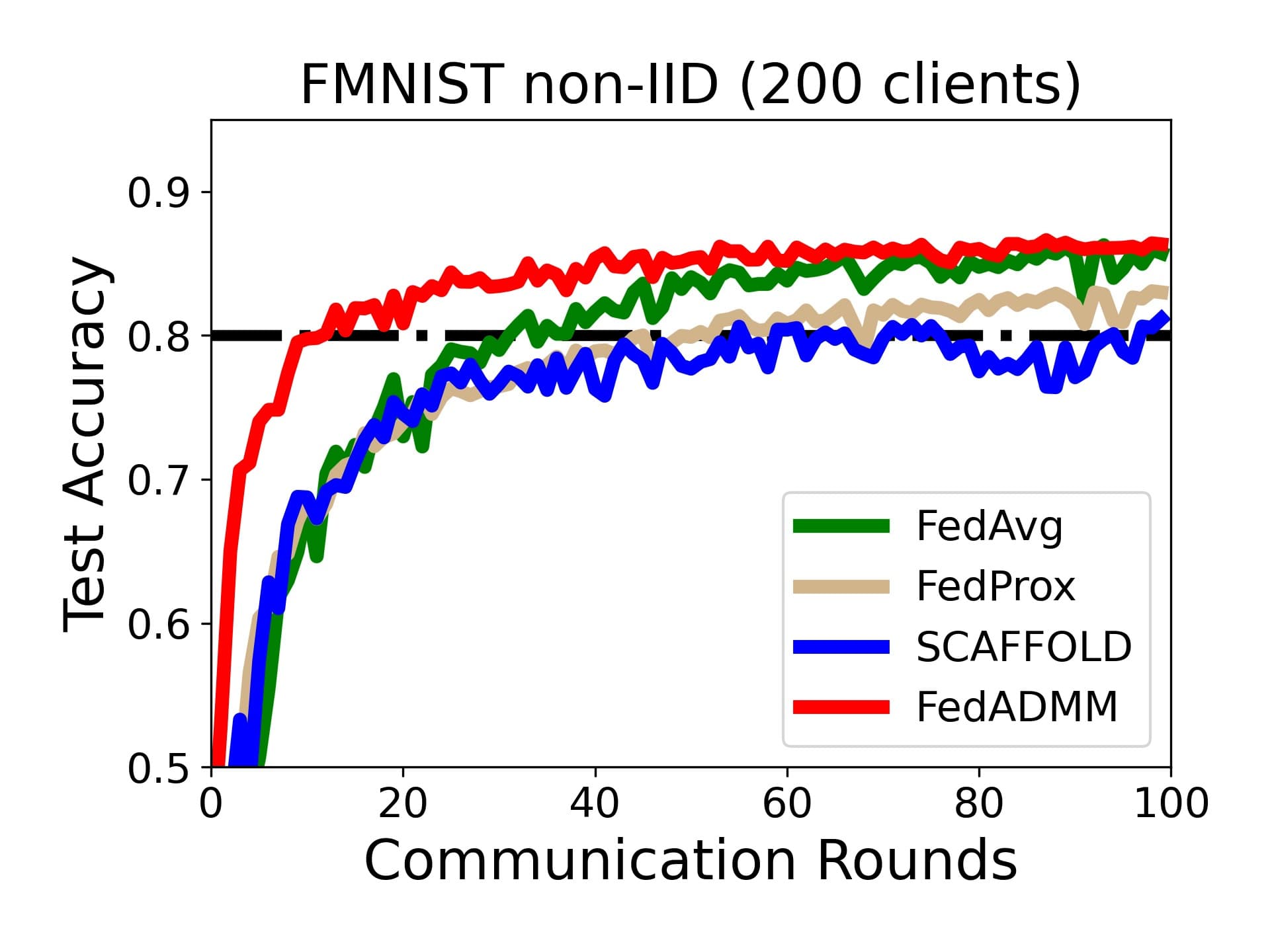}
  \end{subfigure}
    \begin{subfigure}[b]{0.49\linewidth}
    \includegraphics[width=\textwidth]{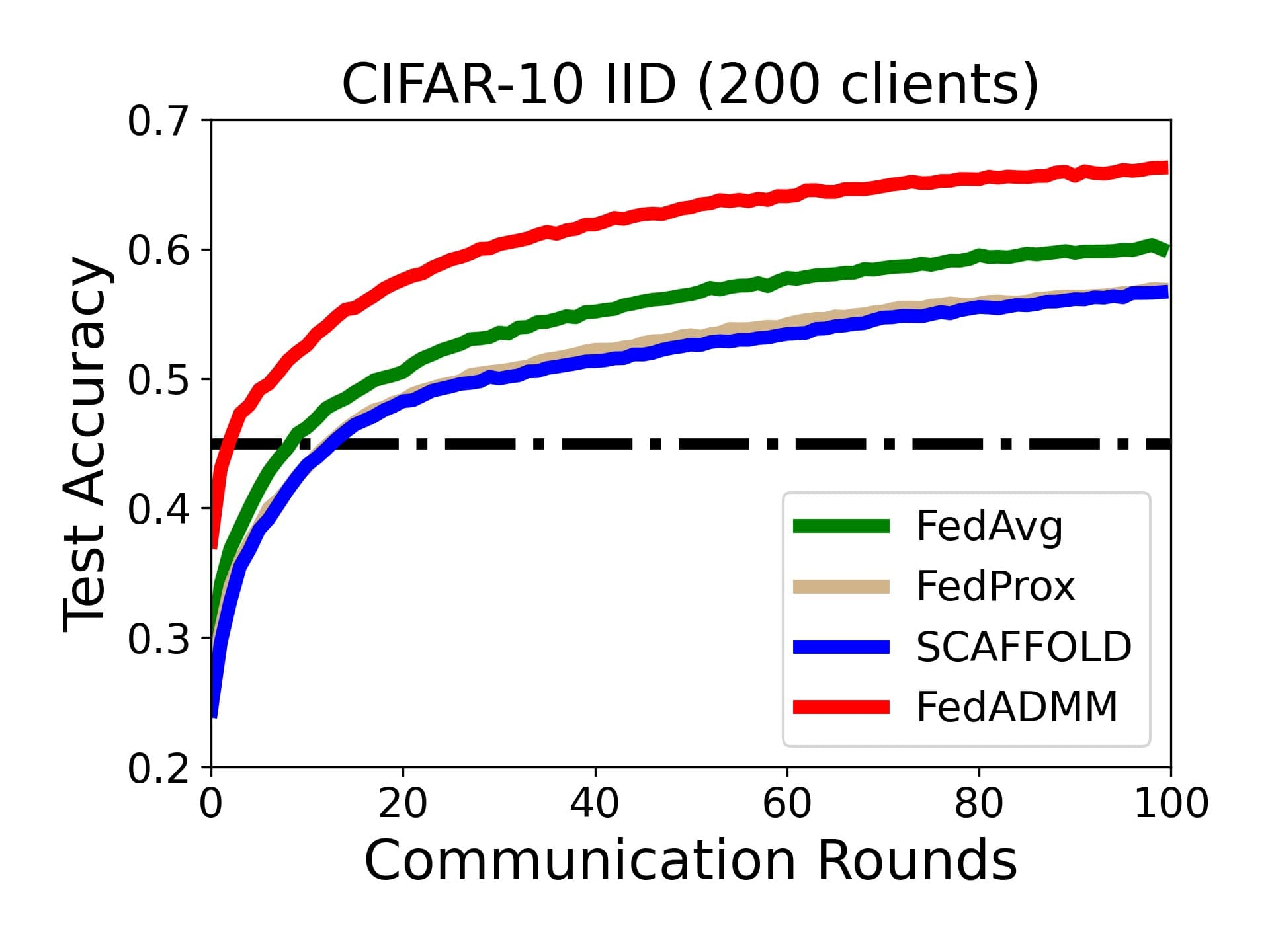}
  \end{subfigure}
    \begin{subfigure}[b]{0.49\linewidth}
    \includegraphics[width=\textwidth]{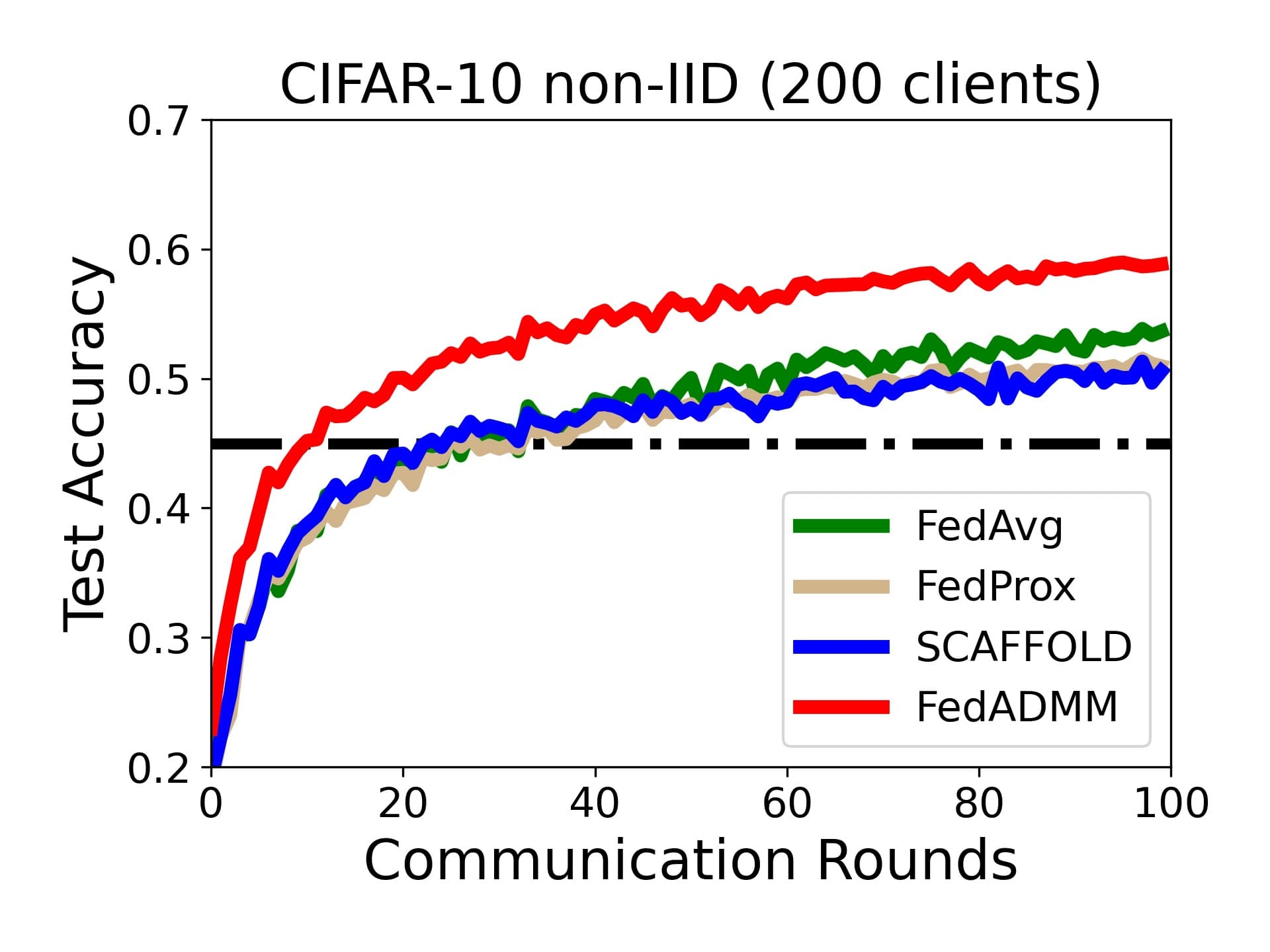}
  \end{subfigure}
  \caption{Comparison of adaptability to heterogeneous data. Target accuracy is $80\%$ for FMNIST and $45\%$ for CIFAR-10. For both cases, $m=200$, $E=10$, $B=50$. In \texttt{FedADMM}, we fix the local learning rates to $0.1$ and $\rho=0.01$, while other algorithms are tuned for best performance. In all scenarios, \texttt{FedADMM} consistently outperforms all baseline methods even with no hyperparameter tuning.}
  \label{adaption_figure}
\end{figure}

\begin{figure}[t]
  \centering
    \begin{subfigure}[b]{0.49\linewidth}
    \includegraphics[width=\textwidth]{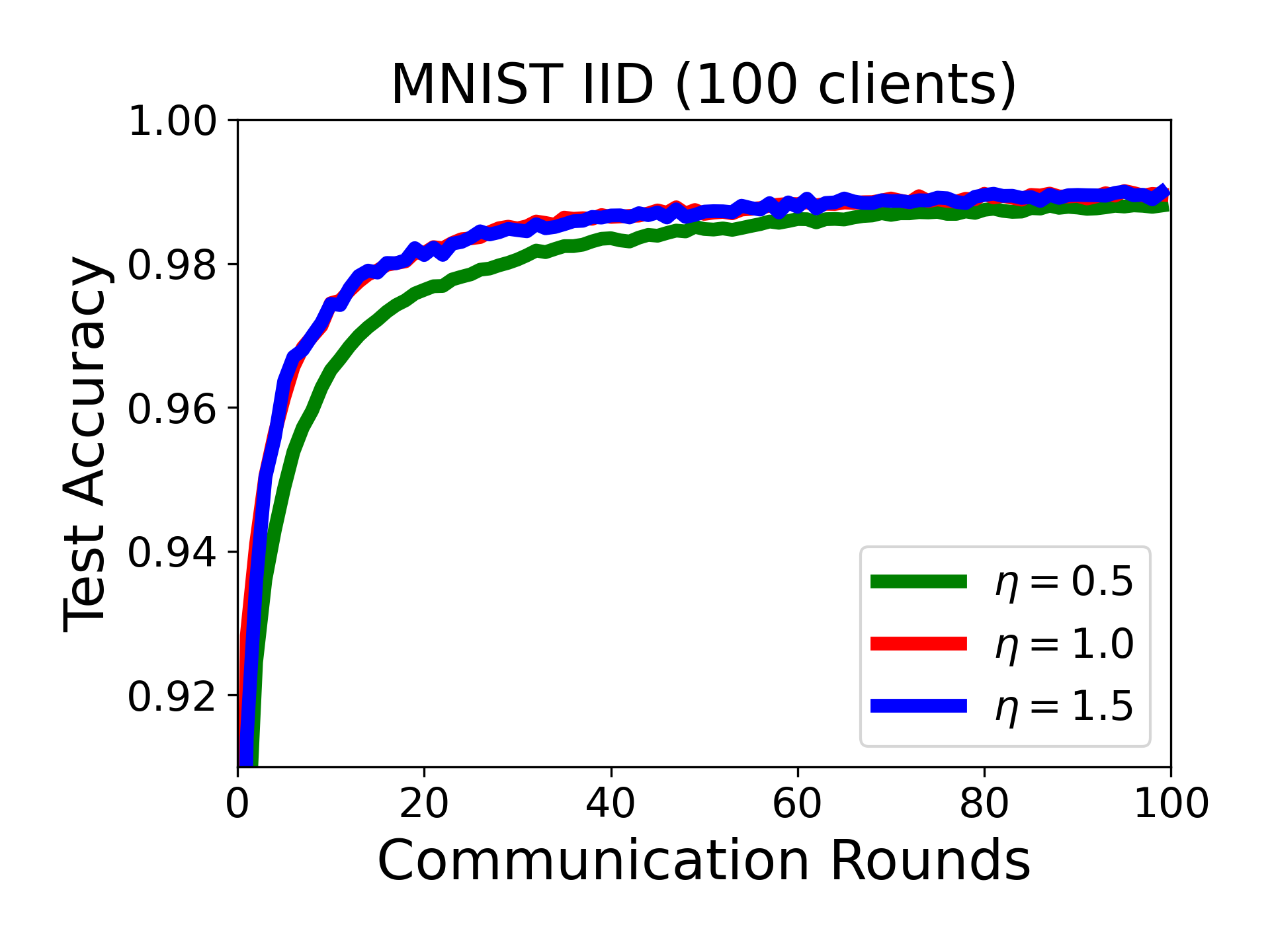}
  \end{subfigure}
  \begin{subfigure}[b]{0.49\linewidth}
    \includegraphics[width=\textwidth]{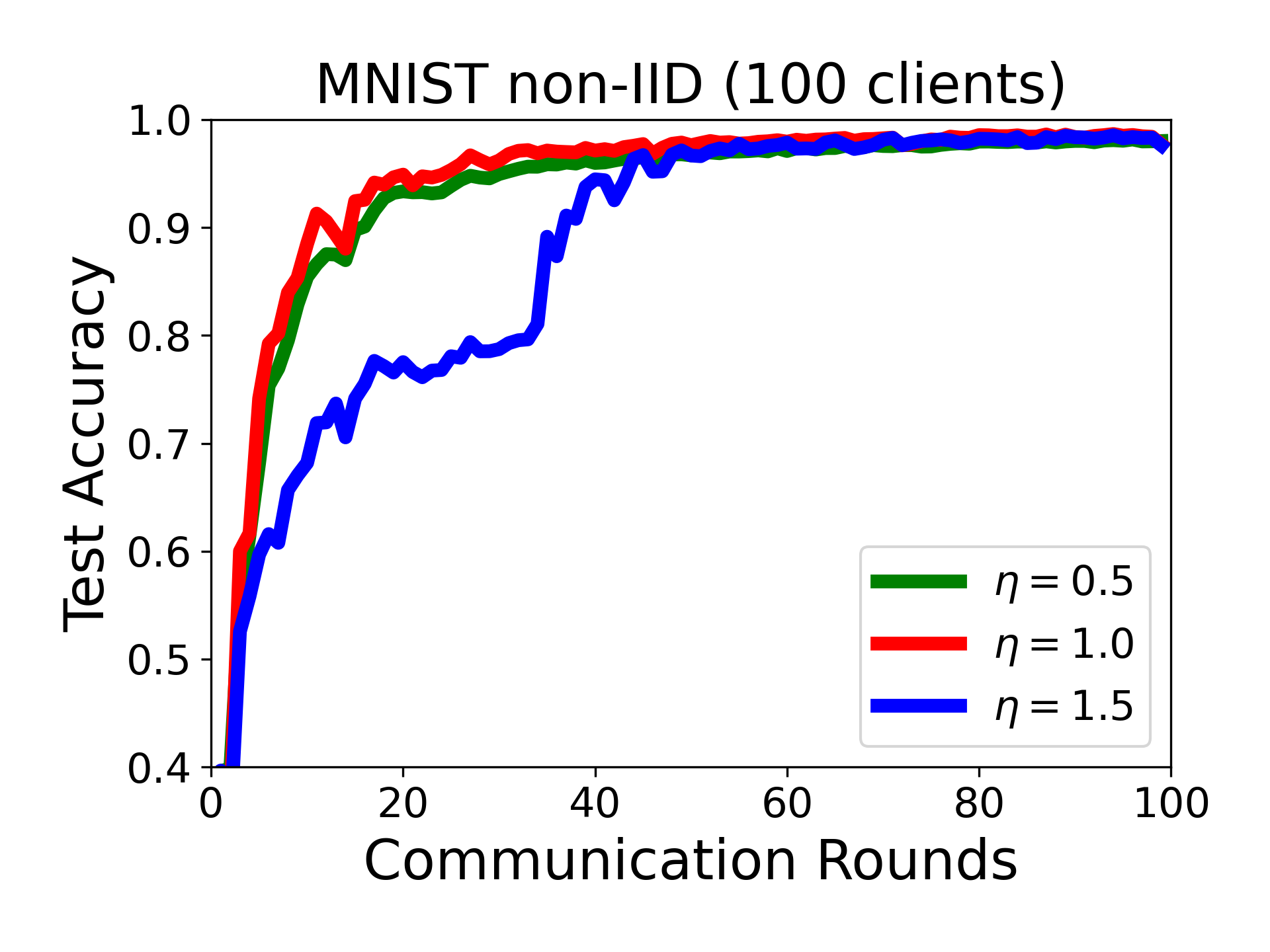}
  \end{subfigure}
      \begin{subfigure}[b]{0.49\linewidth}
    \includegraphics[width=\textwidth]{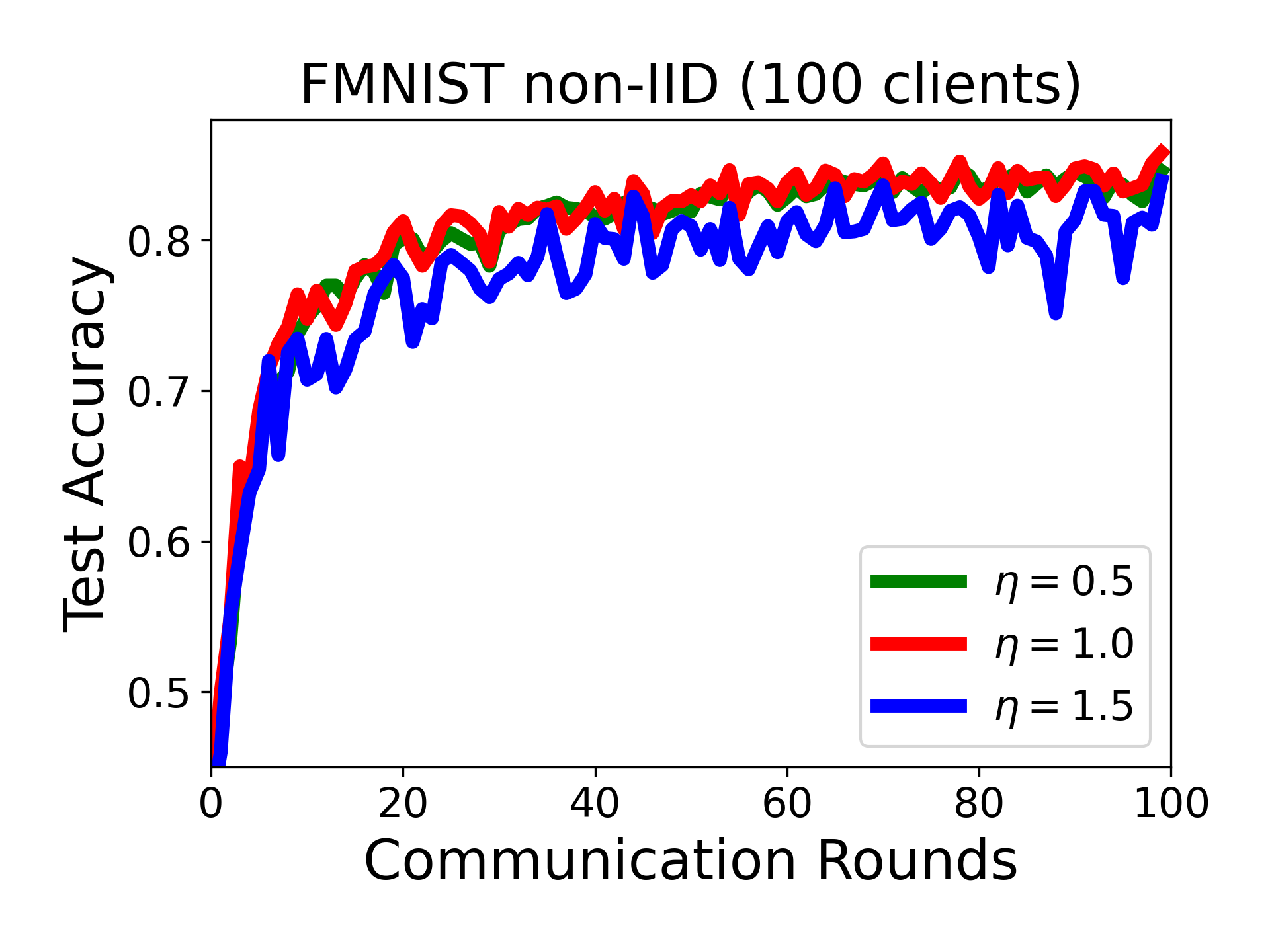}
  \end{subfigure}
  \begin{subfigure}[b]{0.49\linewidth}
    \includegraphics[width=\textwidth]{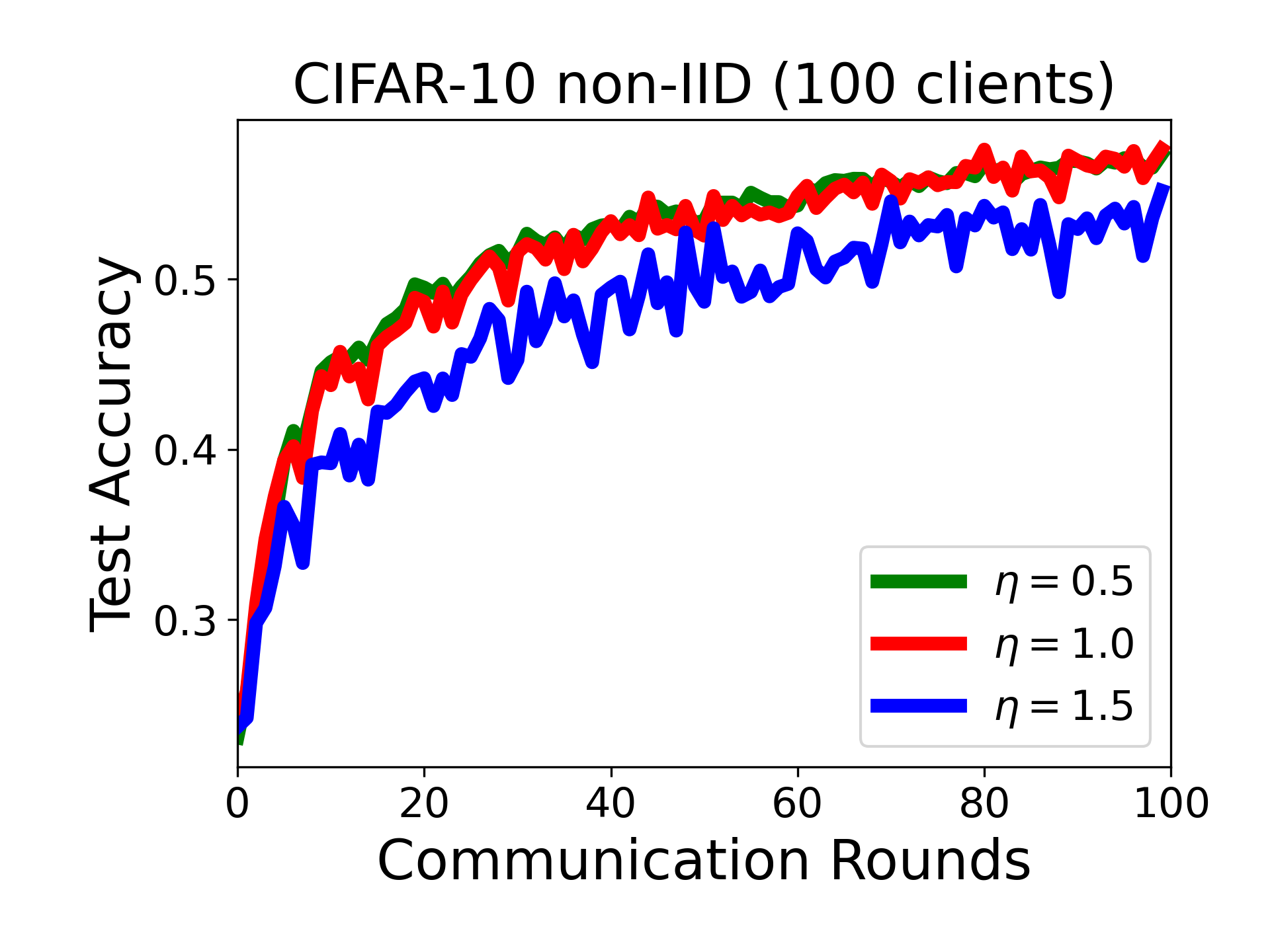}
  \end{subfigure}
    \begin{subfigure}[b]{0.49\linewidth}
    \includegraphics[width=\textwidth]{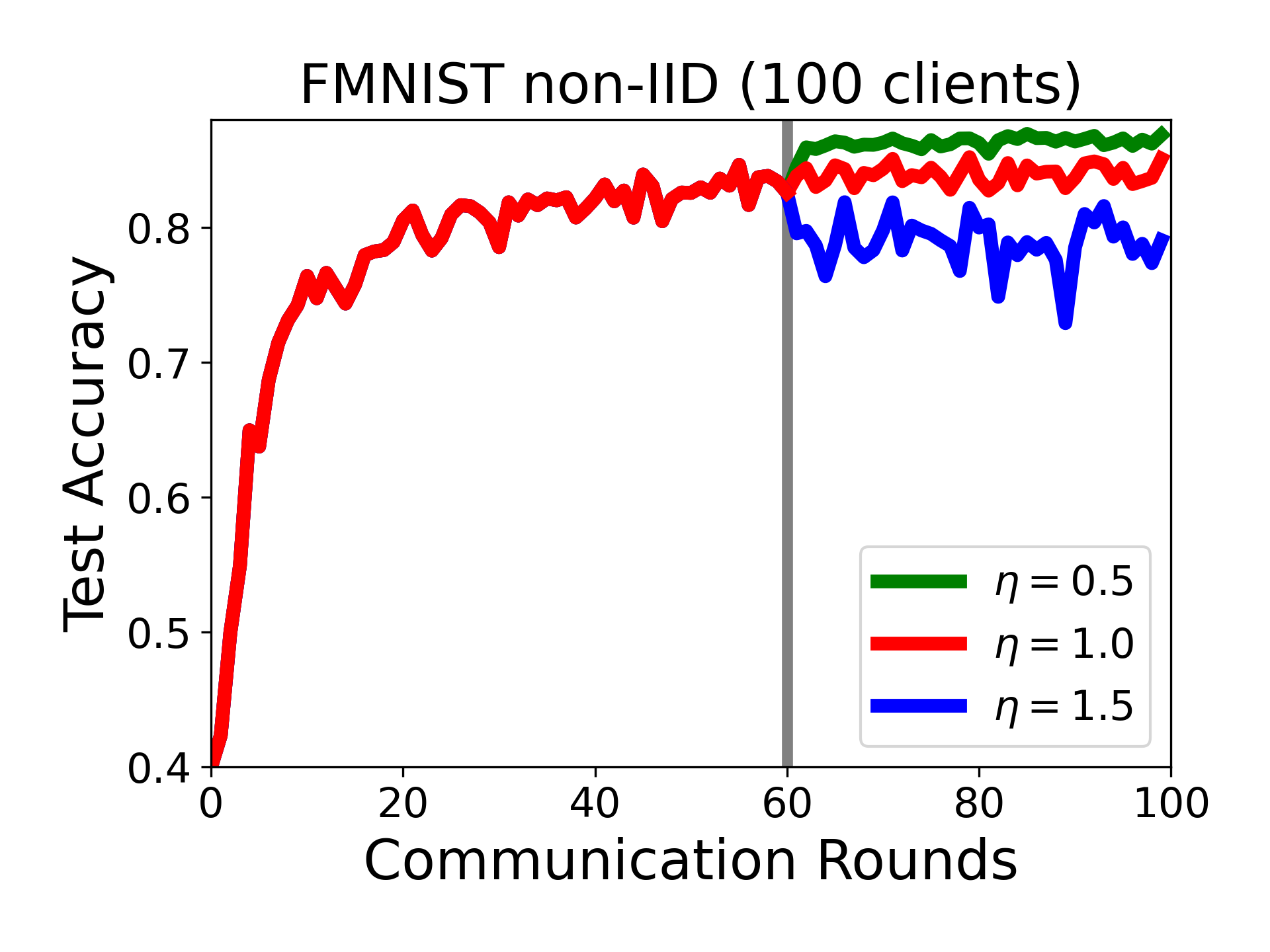}
  \end{subfigure}
      \begin{subfigure}[b]{0.49\linewidth}
    \includegraphics[width=\textwidth]{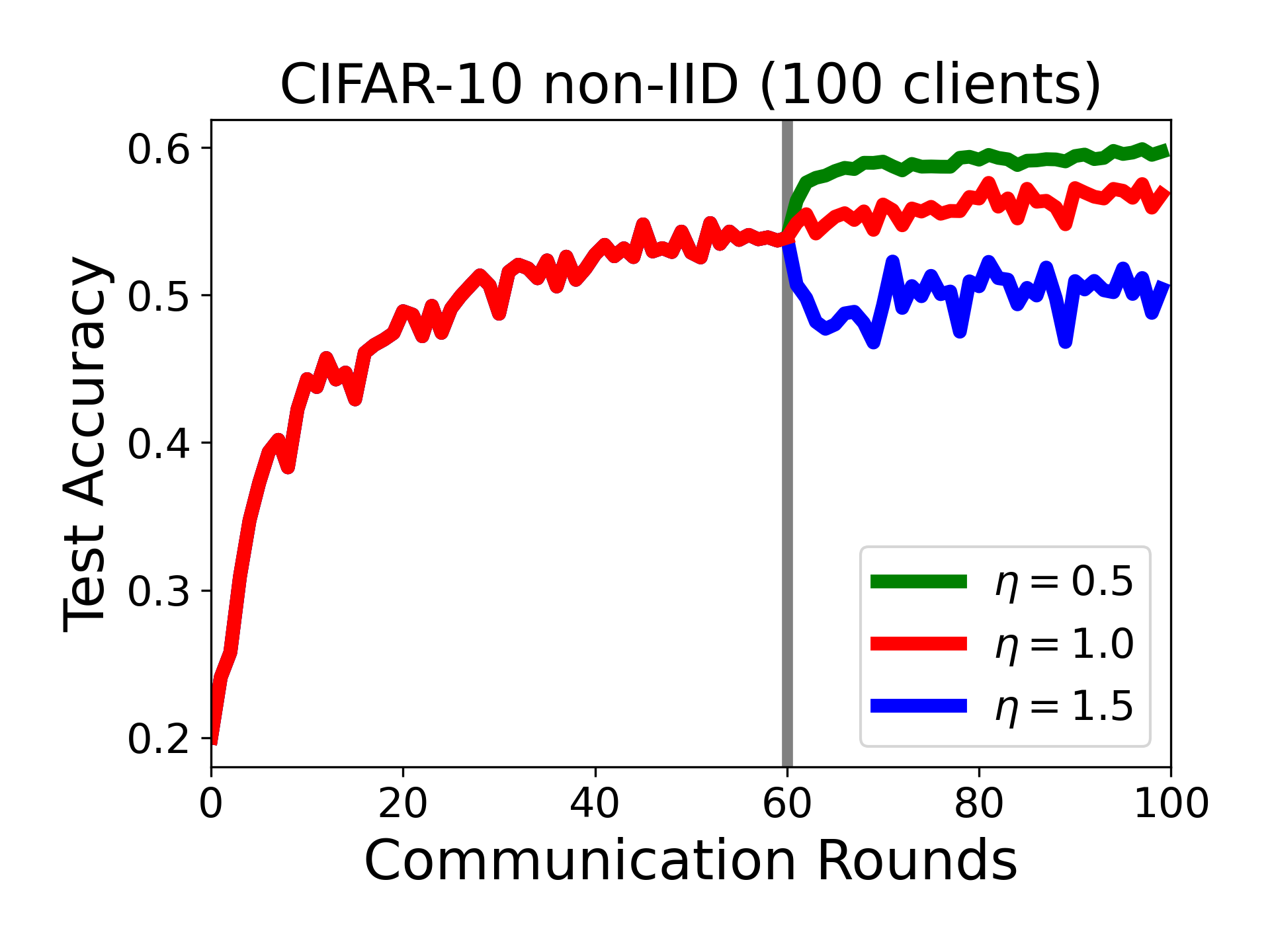}
  \end{subfigure}
  \caption{Performance comparison using different server step sizes for \texttt{FedADMM}. The vertical line indicates the point of when the step size is adjusted. The nominal step size $\eta=1.0$ yields consistent performance, while a decrease of the step size at later stages of the process can further improve the algorithm. }
  \label{stepsize}
\end{figure}

\subsection{Experimental Results}\label{5.2}
We first demonstrate the communication efficiency of \texttt{FedADMM} by comparing the number of rounds needed to reach a prescribed target accuracy. The results accompanied by a detailed description are summarized in Table \ref{big_table}. We note that \texttt{FedADMM} consistently outperforms all baseline methods in all cases tested. In specific, the communication savings achieved by \texttt{FedADMM} compared to the best baseline method in each case averages $72\%$. Recall also that \texttt{FedADMM} has $50\%$ less training computation than \texttt{FedAvg} and \texttt{SCAFFOLD} in all instances, since those were tested without system heterogeneity considerations, unlike \texttt{FedProx} and \texttt{FedADMM}.

\noindent \textbf{Increasing System Scale}. We compare the performance of \texttt{FedADMM} in different system scales using the FMNIST and CIFAR-10 datasets with both IID and non-IID data distributions. We first tuned all algorithms for their best performance in the 100-client setting, and then scale up system sizes with the hyperparameters fixed. The results are illustrated in Fig. \ref{scaling_figure}. 
% However, as we increase the system size while keeping the same fraction ($10\%$) of clients being selected at each round, \texttt{FedADMM} demonstrates significant advantages over baseline methods. We note that existing state-of-the-art methods not only have slower rate at initial stages, but also unstable oscillations may occur as the federated training process proceeds. 
Fig. \ref{figure8} serves to complement Fig. \ref{scaling_figure}: it considers the same datasets but for the reversed setting (IID becomes non-IID and vice versa). We keep the same fraction ($C=0.1$) of participating clients at each round, which translates to the same amount of data being used per round. Note, however, that an increase of the client population comes with the introduction of additional dual variables for \texttt{FedADMM}. In other words, the same amount of data is processed with more guidance, which serves to yield a larger improvement of \texttt{FedADMM} over baseline methods at larger scales. This favorable attribute is intensified in the non-IID setting where \texttt{FedADMM} is shown to achieve an effective incorporation of local information even faster. We conclude that \texttt{FedADMM} has formidable scalability, as also suggested by \textbf{Theorem 1}. In summary, the performance boost at increased scale (especially in the non-IID setting) is attributed to: (i) a higher aggregate number of information exchanges with the server, (ii) a higher total computational power in the network (albeit both at no additional cost per user), and (iii) additional dual variables, which provide extra guidance to the learning process.

\hide{the merits of \texttt{FedADMM} over other baseline methods are most pronounced in \emph{large-scale systems with heterogeneous data distributions}, where the effect of dual variables is most significant.}

\hide{ \noindent \textbf{Computation Heterogeneity}.
We note that \texttt{FedProx} is the only baseline method that allows selected clients to perform different amount of work locally. In Fig. \ref{stragglers}, we compare \texttt{FedADMM} against \texttt{FedProx} with different extent of computational heterogeneity using CIFAR-10 dataset with non-IID data distributions with 100 clients. We let different fractions (indicated by the percent of heterogeneity) of selected clients to perform different amount of local training to simulate computational heterogeneity due to hardware conditions. The corresponding fraction of clients selects a local training epoch number from the interval $[1,10]$ uniformly at random and the remaining fraction perform a fixed number of epochs $E=10$. }

\noindent \textbf{Data Heterogeneity}. To demonstrate the adaptability of \texttt{FedADMM} to heterogeneous data distributions, we fix parameters of \texttt{FedADMM} while tuning other methods for their best performance. Fig. \ref{adaption_figure} reveals the robust performance of \texttt{FedADMM} to statistical variations (with no hyperparameter tuning). This is a positive attribute for FL applications where data distributions are unpredictable and may be hard to characterize due to strict restrictions on clients' data privacy. As discussed in Section \ref{section3}, the dual variables in \texttt{FedADMM} serve as an automatic adaptation mechanism to data heterogeneity, by quantitatively recording the discrepancy between the local model and the global model. 

\noindent \textbf{Server step size}.
We explore the effect of adjusting the server gathering step size $\eta$ 
% using the MNIST and CIFAR-10 datasets 
in a system with 100 clients. In the left figure of the first row of Fig. \ref{stepsize}, we observe that when data distributions are IID across clients, all choices of $\eta$ result in similar performance, with the smallest $\eta=0.5$ being slightly slower at initial stages. This is due to the fact that when data are evenly distributed among clients, more drastic updates are allowed. When data distributions are non-IID, care must be taken in incorporating local information to the global model. This is reflected by the stalling in the case of setting $\eta = 1.5$. The nominal value $\eta=1.0$ showcases consistent performance in all cases. We additionally tested the effect of adjusting $\eta$ at later stages of the process (60 rounds): a decrease of the step size serves to incorporate past information in a finer fashion, thus improving the test accuracy.

\begin{table}[b]
    \caption{Number of rounds required  for \texttt{FedADMM} to reach given accuracy ($97\%$ for MNIST and $45\%$ for CIFAR-10) with variable amount of local training. }
    \centering
    \begin{tabular}{cc|ccc}
    \hline\hline
    & & $E=1$ & $E=5$ & $E=10$ \\
    \hline
     \multirow{2}{*}{MNIST}    & IID  & \textbf{27} & \textbf{10} & \textbf{6}   \\
         & non-IID & 56 & 33 & 32  \\
    \hline
     \multirow{2}{*}{CIFAR-10} & IID  & \textbf{24} & \textbf{12} & \textbf{10}   \\
      &  non-IID & 30 & 14 & 11 \\
      \hline
    \end{tabular}
    \label{local_work_table}
\end{table}
\begin{figure}[t]
   \centering
    \begin{subfigure}[b]{0.49\linewidth}
    \includegraphics[width=\textwidth]{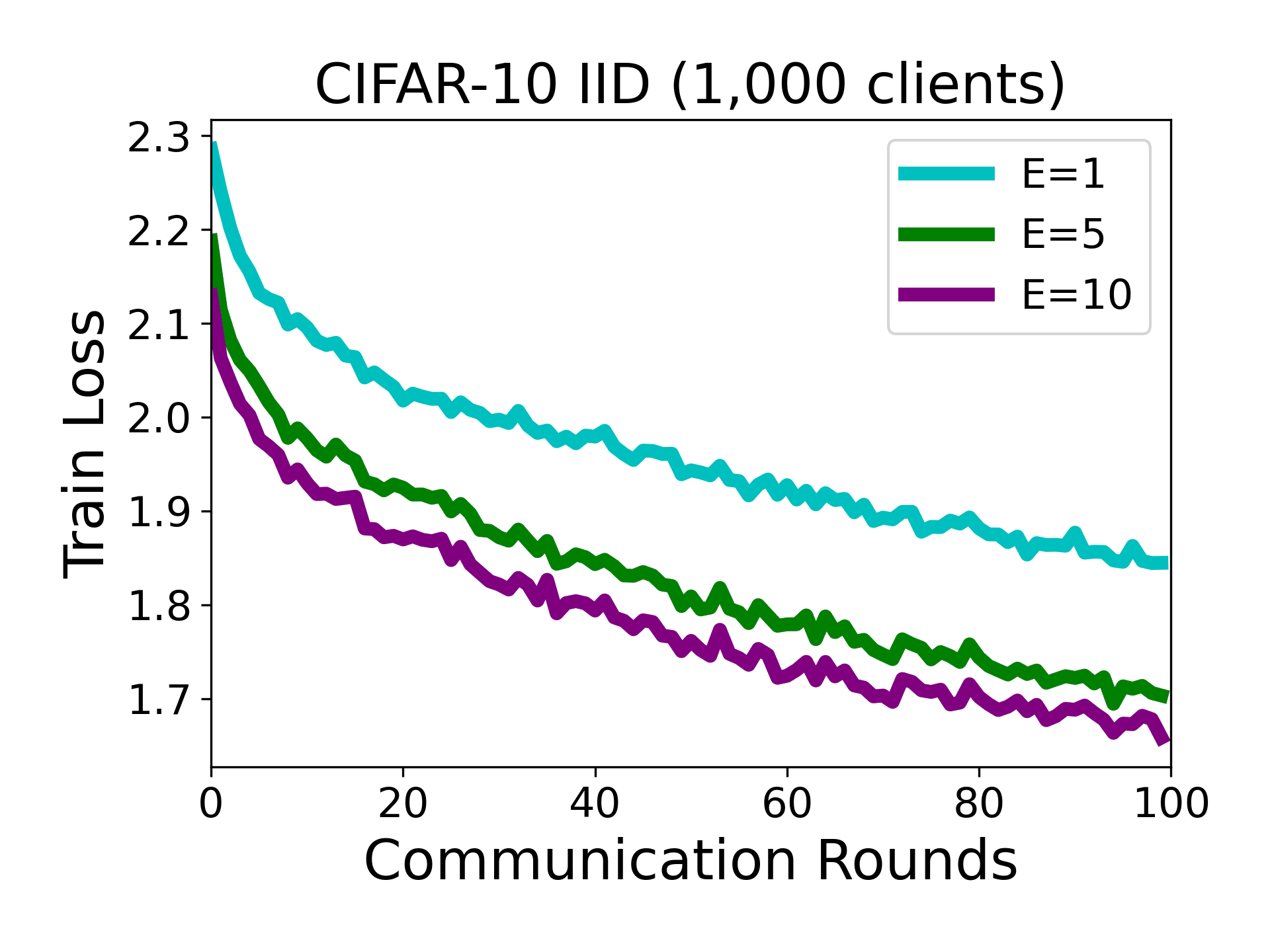}
  \end{subfigure}
    \begin{subfigure}[b]{0.49\linewidth}
    \includegraphics[width=\textwidth]{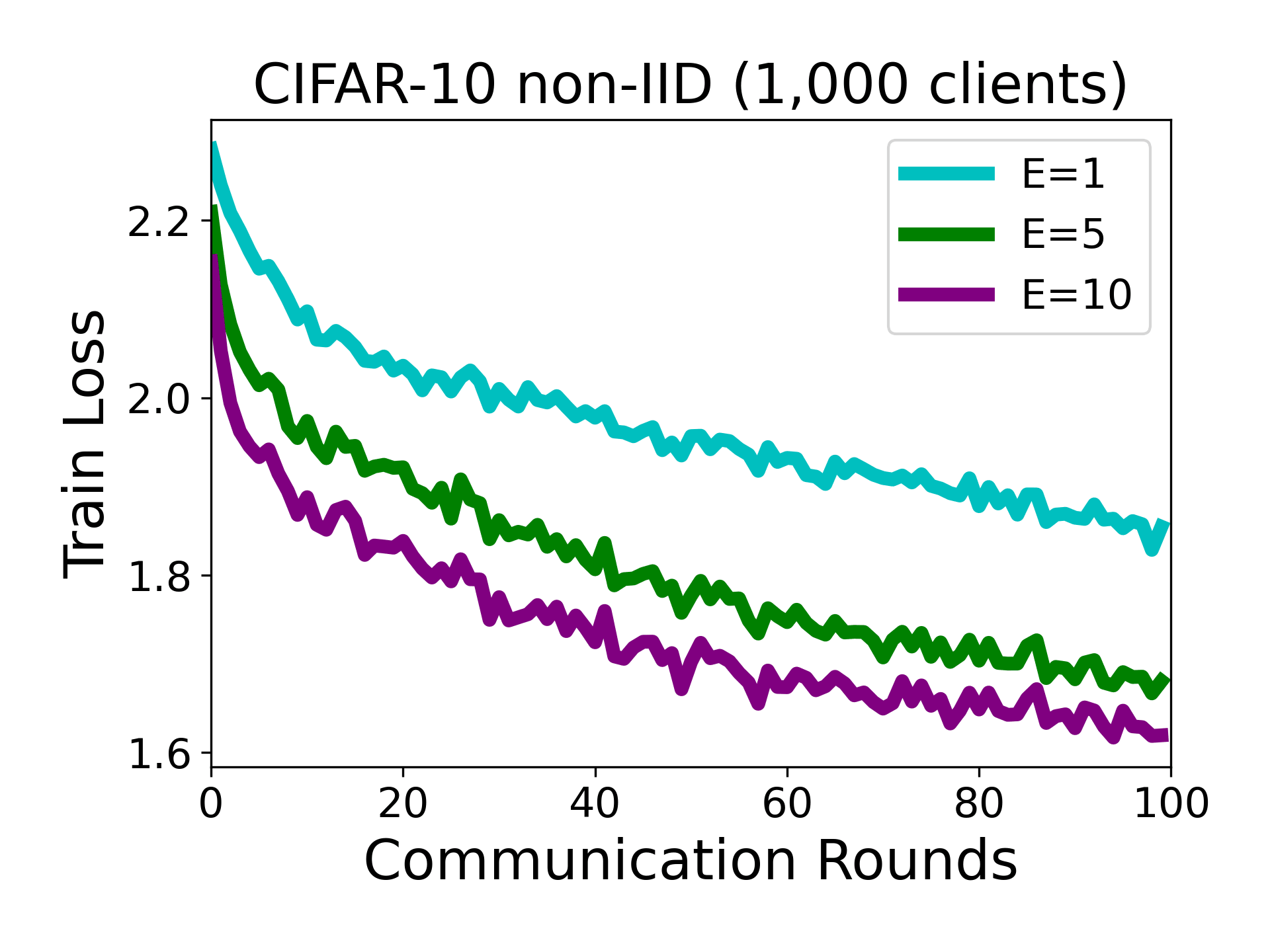}
  \end{subfigure}
  \caption{Performance boost with increased local epoch number $E$. Convergence of \texttt{FedADMM} is always ensured even with fixed learning rate and increasing amount of local training. }
  \label{local_work_figure}
\end{figure}

\noindent\textbf{Increasing Local Work}. 
We investigate how the local epoch number $E$ (capturing the amount of local training) affects the performance of \texttt{FedADMM}.
% while it also reflects the local solution accuracy in (\ref{varepsilon_i}).
Table \ref{local_work_table} and Fig. \ref{local_work_figure} indicate that a more accurate global model can be obtained \hide{within a fixed number of rounds} by increasing the amount of local training. This is in accord with our theoretical analysis (\textbf{Theorem 1}): the fact that the local subproblems are cast as strongly convex guarantees that the larger the local workload, the smaller the $\varepsilon_i$ in (\ref{varepsilon_i}), and thus the better the convergence.

\begin{figure}[t]
  \centering
    \begin{subfigure}[b]{0.49\linewidth}
    \includegraphics[width=\textwidth]{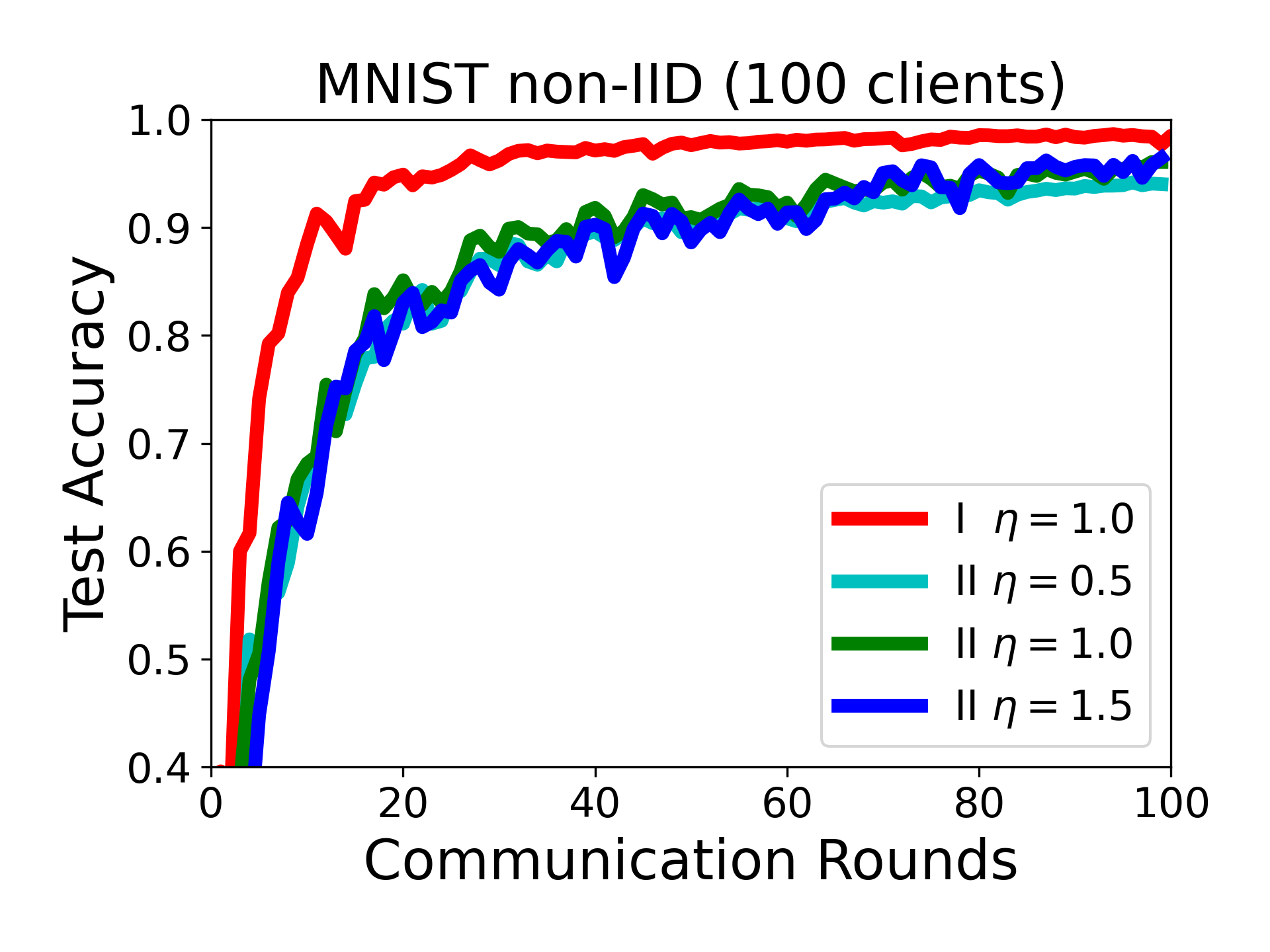}
  \end{subfigure}
  \begin{subfigure}[b]{0.49\linewidth}
    \includegraphics[width=\textwidth]{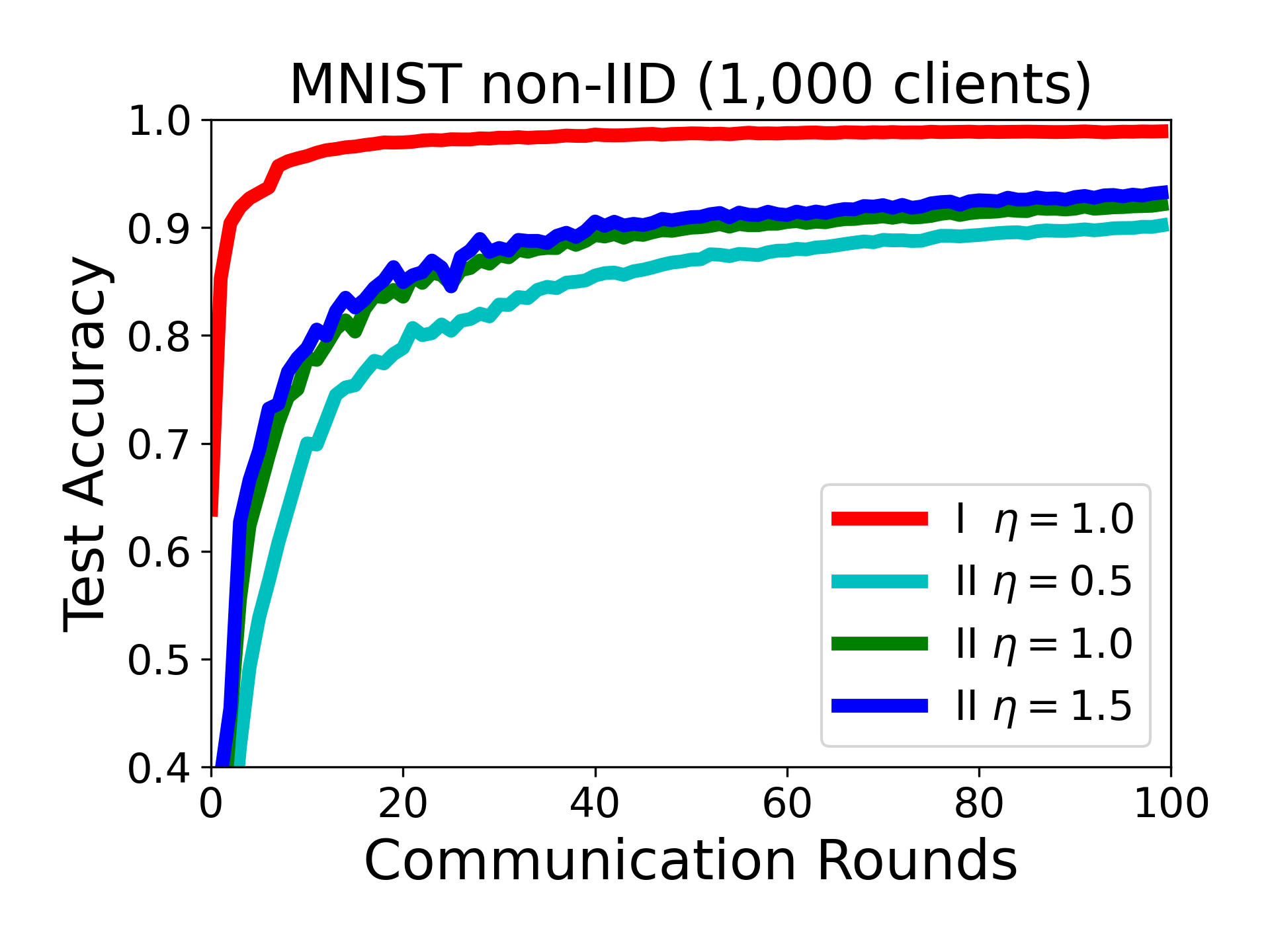}
  \end{subfigure}
      \begin{subfigure}[b]{0.49\linewidth}
    \includegraphics[width=\textwidth]{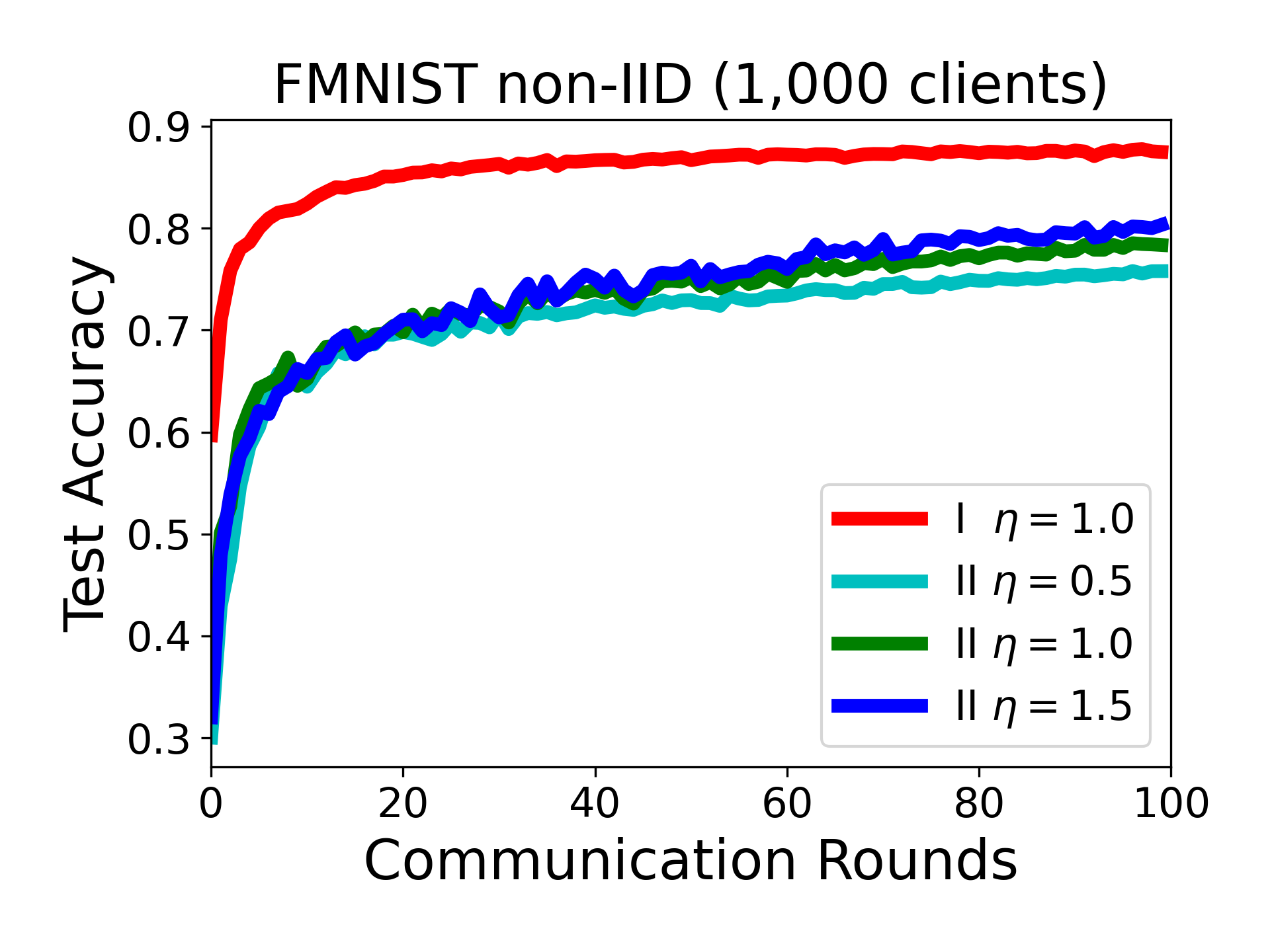}
  \end{subfigure}
  \begin{subfigure}[b]{0.49\linewidth}
    \includegraphics[width=\textwidth]{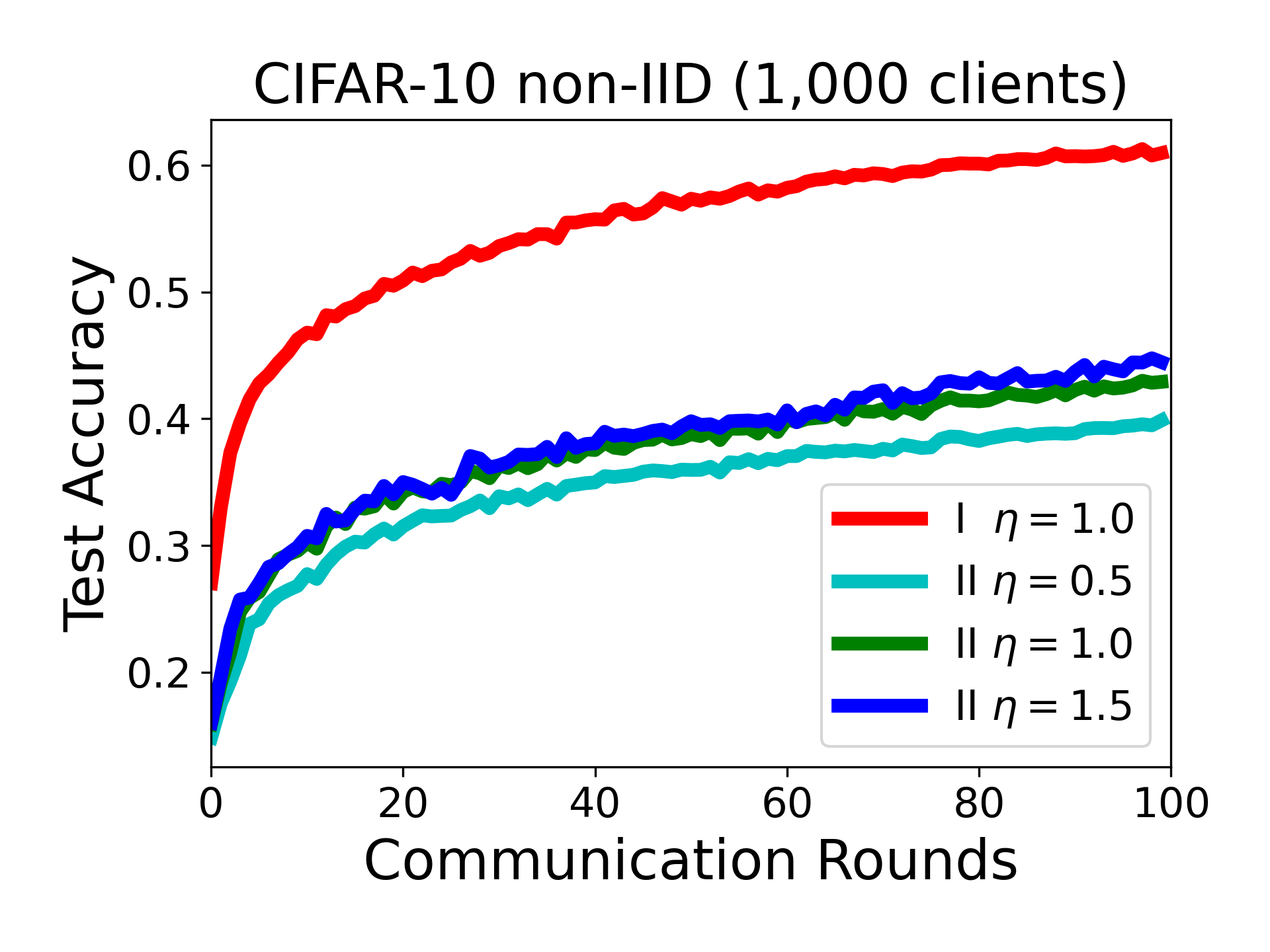}
  \end{subfigure}
  \caption{Different choices on local training initialization for \texttt{FedADMM}: I stands for initializing with the local model $w_i^t$, and II stands for  initializing with the global model $\theta^t$ (over different server step sizes). Warm starting local training with the local model $w_i^t$ yields superior results in all cases.}
  \label{initialization}
\end{figure}  

\begin{table}[b]
    \caption{Number of communication rounds to reach a prescribed accuracy (80\% for FMNIST, 97\% for MNIST). We use fixed $\rho=0.01$ for \texttt{FedADMM} while varying $\rho$ for \texttt{FedProx}. Note that different values of $\rho$ give drastically different results for of \texttt{FedProx}, while \texttt{FedADMM} is consistently superior with constant $\rho$. \hide{Noting the sensitivity of \texttt{FedProx} to the hyperparameter, it is revealed that \texttt{FedADMM} has consistently better performance than all settings of \texttt{FedProx}.} }
    \centering
    \begin{tabular}{ll|cccc}
    \hline\hline
     &  & \multicolumn{2}{l}{\quad 200 clients} & \multicolumn{2}{l}{\quad 500 clients}  \\
    Dataset & \makecell[c]{$\rho$} & IID & non-IID  & IID & non-IID \\
    \hline
    \multirow{4}{*}{MNIST} & \texttt{FedADMM} ($0.01$)  & \textbf{3} & \textbf{29} & \textbf{5} & \textbf{14} \\
     &\texttt{FedProx} ($0.01$)  & 100+ &100+ & 47 & 82 \\
    % \cline{2-5}
             &\texttt{FedProx} (0.1)  & 25 & 83 & 59 & 100+ \\
    % \cline{2-5}
             &\texttt{FedProx} (1)  & 81 & 100+ & 100+ & 100+ \\
    \hline
    \multirow{4}{*}{FMNIST}& \texttt{FedADMM} ($0.01$) & \textbf{2} & \textbf{13} & \textbf{2} & \textbf{9} \\
        &\texttt{FedProx} (0.01) & 5 & 34 & 11 & 45 \\
    % \cline{2-5}
             &\texttt{FedProx} (0.1)  & 7 & 45 & 13 & 54 \\
    % \cline{2-5}
             &\texttt{FedProx} (1)  & 43 & 100+ & 44 & 100+ \\
    \hline
    \end{tabular}
    \label{rho_2_round}
\end{table}

\noindent \textbf{Local Initialization}.
We further study the performance of \texttt{FedADMM} by investigating the impact of different initialization for the local training subproblems at selected clients. As shown in Fig. \ref{initialization}, using client model parameters $w_i^t$ as a warm start for the local SGD demonstrates significant advantages compared to using the global model $\theta^t$. This provides additional motivation for clients to store their local model $w_i$ as compared to using the global model $\theta$ as the starting point for local training.

\begin{figure}[t]
  \centering
  \begin{subfigure}[b]{0.49\linewidth}
    \includegraphics[width=\textwidth]{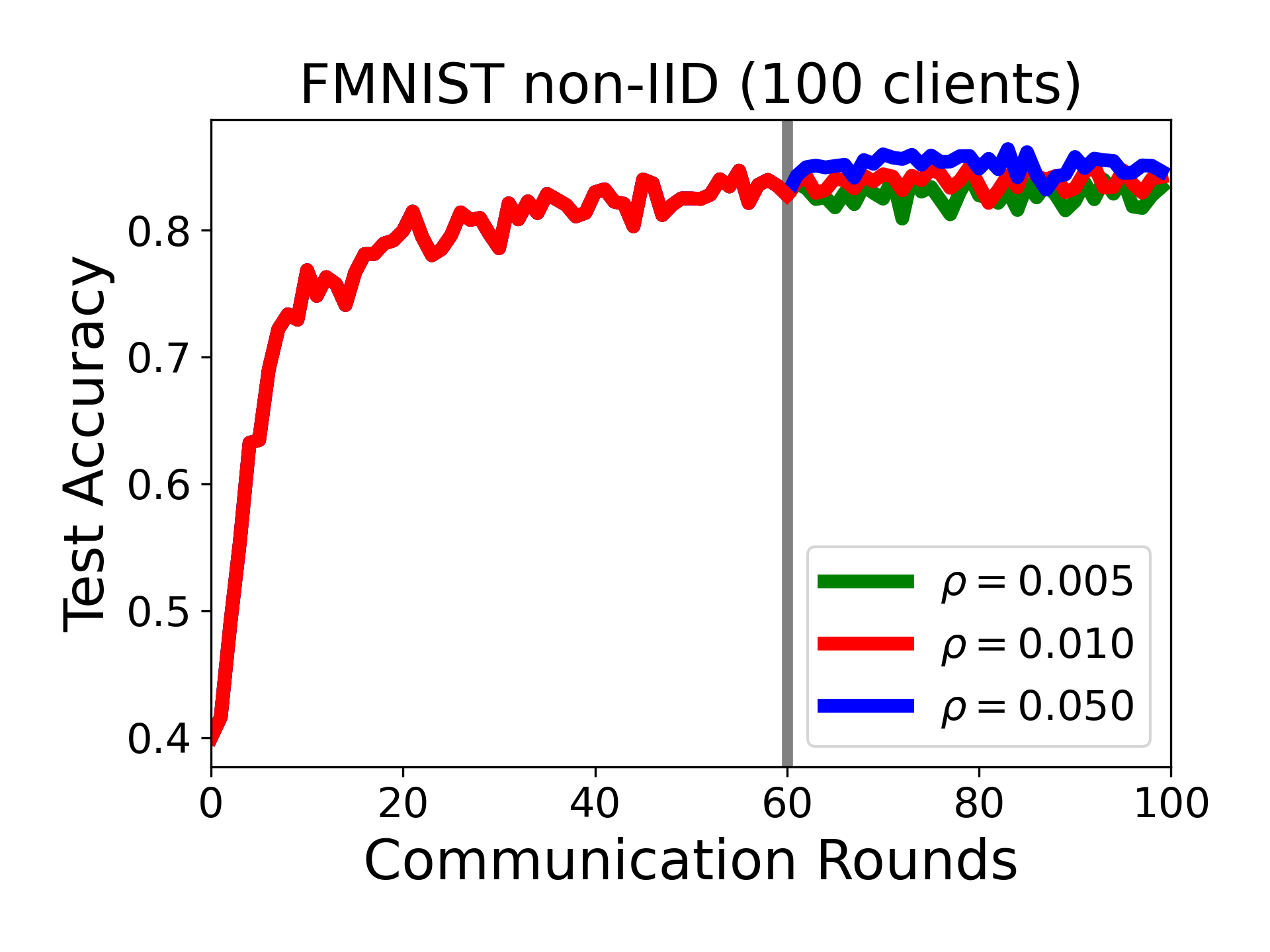}
  \end{subfigure}
      \begin{subfigure}[b]{0.49\linewidth}
    \includegraphics[width=\textwidth]{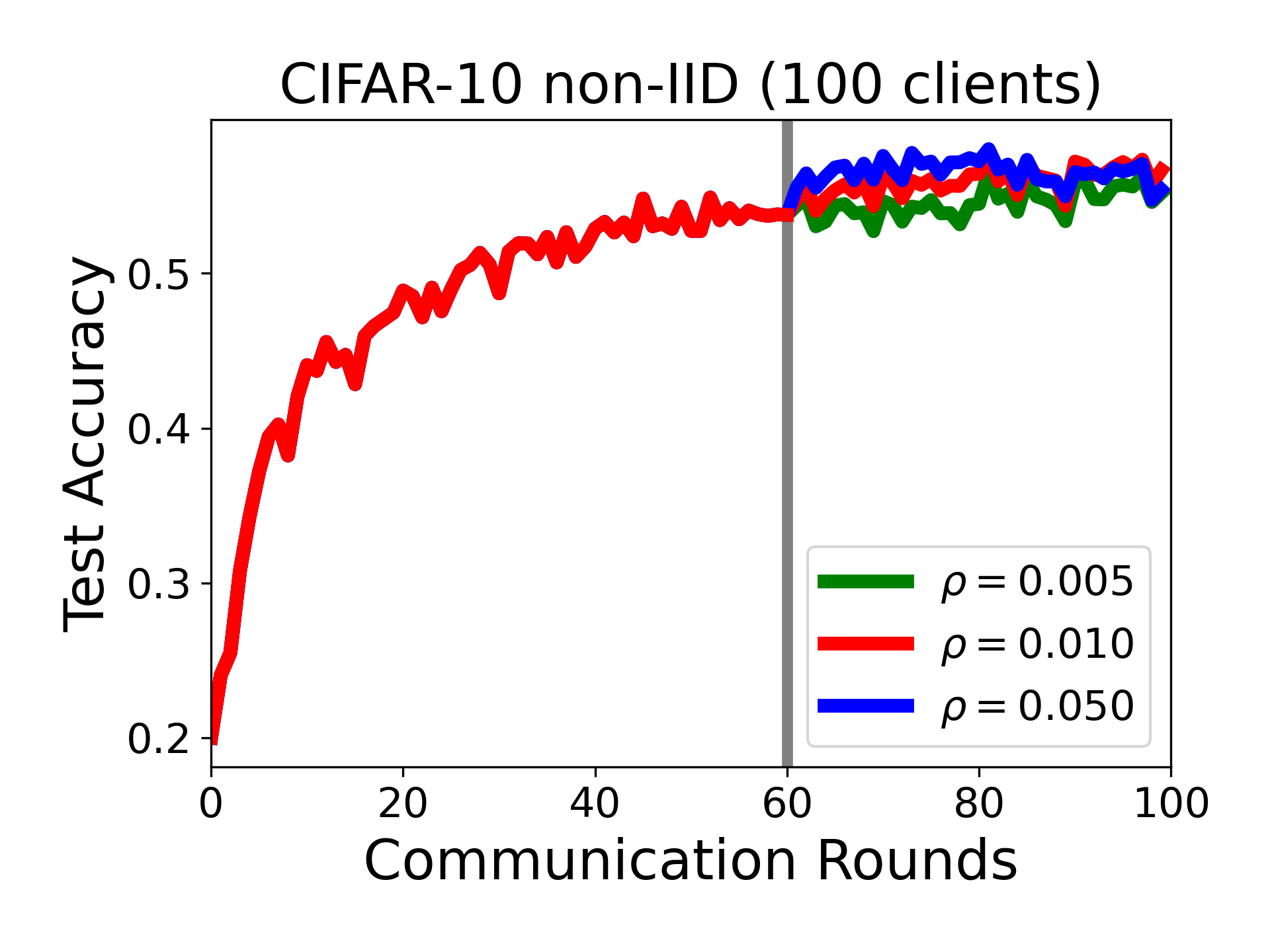}
  \end{subfigure}
  \caption{Performance comparison with different value of proximal hyperparameter $\rho$ in \texttt{FedADMM}. The vertical line indicates the point when $\rho$ is changed.}
  \label{rho}
\end{figure}

\noindent \textbf{Proximal Parameter $\rho$}. To counter \textit{client drift} in non-IID data distributions,  \texttt{FedADMM} and \texttt{FedProx} both use a quadratic proximal term for the local training problem. The proximal coefficient $\rho$ in \texttt{FedProx} has to be carefully tuned to achieve competitive performance across different settings \cite{Fedprox}, \cite{scaffold}. Such tuning is dependent on system sizes and data distributions, as we also demonstrate in Table \ref{rho_2_round}. Note that the best $\rho$ value (0.01) for \texttt{FedProx} in the FMNIST dataset gives the worst performance for MNIST in the 200-client setting. Moreover, the performance of \texttt{FedProx} with respect to $\rho$ is not monotone, which makes tuning even more challenging.\hide{First, the best value of $\rho$ varies with the dataset and the system size (i.e., in the 200-client settings, the best $\rho$ in FMNIST introduces the worst performance in MNIST). Second, the performance will have significantly variation depend on the data distribution, which is dynamically changing in FL scenarios.} On the other hand, \texttt{FedADMM} dominates all tested instances of \texttt{FedProx} with fixed $\rho=0.01$. This is also supported by our theoretical analysis (\textbf{Theorem 1} and \textbf{Remark 1} supports a constant $\rho$).
Additional insights can be gained by a simple dynamic adaptation of $\rho$ for \texttt{FedADMM} in Fig. \ref{rho}. A smaller value (0.01) at initial stages of training allows efficient incorporation of local data when the global model is not informed, while an increase of $\rho$ at later stages reduces discrepancies between client models and the global model.

\begin{table}[b]
    \caption{Statistics of imbalanced datasets. }
    \centering
    \begin{tabular}{lcccc}
    \hline\hline
    Dataset & Clients & Samples & Mean & Stdev \\
    \hline
    FMNIST & 200 & 60,000 & 300 & 171.03 \\
    CIFAR-10 & 200 & 50,000 & 250 & 142.52 \\
    \hline
    \end{tabular}
    \label{unequal_details}
\end{table}

\begin{figure}[b]
  \centering
    \begin{subfigure}[b]{0.49\linewidth}
    \includegraphics[width=\textwidth]{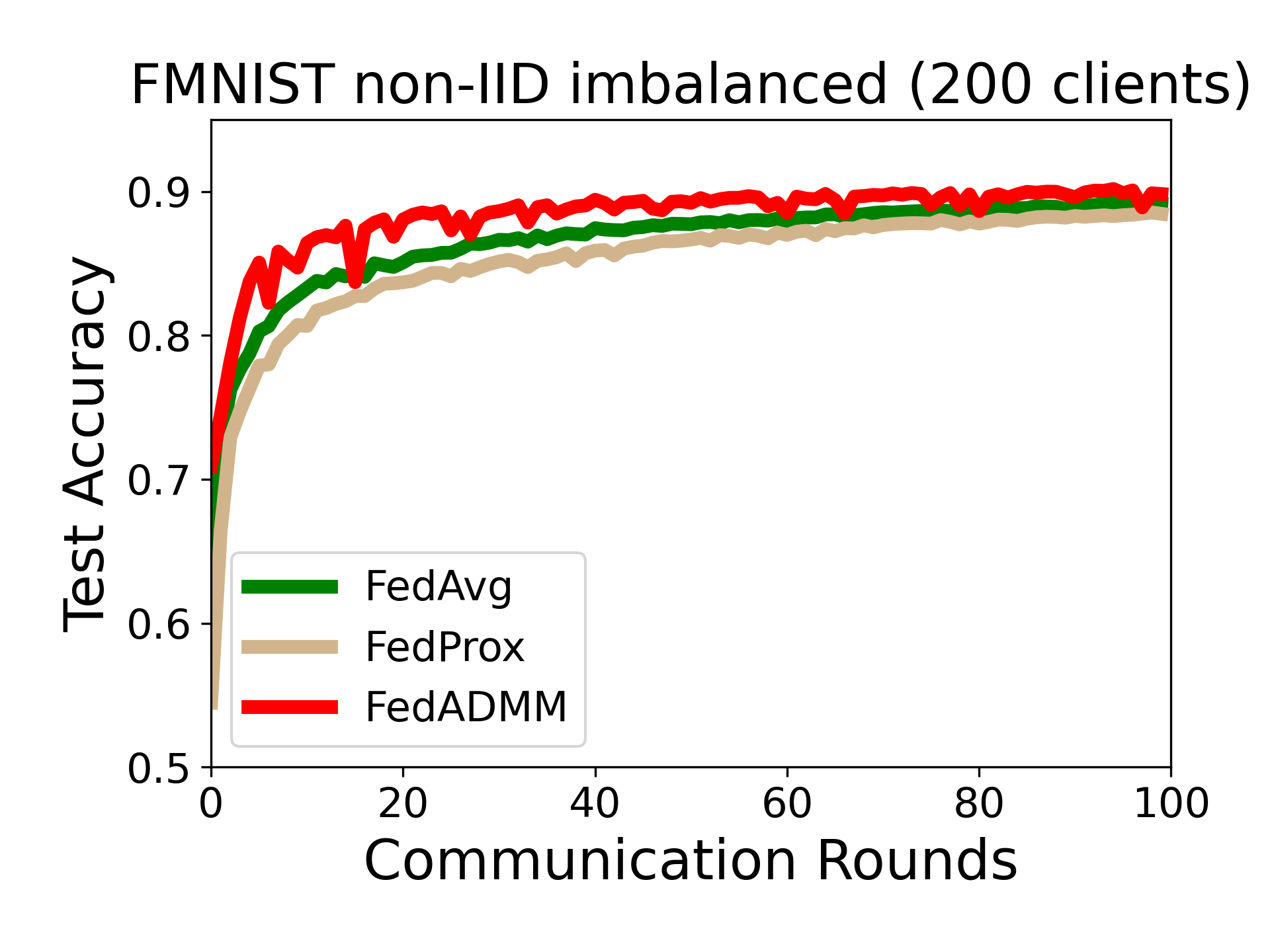}
  \end{subfigure}
  \begin{subfigure}[b]{0.49\linewidth}
    \includegraphics[width=\textwidth]{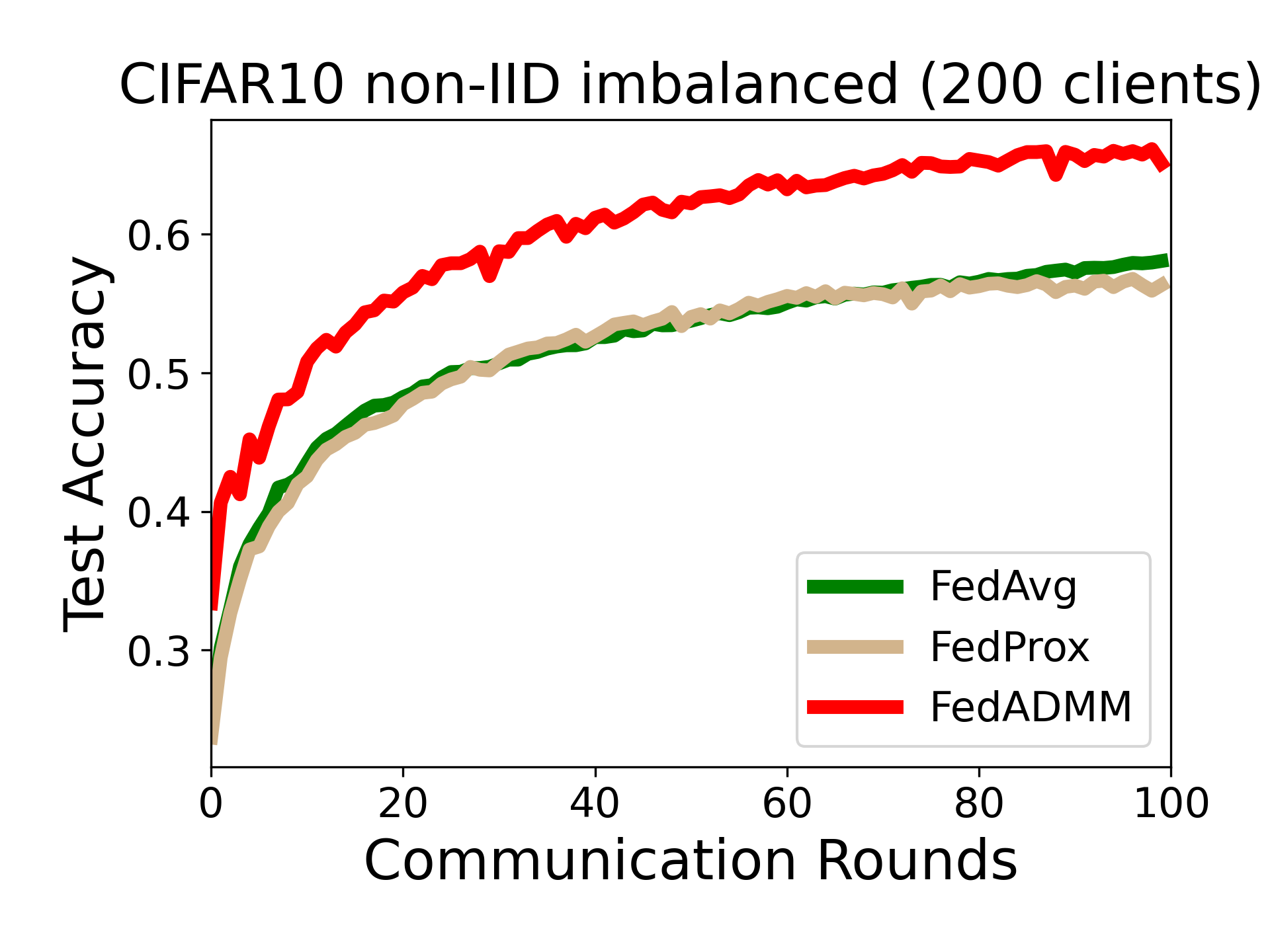}
  \end{subfigure}
  \caption{\texttt{FedADMM} achieves the best performance over baseline algorithms for imbalanced datasets ($E=10$, $B=50$).}
  \label{imbalanced}
\end{figure}

\noindent \textbf{Imbalanced Data Volumes}. We explore the more realistic scenario that clients hold different data volumes. In this setting, we first sort the training data points by labels, then divide the training data into $10,000$ shards, each with 5 data points for CIFAR-10 and 6 data points for FMNIST, respectively. We divide 200 clients evenly into 100 groups. Each member of the group is assigned with a  number of shards that equals the group index, except for the last group that collects the remaining data.\hide{This setting aims to reflect a rather extreme case of data imbalance in the system.} The setting is summarized in Table \ref{unequal_details}. Fig. \ref{imbalanced} shows that \texttt{FedADMM} achieves higher accuracy than other baselines, especially in the CIFAR-10 dataset.

\section{Conclusion}
\texttt{FedADMM} is a new federated learning method that handles both statistical and system heterogeneity, without inducing extra communication costs per round. By storing an extra dual variable at each client, a significant speedup over the state-of-the-art is achieved which translates to substantial communication savings. Moreover, FedADMM features automatic adaptation to statistical variations within the system without the need for hyperparameter tuning. Such adaptation safeguards against client drift that causes \texttt{FedAvg} to diverge. We have established an optimal convergence rate (which matches the problem complexity lower bound), with no assumptions on data dissimilarity or gradient boundedness. The analytical results are corroborated by extensive experiments that reveal that FedADMM is well-suited for FL applications with large system sizes and heterogeneous data distributions.

\section{Full proof}\label{full proof}

In this section, we present the full proof of our theoretical results. Our proof of convergence is inspired by \cite{hongadmm} but differs in the following aspects allusive to our specific design: (i) the server updates at each round; (ii) the local training problems are solved inexactly with the level of inexactness captured by the local parameter $\varepsilon_i$ as in (\ref{varepsilon_i}); (iii) an additional step size $\eta$ for server aggregation is included.
For ease of exposition, we fix $\eta=\abs{S^t}/m$ in the analysis and stack variables as follows:
$
    w^{t} = [(w_1^t)^\top,\dots,(w_m^t)^\top]^\top,
    y^{t} = [(y_1^t)^\top,\dots,(y_m^t)^\top]^\top \in \mathbb{R}^{md}
$. We define the \emph{virtual} dual update as follows:
\begin{gather}
   \norm{\nabla_{w_i} \mathcal{L}_i(\widehat{w}_i^{t+1},y_i^t,\theta^t)}^2 \leq \varepsilon_i, \label{virtual_primal}\\
    \widehat{y}_i^{t+1} = y_i^t+\rho (\widehat{w}_i^{t+1}-\theta^t).\label{virtual}
\end{gather}
In other words, the virtual update for client $i$ is the value for $(w_i^{t+1},y_i^{t+1})$ if it participates in the current round, i.e., if $i\in S^t$, $w_i^{t+1}=\widehat{w}_i^{t+1}$ and $y_i^{t+1}=\widehat{y}_i^{t+1}$. As a consequence, we have: 
$$
    \mathbb{E}^t[w_i^{t+1}-w_i^t]= 
     p_i(\widehat{w}_i^{t+1}-w_i^t),
$$
where $\mathbb{E}^t[\cdot]$ denotes the conditional expectation at step $t$, and $p_i$ denotes the probability of client $i$ being active. We proceed to bound $\norm{y_i^{t+1}-y_i^t}$ in the following lemma.

\noindent \textbf{Lemma 1}: Recall assumption 1 and the local accuracy $\varepsilon_i$ in (\ref{varepsilon_i}). The consecutive difference between dual variables can be bounded as follows: 
\begin{gather}
  \norm{\widehat{y}_i^{t+1}-y_i^t}^2\leq 8 \varepsilon_i+2L^2\norm{\widehat{w}_i^{t+1}-w_i^t}^2, \label{lemma1_1}\\
        \norm{y_i^{t+1}-y_i^t}^2\leq 8 \varepsilon_i+2L^2\norm{w_i^{t+1}-w_i^t}^2.\label{lemma1_2}
\end{gather}
\textit{Proof}: We first prove (\ref{lemma1_1}) by defining the local error term as 
\begin{align}
    e_i^{t+1} &=\nabla_{w_i}\mathcal{L}_i(\widehat{w}_i^{t+1},y_i^t,\theta^t)\nonumber \\
    &= \nabla f_i(\widehat{w}_i^{t+1})+y_i^t+ \rho(\widehat{w}_i^{t+1}-\theta^t) \nonumber \\
    &= \nabla f_i(\widehat{w}_i^{t+1})+\widehat{y}_i^{t+1}, \label{error}
\end{align}
where the last equality holds from the definition of virtual update rule (\ref{virtual}). Note that $\norm{e_i^{t+1}}=\norm{\nabla_{w_i}\mathcal{L}_i(\widehat{w}_i^{t+1},y_i^t,\theta^t)}\leq\sqrt{\varepsilon_i}$. By rearranging and taking the difference, 
\begin{align*}
    \norm{\widehat{y}_i^{t+1}-y_i^t} &= \norm{e_i^{t+1}-e_i^{t}+\nabla f_i(w_i^{t})-\nabla f_i(\widehat{w}_i^{t+1})}\\
    &\leq \norm{e_i^{t+1}}+\norm{e_i^t}+\norm{\nabla f_i(w_i^t)-\nabla f_i(\widehat{w}_i^{t+1})} \\
    &\leq 2\sqrt{\varepsilon_i}+L\norm{\widehat{w}_i^{t+1}-w_i^t}.
\end{align*}
Using the inequality $\norm{\sum_{i=1}^m a_i }^2\leq m \sum_{i=1}^m \norm{a_i}^2$, we obtain: 
$$
    \norm{\widehat{y}_i^{t+1}-y_i^t}^2 \leq 8\varepsilon_i+2L^2 \norm{\widehat{w}_i^{t+1}-w_i^t}^2,
$$
which is the desired. We note that (\ref{lemma1_1}) holds for all $i\in [m]$ and proceed to show (\ref{lemma1_2}) as follows. For $i\in S^t$, $y_i^{t+1}=\widehat{y}_i^{t+1}$, $w^{t+1}_i=\widehat{w}^{t+1}_i$, and therefore (\ref{lemma1_2}) reduces to (\ref{lemma1_1}). For $i\notin S^t$, $w_i^{t+1}=w_i^t$, $y_i^{t+1}=y_i^t$, and therefore (\ref{lemma1_2}) trivially holds.    \QEDB

We denote the aggregated Lagrangian as $\mathcal{L}(w^t,y^t,\theta^t)=\sum_{i=1}^m \mathcal{L}_i(w_i^t,y_i^t,\theta^t)$ and proceed to bound its change after one round. 

\noindent \textbf{Lemma 2}: The aggregated Lagrangian $\mathcal{L}^t\equiv \mathcal{L}(w^t,y^t,\theta^t)$ iterates satisfy: 
\begin{gather}
    \mathcal{L}^{t+1}-\mathcal{L}^t\leq \sum_{i\in S^t}\Bigg(\bigg(\tfrac{2L-\rho}{2}+\tfrac{2L^2}{\rho }\bigg)\norm{w_i^{t+1}-w_i^t}^2\nonumber\\
    +\tfrac{\varepsilon_i}{2L}+\tfrac{8\varepsilon_i}{\rho}\Bigg)-\tfrac{m\rho}{2}\norm{\theta^{t+1}-\theta^t}^2.
\end{gather}
\textit{Proof}: We decompose the difference into three parts:
\begin{gather}
    \mathcal{L}^{t+1}-\mathcal{L}^t= \mathcal{L}(w^{t+1},y^{t+1},\theta^{t+1})-\mathcal{L}(w^{t+1},y^{t+1},\theta^{t})\nonumber\\
    +\mathcal{L}(w^{t+1},y^{t+1},\theta^{t})-\mathcal{L}(w^{t+1},y^{t},\theta^{t})\nonumber\\
    +\mathcal{L}(w^{t+1},y^{t},\theta^{t})- \mathcal{L}(w^{t},y^{t},\theta^{t}). \label{AL diff}
\end{gather}
The three difference terms in (\ref{AL diff}) correspond to the $\theta^{t+1}$-update, $y^{t+1}$-update, and $w^{t+1}$-update, respectively. We proceed to bound each term separately. Note that $\mathcal{L}(\cdot)$ is strongly convex with respect to $\theta$ with parameter $m\rho$. Therefore, 
\begin{gather}
    \mathcal{L}(w^{t+1},y^{t+1},\theta^t) \geq \mathcal{L}^{t+1} +\langle\nabla_{\theta}\mathcal{L}^{t+1},\theta^t-\theta^{t+1}\rangle\nonumber\\
    +\tfrac{m\rho}{2}\norm{\theta^{t+1}-\theta^t}^2.\label{first term}
\end{gather}
By definition, 
\begin{gather}
    \nabla_{\theta}\mathcal{L}^{t+1}=\rho\left(m\theta^{t+1}-\sum_{i=1}^m (w_i^{t+1}+\tfrac{1}{\rho}y_i^{t+1})\right). \label{theta derivative}
\end{gather}
When the server step size is chosen as $\eta=\abs{S^t}/m$, we have
\begin{align*}
    \theta^{t+1}&=\theta^t+\tfrac{1}{m}\sum_{i\in S^t}\left(w_i^{t+1}+\tfrac{1}{\rho}y_i^{t+1}-(w_i^t+\tfrac{1}{\rho}y_i^t)\right)\\
    &= \theta^t+\tfrac{1}{m}\sum_{i=1}^m (u_i^{t+1}-u_i^t),
\end{align*}
where we have defined the \emph{augmented model} $u_i^{t} = w_i^{t}+\tfrac{1}{\rho}y_i^{t}$ and used the fact that for $i\notin S^t$, $u_i^{t+1}-u_i^t=0$. After telescoping, we obtain 
$$
    \theta^{t+1} = \theta^0+\tfrac{1}{m}\sum_{i=1}^m(u_i^{t+1}-u_i^0).
$$ From our initialization, $w_i^0=\theta^0$ and $y_i^0=0$, it follows that $\tfrac{1}{m}\sum_{i=1}^m u_i^0=\theta^0$. Therefore, 
$
    \theta^{t+1} = \tfrac{1}{m}\sum_{i=1}^m u_i^{t+1} 
$. Using this fact along with (\ref{theta derivative}), we obtain 
\begin{gather}
    \nabla_{\theta}\mathcal{L}^{t+1}=\rho\sum_{i=1}^m\left( u_i^{t+1}-(w_i^{t+1}+\tfrac{1}{\rho}y_i^{t+1})\right)=0 \label{zero_deri},
\end{gather}
where we used the definition $u_i^{t+1}=w_i^{t+1}+\tfrac{1}{\rho}y_i^{t+1}$. Therefore, we can rewrite (\ref{first term}) as
\begin{gather}
    \mathcal{L}^{t+1}-\mathcal{L}(w^{t+1},y^{t+1},\theta^t)\leq -\tfrac{m\rho}{2}\norm{\theta^{t+1}-\theta^t}^2, \label{one}
\end{gather}
which is the bound for the first difference term in (\ref{AL diff}). By definition, the second difference term in (\ref{AL diff}) can be expressed as:
\begin{align*}
    &\mathcal{L}(w^{t+1},y^{t+1},\theta^t)-\mathcal{L}(w^{t+1},y^t,\theta^t) \qquad\\
    &= \sum_{i\in S^t} \langle y_i^{t+1}-y_i^t, w_i^{t+1}-\theta^t \rangle \\
    &=\sum_{i \in S^t} \tfrac{1}{\rho}\norm{y_i^{t+1}-y_i^t}^2,
\end{align*}
where the last equality follows from (\ref{virtual}) and the fact that for $i\in S^t$, $y_i^{t+1}=\widehat{y}_i^{t+1}$, $w_i^{t+1}=\widehat{w}_i^{t+1}$. Using Lemma 1, we obtain
\begin{gather}
    \mathcal{L}(w^{t+1},y^{t+1},\theta^t)-\mathcal{L}(w^{t+1},y^t,\theta^t) \nonumber\\
    \leq \sum_{i\in S^t}\bigg(\tfrac{8\varepsilon_i}{\rho}+\tfrac{2L^2}{\rho} \norm{w_i^{t+1}-w_i^t}^2\bigg),\label{second}
\end{gather}
which is the bound for the second difference term in (\ref{AL diff}). For the third difference term, we first note that $-\nabla f_i(\cdot)$ is Lipschitz as well. Therefore, the following holds:
\begin{gather}
    -f_i(w_i^t)\leq -f_i(w_i^{t+1})+\langle -\nabla f_i(w_i^{t+1}), w_i^t-w_i^{t+1}  \rangle\nonumber\\
    +\tfrac{L}{2}\norm{w_i^{t+1}-w_i^t}^2, \label{-Lip}
\end{gather}
and $\forall\,i\in S^t$:
\begin{align}
    &\mathcal{L}_i(w_i^{t+1},y_i^t,\theta^t)-\mathcal{L}_i(w_i^t,y_i^t,\theta^t)\nonumber\\
    &= f_i(w_i^{t+1})-f_i(w^t_i)
    +\langle y_i^t,w_i^{t+1}-w_i^t \rangle\nonumber\\
    &\quad +\tfrac{\rho}{2}\left(\norm{w_i^{t+1}-\theta^t}^2-\norm{w_i^t-\theta^t}^2\right) \nonumber\\
    &\underset{(\text{i})}{\leq} \langle \nabla f_i(w_i^{t+1}),w_i^{t+1}-w_i^t  \rangle+\tfrac{L}{2}\norm{w_i^{t+1}-w_i^t}^2\nonumber\\
    &\quad +\langle y_i^t,w_i^{t+1}-w_i^t \rangle+\tfrac{\rho}{2}\langle w_i^{t+1}+w_i^t-2\theta^t,w_i^{t+1}-w_i^t \rangle \nonumber\\
    &\underset{(\text{ii})}{=}\langle \nabla f_i(w_i^{t+1})+y_i^t+\tfrac{\rho}{2}\cdot 2(w_i^{t+1}-\theta^t),w_i^{t+1}-w_i^t\rangle \nonumber \\
    &\quad +\tfrac{L}{2}\norm{w_i^{t+1}-w_i^t}^2 +\tfrac{\rho}{2}\langle w_i^t-w_i^{t+1},w_i^{t+1}-w_i^t\rangle\nonumber\\
    &\underset{(\text{iii})}{\leq} \tfrac{1}{2L}\norm{\nabla f_i(w_i^{t+1})+y_i^{t+1}}^2+\tfrac{2L-\rho}{2}\norm{w_i^{t+1}-w_i^t}^2 \nonumber \\
    &\underset{(\text{iv})}{\leq} \tfrac{\varepsilon_i}{2L}+\tfrac{2L-\rho}{2}\norm{w_i^{t+1}-w_i^t}^2.\label{3rd_diff}
\end{align}
In (\ref{3rd_diff}): (i) follows from (\ref{-Lip}) and the identity $\norm{a}^2-\norm{b}^2= (a+b)^\top(a-b)$; (ii) follows from splitting $w_i^{t+1}+w_i^t-2\theta^t =2(w_i^{t+1}-\theta^t)+w_i^t-w_i^{t+1}$; (iii) follows from the dual update: $y_i^{t+1} = y_i^t + \rho(w_i^{t+1}-\theta^t)$ and the identity $a^\top b \leq \tfrac{1}{2L}\norm{a}^2+\tfrac{L}{2}\norm{b}^2$; (iv) follows from (\ref{error}) and (\ref{virtual_primal}).
Therefore, the third difference term in (\ref{AL diff}) can be bounded as:
\begin{gather}
    \mathcal{L}(w^{t+1},y^t,\theta^t)-\mathcal{L}^t \leq \sum_{i\in S^t}\tfrac{2L-\rho}{2}\norm{w_i^{t+1}-w_i^t}^2
    +\sum_{i\in S^t}\tfrac{\varepsilon_i}{2L}.\label{final}
\end{gather}
After combining (\ref{one}), (\ref{second}) and (\ref{final}), we obtain the desired. \QEDB

We proceed to establish a lower bound for the augmented Lagrangian in the following lemma.   

\noindent \textbf{Lemma 3}. Recall the lower bound $\sum_{i=1}^m f_i(w)\geq f^\star$ in assumption 2. For $t\geq0$, the augmented Lagrangian is lower bounded as $\mathcal{L}^{t+1}\geq f^\star-\tfrac{1}{2L}\sum_{i=1}^m \varepsilon_i$ by selecting $\rho\geq 2L$.

\noindent \textit{Proof}: By definition, 
\begin{align}
    \mathcal{L}^{t+1}
    &=\sum_{i=1}^m \bigg(f_i(w_i^{t+1})+\langle y_i^{t+1},w_i^{t+1}-\theta^{t+1} \rangle\nonumber\\\
    &\quad +\tfrac{\rho}{2}\norm{w_i^{t+1}-\theta^{t+1}}^2\bigg) \nonumber \\
    &\underset{(\text{i})}{=}\sum_{i=1}^m \bigg(f_i(w_i^{t+1})+\langle \nabla f_i(w_i^{t+1}),\theta^{t+1}-w_i^{t+1} \rangle\nonumber\\
    &\quad+\tfrac{\rho}{2}\norm{w_i^{t+1}-\theta^{t+1}}^2+\langle e_i^{t+1},w_i^{t+1}-\theta^{t+1}\rangle\bigg) \nonumber \\
    &\underset{(\text{ii})}{\geq}\sum_{i=1}^m \bigg(f_i(\theta^{t+1})+\tfrac{\rho-2L}{2}\norm{w_i^{t+1}-\theta^{t+1}}^2-\tfrac{1}{2L}\varepsilon_i\bigg) \nonumber \\
    &\geq \sum_{i=1}^m f_i(\theta^{t+1})-\tfrac{1}{2L}\sum_{i=1}^m \varepsilon_i\geq f^\star-\tfrac{1}{2L}\sum_{i=1}^m\varepsilon_i.
\end{align}
To show equality (\text{i}) holds, it suffices to show: $\forall\,i$, $y_i^{t+1}=e_i^{t+1}-\nabla f_i(w_i^{t+1})$. This holds for $i\in S^t$ by (\ref{error}) and the fact that $\widehat{w}_i^{t+1}=w_i^{t+1}, \widehat{y}_i^{t+1}=y_i^{t+1}$. For $i\notin S^t$, $w_i^{t+1}=w_i^t=w_i^{t_i+1}$, where we denote $t_i$ as the most recent time step when client $i$ is selected before step $t$. Therefore, for $i\notin S^t$, $y_i^{t+1}=y_i^t=y_i^{t_i+1}=e_i^{t_i+1}-\nabla f_i(w_i^{t_i+1})=e_i^{t+1}-\nabla f_i(w_i^{t+1})$. The inequality \text(ii) follows from a consequence of assumption 1, the identity $a^\top b\geq -\tfrac{L}{2}\norm{a}^2-\tfrac{1}{2L}\norm{b}^2$, and $\norm{e_i^{t+1}}^2\leq \varepsilon_i$. \QEDB

Recall the non-negative function $V^t\equiv V(w^t,y^t,\theta^t)$ defined in (\ref{V^t}):
\begin{gather*}
    V^t = \norm{\nabla_\theta \mathcal{L}^t}^2+\sum_{i=1}^m \left(\norm{\nabla_{w_i}\mathcal{L}^t}^2+\norm{w_i^t-\theta^t}^2\right).
\end{gather*}
We establish the convergence of the \texttt{FedADMM} in the following.

\noindent \textbf{Theorem 1}. Let assumptions 1 and 2 hold. Assume each client has a probability of participating at a given  round that is lower bounded by a positive constant $p_{\mathrm{min}}>0$. Consider $\eta=\abs{S^t}/m$, and select $\rho>(1+\sqrt{5})L$, then the following holds:
\begin{gather}
   \tfrac{1}{mT}\sum_{t=0}^{T-1} \mathbb{E}[V^t] \leq \tfrac{1}{mT}\tfrac{c_2}{c_1}\left(\mathcal{L}^0-f^\star+\tfrac{m}{2L}\varepsilon_{\mathrm{max}}\right)+c_3\varepsilon_{\mathrm{max}} ,\tag{\ref{theorem1}}
\end{gather}
where $\varepsilon_\mathrm{max}= \max_i\,\varepsilon_i$, $c_1 = p_{\mathrm{min}}\left(\tfrac{\rho-2L}{2}-\tfrac{2L^2}{\rho}\right)$, $c_2 =3(L^2+\rho^2)+2(1+\tfrac{2L^2}{\rho^2})$, and $c_3 = 3+\tfrac{16}{\rho^2}+\tfrac{c_2}{c_1}\cdot \tfrac{\rho+16L}{2L\rho}$.

\noindent \textit{Proof}: By definition, the following holds:
$$
     \norm{\nabla_{w_i}\mathcal{L}_i^t}^2 = \norm{\nabla f_i(w_i^t)+y_i^t+\rho(w_i^t-\theta^t)}^2.
$$
Recall (\ref{error}) where 
$$
       e_i^{t+1} = \nabla f_i(\widehat{w}_i^{t+1})+y_i^t+ \rho(\widehat{w}_i^{t+1}-\theta^t). \nonumber 
$$
Therefore, we obtain
\begin{align*}
 \norm{\nabla_{w_i}\mathcal{L}_i^t}&=\norm{\nabla f_i(w_i^t)+y_i^t+\rho(w_i^t-\theta^t)}\\
 & = \bigg|\bigg|\nabla f_i(w_i^t)+y_i^t+\rho(w_i^t-\theta^t)\\
 &\quad-\bigg(\nabla f_i(\widehat{w}_i^{t+1})+y_i^t+ \rho(\widehat{w}_i^{t+1}-\theta^t)\bigg)+e_i^{t+1}\bigg|\bigg|\\
 =&\norm{\nabla f_i(w_i^{t})-\nabla f_i(\widehat{w}_i^{t+1})+\rho(w_i^t-\widehat{w}_i^{t+1})+e_i^{t+1}} \\
 \leq &  \norm{\nabla f_i(w_i^t)-\nabla f_i(\widehat{w}_i^{t+1})}+\rho\norm{w_i^t-\widehat{w}_i^{t+1}}+\norm{e_i^{t+1}}
\end{align*}
Using assumption 1, and the accuracy for inexact primal updates, we obtain 
\begin{align*}
    \norm{\nabla_{w_i}\mathcal{L}_i^t}^2\leq 3\bigg(\left(L^2+\rho^2\right)\norm{\widehat{w}_i^{t+1}-w_i^t}^2+\varepsilon_i\bigg).
\end{align*}
Therefore, it holds that
\begin{gather}
    \sum_{i=1}^m \norm{\nabla_{w_i}\mathcal{L}_i^t}^2 \leq 3 \sum_{i=1}^m \left( \bigg(L^2+\rho^2\bigg)\norm{\widehat{w}^{t+1}_i-w_i^t}^2+\varepsilon_i\right).\label{theorem_comp1}
\end{gather}
Moreover, 
\begin{align}
    \sum_{i=1}^m \norm{w_i^t-\theta^t}^2 &=\sum_{i=1}^m \norm{w_i^t-\widehat{w}_i^{t+1}+\widehat{w}_i^{t+1}-\theta^t}^2  \nonumber\\
    &\leq \sum_{i=1}^m 2\bigg(\norm{\widehat{w}_i^{t+1}-w_i^t}^2+\norm{\widehat{w}_i^{t+1}-\theta^t}^2\bigg). \nonumber
\end{align}
From the virtual update for $\widehat{y}_i^{t+1}$ in (\ref{virtual}), 
\begin{gather}
    \norm{\widehat{w}_i^{t+1}-\theta^t}^2= \tfrac{1}{\rho^2}\norm{\widehat{y}_i^{t+1}-y_i^t}^2.\label{1_rho_y}
\end{gather}
Therefore, we obtain:
\begin{gather}
    \sum_{i=1}^m \norm{w_i^t-\theta^t}^2 \leq \sum_{i=1}^m 2\bigg(\norm{\widehat{w}_i^{t+1}-w_i^t}^2+\tfrac{1}{\rho^2}\norm{\widehat{y}_i^{t+1}-y_i^t}^2\bigg) \nonumber\\
    \leq 2\sum_{i=1}^m \left(\bigg(1+\tfrac{2 L^2}{\rho^2}\bigg)\norm{\widehat{w}_i^{t+1}-w_i^t}^2 +\tfrac{8\varepsilon_i}{\rho^2}\right),\label{theorem1_comp2}
\end{gather}
where the last inequality follows from Lemma 1. From (\ref{zero_deri}) we obtain:
\begin{align}
    V^t &= \underbrace{\norm{\nabla_{\theta}\mathcal{L}^t}^2}_{\text{ $=0$}}+\sum_{i=1}^m \bigg( \norm{\nabla_{w_i}\mathcal{L}_i^t}^2+\norm{w_i^t-\theta^t}^2\bigg)\nonumber\\
    &\underset{(\ref{theorem_comp1})-(\ref{theorem1_comp2})}{\leq} \sum_{i=1}^m \bigg(3(L^2+\rho^2)+2(1+\tfrac{2L^2}{\rho^2})\bigg)\norm{\widehat{w}_i^{t+1}-w_i^t}^2\nonumber\\
    &\quad +\sum_{i=1}^m \bigg(3
    +\tfrac{16}{\rho^2}\bigg)\varepsilon_i\nonumber\\
    &\leq \sum_{i=1}^m c_2 \norm{\widehat{w}_i^{t+1}-w_i^t}^2+\left(3m+\tfrac{16m}{\rho^2}\right)\varepsilon_{\mathrm{max}},\label{VV}
\end{align}
where the last inequality follows from the definitions: $
  \varepsilon_{\mathrm{max}}= \underset{i}{\mathrm{max}}\,\,\varepsilon_i ,
    c_2 = 3(L^2+\rho^2)+2(1+\tfrac{2L^2}{\rho^2})
$. From Lemma 2, we have
$
    \mathcal{L}^{t+1}-\mathcal{L}^t\leq \sum_{i\in S^t}\Bigg(\bigg(\tfrac{2L-\rho}{2}+\tfrac{2L^2}{\rho }\bigg)\norm{w_i^{t+1}-w_i^t}^2
    +\tfrac{\varepsilon_i}{2L}+\tfrac{8\varepsilon_i}{\rho}\Bigg)
    -\tfrac{m\rho}{2}\norm{\theta^{t+1}-\theta^t}^2.
$
After taking conditional expectation at step $t$, by using linearity of expectation and rearranging, we obtain:
\begin{gather}
     \mathbb{E}^t\left[\mathcal{L}^t-\mathcal{L}^{t+1}\right]\geq p_{\mathrm{min}}\sum_{i=1}^m \left(\tfrac{\rho-2L}{2}-\tfrac{2L^2}{\rho}\right)\norm{\widehat{w}_i^{t+1}-w_i^t}^2\nonumber\\
     -m\left(\tfrac{\varepsilon_{\mathrm{max}}}{2L}+\tfrac{8\varepsilon_{\mathrm{max}}}{\rho}\right)+\tfrac{m\rho}{2}\norm{\overline{\theta}^{t+1}-\theta^t}^2, \label{condi}
\end{gather}
where we denote $\mathbb{E}^t\left[\theta^{t+1}\right]=\overline{\theta}^{t+1}$ and used Jensen's inequality:
$
    \mathbb{E}^t\left[\norm{\theta^{t+1}-\theta^t}^2\right]\geq \norm{\mathbb{E}^t\left[\theta^{t+1}\right]-\theta^t}^2
    =\norm{\overline{\theta}^{t+1}-\theta^t}^2.
$
Defining
$
    c_1 = p_{\mathrm{min}}(\tfrac{\rho-2L}{2}-\tfrac{2L^2}{\rho})
$, we rewrite (\ref{condi}) as
\begin{gather}
    \mathbb{E}^t[\mathcal{L}^t-\mathcal{L}^{t+1}]+m\left(\tfrac{\varepsilon_{\mathrm{max}}}{2L}+\tfrac{8\varepsilon_{\mathrm{max}}}{\rho}\right)\geq c_1 \sum_{i=1}^m \norm{\widehat{w}_i^{t+1}-w_i^t}^2.\label{condi AL}
\end{gather}
Using (\ref{VV}) and (\ref{condi AL}) leads to:
\begin{gather}
    V^t\leq \tfrac{c_2}{c_1}\mathbb{E}^t[\mathcal{L}^t-\mathcal{L}^{t+1}]+mc_3\varepsilon_{\mathrm{max}},
\end{gather}
where $c_3 = 3+\tfrac{16}{\rho^2}+\tfrac{c_2}{c_1}(\tfrac{1}{2L}+\tfrac{8}{\rho})$. Taking expectation on both sides gives 
$$
    \mathbb{E}[V^t] \leq \tfrac{c_2}{c_1}\mathbb{E}[\mathcal{L}^t-\mathcal{L}^{t+1}]+mc_3 \varepsilon_{\mathrm{max}}.
$$After telescoping the above and dividing by $mT$, we obtain:
\begin{gather*}
    \tfrac{1}{mT}\sum_{t=0}^{T-1} \mathbb{E}\left[V^t\right]\leq \tfrac{1}{mT} \tfrac{c_2}{c_1}\mathbb{E}\left[\mathcal{L}^0-\mathcal{L}^{T-1}\right]+c_3\varepsilon_{\mathrm{max}}.
\end{gather*}From Lemma 3, it holds that $\mathbb{E}[\mathcal{L}^0-\mathcal{L}^{T-1}]\leq \mathcal{L}^0-f^\star+\tfrac{m}{2L} \varepsilon_{\mathrm{max}}$. Therefore, the following holds:
\begin{gather*}
    \tfrac{1}{mT}\sum_{t=0}^{T-1} \mathbb{E}[V^t] \leq \tfrac{1}{mT}\tfrac{c_2}{c_1}\left(\mathcal{L}^0-f^\star+\tfrac{m}{2L}\varepsilon_{\mathrm{max}}\right)+c_3\varepsilon_{\mathrm{max}},
\end{gather*}
which is the desired. 
\QEDB

\newpage
\bibliographystyle{ieeetr}
\bibliography{Ref.bib}

\end{document}